\documentclass{article}

     \PassOptionsToPackage{numbers,sort,compress}{natbib}

\usepackage[preprint]{neurips_2024}

\usepackage[utf8]{inputenc} 
\usepackage[T1]{fontenc}    
\usepackage[backref=page]{hyperref}       
\usepackage{url}            
\usepackage{booktabs}       
\usepackage{microtype}      
\usepackage[dvipsnames,table]{xcolor}  
\definecolor{yellowmine}{RGB}{205,63,0}
\definecolor{redmine}{RGB}{255,0,63}

\hypersetup{
	hidelinks,
	colorlinks=true,
    linkcolor=Blue,
	filecolor=Blue,
	citecolor=Blue,
}
\usepackage{amsfonts}       
\usepackage{nicefrac}       
\usepackage{amsmath}
\usepackage{amssymb}
\usepackage{mathtools}
\usepackage{amsthm}
\usepackage{bm}
\usepackage{multirow}

\usepackage[capitalize]{cleveref}

\usepackage[edges]{forest}
\usepackage{graphicx}
\usepackage[labelformat=simple]{subcaption}

\graphicspath{{./media/}}
\DeclareGraphicsExtensions{.pdf,.png}

\newenvironment{nscenter}
 {\parskip=0pt\par\nopagebreak\centering}
 {\par\noindent\ignorespacesafterend}

\def\ie{\emph{i.e.,}~}
\def\eg{\emph{e.g.,}~}

\def\eqref#1{equation~\ref{#1}}

\def\1{\bm{1}}

\def\rc{{\textnormal{c}}}

\def\rvl{{\mathbf{l}}}

\def\rvn{{\mathbf{n}}}

\def\rvs{{\mathbf{s}}}

\def\rvx{{\mathbf{x}}}
\def\rvy{{\mathbf{y}}}

\def\rmA{{\mathbf{A}}}

\def\ermA{{\textnormal{A}}}

\def\vmu{{\bm{\mu}}}
\def\valpha{{\bm{\alpha}}}
\def\vtheta{{\bm{\theta}}}

\def\vvarphi{{\bm{\varphi}}}

\def\vb{{\bm{b}}}

\def\vl{{\bm{l}}}

\def\vp{{\bm{p}}}
\def\vq{{\bm{q}}}

\def\vs{{\bm{s}}}

\def\vx{{\bm{x}}}
\def\vy{{\bm{y}}}

\def\mA{{\bm{A}}}

\def\mI{{\bm{I}}}

\def\mP{{\bm{P}}}

\def\mX{{\bm{X}}}

\def\mSigma{{\bm{\Sigma}}}

\DeclareMathAlphabet{\mathsfit}{\encodingdefault}{\sfdefault}{m}{sl}
\SetMathAlphabet{\mathsfit}{bold}{\encodingdefault}{\sfdefault}{bx}{n}

\def\gC{{\mathcal{C}}}

\def\sR{{\mathbb{R}}}

\def\emA{{A}}

\def\emP{{P}}

\newcommand{\E}{\mathbb{E}}

\newcommand{\R}{\mathbb{R}}

\newcommand{\softmax}{\mathrm{softmax}}

\DeclareMathOperator*{\argmin}{arg\,min}

\newcommand{\beginsupplement}{%
    \renewcommand{\theequation}{A\arabic{equation}}
    \renewcommand{\thefigure}{A\arabic{figure}}
    \setcounter{figure}{0}
    \setcounter{equation}{0}  
}

\title{Hierarchical Uncertainty Exploration via\\Feedforward Posterior Trees}

\author{%
  Elias Nehme, \\
  Electrical and Computer Engineering\\
  Technion--Israel Institute of Technology\\
  Haifa, Israel \\
  \texttt{seliasne@campus.technion.ac.il} \\
  \And
  Rotem Mulayoff, \\
  Electrical and Computer Engineering\\
  Technion--Israel Institute of Technology\\
  Haifa, Israel \\
  \texttt{smulayof@campus.technion.ac.il} \\
  \And
  Tomer Michaeli, \\
  Electrical and Computer Engineering\\
  Technion--Israel Institute of Technology\\
  Haifa, Israel \\
  \texttt{tomer.m@ee.technion.ac.il} \\
}

\begin{document}

\maketitle

\begin{abstract}

When solving ill-posed inverse problems, one often desires to explore the space of potential solutions rather than be presented with a single plausible reconstruction. Valuable insights into these feasible solutions and their associated probabilities are embedded in the posterior distribution. However, when confronted with data of high dimensionality (such as images), visualizing this distribution becomes a formidable challenge, necessitating the application of effective summarization techniques before user examination.
In this work, we introduce a new approach for visualizing posteriors across multiple levels of granularity using \emph{tree}-valued predictions. Our method predicts a tree-valued hierarchical summarization of the posterior distribution for any input measurement, in a single forward pass of a neural network.
We showcase the efficacy of our approach across diverse datasets and image restoration challenges, highlighting its prowess in uncertainty quantification and visualization. Our findings reveal that our method performs comparably to a baseline that hierarchically clusters samples from a diffusion-based posterior sampler, yet achieves this with orders of magnitude greater speed.

\end{abstract}

\section{Introduction}\label{sec:intro}
Communicating prediction uncertainty is a crucial element in advancing the reliability of machine learning models. This is particularly important in the realm of imaging inverse problems, in which a given input can typically be associated with a multitude of plausible solutions. In such cases, it is advantageous to equip users with efficient tools for exploring and visualizing the set of admissible solutions. Such tools may be of high value especially in safety-critical domains, such as scientific and medical image analysis \citep{ounkomol2018label, christiansen2018silico, rivenson2019virtual, falk2019u}, where inaccuracies in predictions could impact human lives.

Information about the plausible solutions and their respective likelihoods is encapsulated within the posterior distribution. However, high-dimensional posteriors are challenging to visualize. One popular approach for communicating uncertainty is to generate samples from the posterior \citep{song2021solving,ohayon2021high,kawar2022denoising,wang2023ddnm,chung2022diffusion,li2022mat,lugmayr2022repaint,saharia2022image,saharia2022palette,bendel2022regularized}. Yet, in complex domains with high uncertainty levels, users may need to sift through hundreds of posterior samples per input to be able to confidently validate or refute suspicions about the unobserved ground-truth image \citep{cohen2023posterior}. Several methods have been proposed for generating a concise set of samples that highlight posterior diversity \citep{sehwag2022generating, cohen2023posterior}. However, choosing this set to be small is often insufficient for summarizing the posterior, while choosing it to be large hinders the user's ability to properly inspect all solutions in the set. Other methods proposed to visualize uncertainty by allowing the user to navigate along the principal components (PCs) of the posterior \citep{nehme2023uncertainty,belhasin2023principal,manor2024posterior,yair2023uncertainty}. While more interactive in nature, the projection onto a low dimensional principal subspace commonly accounts for only a small fraction of the reconstruction error.

\begin{figure}
\begin{subfigure}{1cm}
    \centering
    $ \scriptstyle \text{GT}$\\
    \includegraphics[scale=0.12]{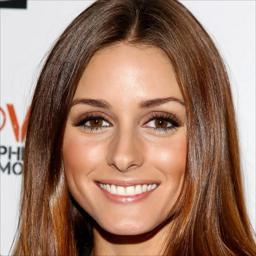}
\end{subfigure}
\begin{subfigure}{1cm}
    \centering
    $ \scriptstyle \text{Input}$\\
\includegraphics[scale=0.12]{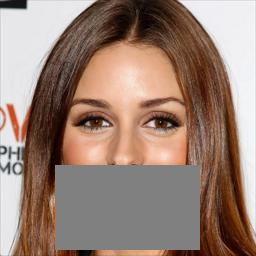}
\end{subfigure}

\vspace{-1.2cm}
\centering
\begin{forest}
  before typesetting nodes={
    for tree={
        font=\footnotesize,
      content/.wrap 2 pgfmath args={    {\footnotesize #2}\\\includegraphics[scale=0.12]{#1}}{content()}{title()},
    },
    where={isodd(n_children())}{calign=child, calign child/.wrap pgfmath arg={#1}{int((n_children()+1)/2)}}{},
  },
  forked edges,
  /tikz/every node/.append style={font=\footnotesize},
  for tree={
    parent anchor=children,
    child anchor=parent,
    align=center,
    l sep'= 1mm,
    s sep'= 2.5mm,
    fork sep'=0mm,
    inner sep = -2.5pt,
    edge={line width=0.5pt},
    font=\footnotesize,
  },
  [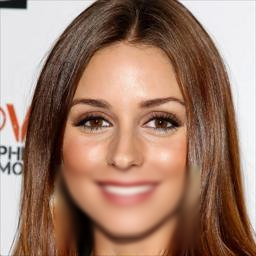, title={ {$ \scriptstyle \text{MMSE}$}}
    [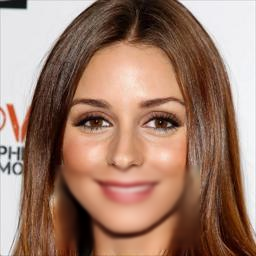, title={ $\scriptstyle p=0.17$}
      [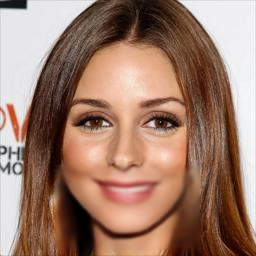, title={ $\scriptstyle p=0.13$}]
      [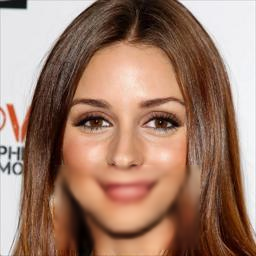, title={ $\scriptstyle p=0.00$}]
      [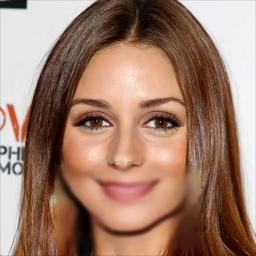, title={ $\scriptstyle p=0.04$}]
    ]
    [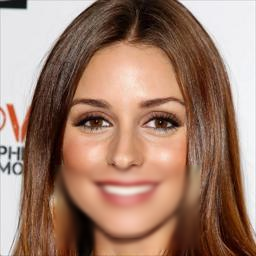, title={ $\scriptstyle p=0.50$}
      [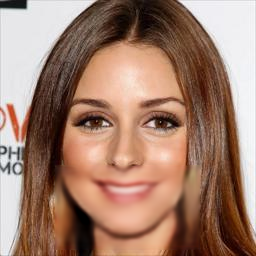, title={ $\scriptstyle p=0.06$}]
      [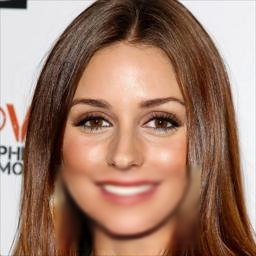, title={ $ \scriptstyle p=0.14$}]
      [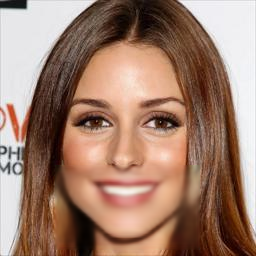, title={ $\scriptstyle p=0.30$}]
    ]
    [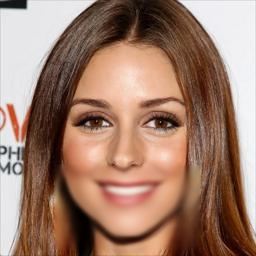, title={ $\scriptstyle p=0.33$}
      [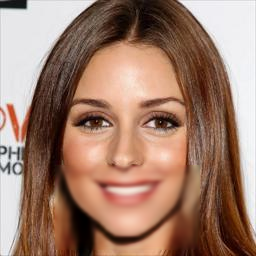, title={{ $\scriptstyle p=0.13$}}]
      [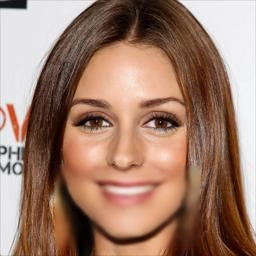, title={$\scriptstyle p=0.13$}]
      [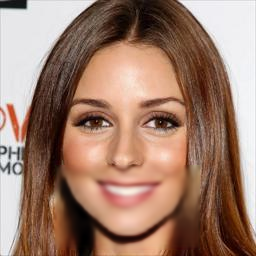, title={ $\scriptstyle p=0.07$}]
    ]
  ]
\end{forest}
\caption{\textbf{Hierarchical decomposition of the minimum-MSE predictor into prototypes in the task of mouth inpainting}. The predicted tree explores the different options of bigger/smaller lips, mouth opening/closing, round/square jawline, etc.}
\label{fig:teaser}
\end{figure}

In this work, we propose to model posterior uncertainty through \emph{hierarchical} clustering, by using tree-structured outputs. In stark contrast to all existing methods, our approach supports efficient user interaction, allowing examination of only a small number of hypotheses en route to accepting or refuting a hypothesis about the unknown image. Specifically, our model receives a degraded measurement and outputs a hierarchy of predictions that cluster the solution set at multiple levels of granularity, each accompanied by their probability. This allows us to visualize the likelihood of different posterior modes, facilitating an informed visual exploration of posterior uncertainty (\cref{fig:teaser}). Our method, which we coin \emph{posterior trees}, is a generic technique for uncertainty visualization that is seamlessly transferable across tasks and datasets. Compared to the popular practice of training point estimators with mean square error (MSE) minimization, our method is an ideal drop-in replacement. This is because it adds a minor additional computational burden, while simultaneously accompanying the point estimate with its decomposition to the constituting clusters, revealing the underlying modes of variation and disclosing posterior uncertainty.

We showcase our method on multiple inverse problems in imaging, demonstrating the practical benefit of tree-valued predictions. We also quantitatively compare our learned posterior trees to a two-step baseline that generates samples using a diffusion-based posterior sampler and then applies hierarchical K-means clustering. Our approach achieves a comparable performance across tasks and datasets while enjoying a significantly faster runtime.

\forestset{
  declare toks={title}{},
}


\section{Related Work}\label{sec:related}

The common avenue for probing data uncertainty (also known as aleatoric uncertainty \cite{kendall2017uncertainties,wang2019aleatoric,ohayon2021high,angelopoulos2022image,kawar2022denoising,wang2023ddnm}) involves posterior sampling using conditional generative models like \citep{lugmayr2020srflow, ohayon2021high, song2021solving, kawar2021snips, kawar2022denoising, chung2022diffusion, li2022mat, lugmayr2022repaint, saharia2022palette, saharia2022image, bendel2022regularized, wang2023ddnm, chung2023fast,feng2023score}, with the dominance of score-based/diffusion models \citep{hyvarinen2005estimation, sohl2015deep, song2019generative, ho2020denoising} in recent years. While posterior sampling theoretically offers a sizable solution set per input, allowing useful uncertainty estimates through summarization methods like PCA \citep{belhasin2023principal} or $K$-means, its implementation proves impractically slow for contemporary state-of-the-art models. Despite attempts to enhance speed \citep{salimans2022progressive, luhman2021knowledge, meng2022on, song2023consistency}, the strategy remains plagued by extended run times.

In an effort to expedite inference, some studies proposed approximating posteriors with simpler distributions considering pixel correlations, such as correlated Gaussians \citep{dorta2018structured, monteiro2020stochastic, meng2021estimating, nussbaum2022structuring}. More recently, two approaches emerged, directly outputting the top principal components (PCs) of the posterior distribution without imposing distributional assumptions \citep{nehme2023uncertainty,manor2024posterior}. Additionally, subsequent work by \citet{yair2023uncertainty} visualized the \textit{projected} posterior distribution onto the subspace spanned by the top PCs, unveiling its multi-modal nature at heightened uncertainty levels.

\emph{Multiple choice learning} (MCL) \citep{guzman2012multiple,guzman2014efficiently,lee2015m,lee2016stochastic,lee2017confident,rupprecht2017learning,firman2018diversenet,ilg2018uncertainty,tian2019versatile,stein2022double,letzelter2023resilient} was proposed as a framework that can provide a multi-modal set of predictions when confronted with ambiguous inputs. The idea was originally proposed by \citet{guzman2012multiple} using structured support vector machines, and was later introduced to deep models in \citep{lee2015m,lee2016stochastic}. The main working principle of MCL is to train an architecture with multiple output heads using the \emph{oracle} loss, such that for a given input, only the head that is closest to the desired output gets updated. This winner-takes-all strategy encourages each head to specialize in a different mode of the output space, effectively predicting the centers of a Voronoi tessellation of all possible outputs \citep{rupprecht2017learning}. Later works such as \citep{tian2019versatile,letzelter2023resilient} proposed to complement each prediction in the set with its likelihood. 
Nonetheless, existing multi-hypothesis methods are mainly tailored to classification settings, and are still limited in their ability to organize the predicted set of hypotheses effectively. Categorizing the solution set \emph{hierarchically}, as we propose here, can significantly accelerate uncertainty exploration. For example, it allows iterative user interaction, focusing user efforts on the examination of a small number of hypotheses (not necessarily the most likely ones) to confirm/refute a suspicion about the signal. Moreover, the hierarchical structure can regularize the number of prototypes devoted to high-density modes, highlighting rare cases.


\section{Method}\label{sec:method}
Our goal is to predict a clean signal $\vx\in\mathcal{X}\subseteq\mathbb{R}^{d_x}$, given some degraded measurements $\vy\in\mathcal{Y}\subseteq\mathbb{R}^{d_y}$. We assume that $\vx$ and $\vy$ are realizations of random vectors $\rvx$ and $\rvy$ with an unknown joint distribution $p_{\rvx,\rvy}(\vx,\vy)$, from which we have a training set of \emph{i.i.d}.\ samples $\smash{\mathcal{D}=\{(\vx_i,\vy_i)\}_{i=1}^{N}}$. The objective of our predicted trees is therefore to visualize the uncertainty in the posterior distribution $p_{\rvx|\rvy}(\vx|\vy)$ via a hierarchical set of a few prototypes.

\subsection{Multiple Output Prediction}\label{subsec:multi-out}

Many image restoration methods output a single prediction $\hat{\vx}$ for any given input $\vy$. A common choice is to aim for the posterior mean $\hat{\vx}=\E[\rvx|\rvy=\vy]$, which is the predictor that minimizes the MSE. However, a single prediction does not convey to the user the uncertainty in the restoration, especially if the solution set is multi-modal. 
When confronted with such ambiguous tasks, it is more natural to predict a small set of prototypical restorations $\{\hat{\vx}_1,\dots,\hat{\vx}_K\}$ that portray the different options. In MCL, this is achieved by minimizing the so-called oracle/winner-takes-all loss given by
\begin{equation}
     \mathcal{L}_{\mathrm{MCL}}\left(\vx, \{\hat{\vx}_i\}_{i=1}^K\right) = \min_{i=1,\dots,K} {\ell(\vx, \hat{\vx}_i)},
\label{eq:mcl-obj}
\end{equation}
where $\ell(\cdot,\cdot)$ is a loss function that measures prediction distance such as the $\ell_2$ loss. Note that $\mathcal{L}_{\mathrm{MCL}}$ is minimized as long as one restoration in the predicted set is close to $\vx$. This encourages prediction diversity, such that each restoration specializes in a different mode of the posterior. \citet{rupprecht2017learning} proposed a probabilistic interpretation of MCL, showing that the optimal minimizers of \cref{eq:mcl-obj} form a centroidal Voronoi tessellation (CVT) of the posterior, where $\{\hat{\vx}_i^{\star}\}_{i=1}^K$ are given by the conditional means/centroids of the resulting $K$ Voronoi cells. In a discrete setting where we are given samples from $p_{\rvx|\rvy}(\vx|\vy)$, CVT with the $\ell_2$ loss is equivalent to $K$-means clustering.

Our goal here is to generalize the concept of predicting a set of restorations. Given an input $\vy$, instead of predicting an unordered set $\{\hat{\vx}_1(\vy),\dots,\hat{\vx}_K(\vy)\}$, we propose to output an input-adaptive tree $\mathcal{T}(\vy)$ representing a hierarchical clustering of the posterior $p_{\rvx|\rvy}(\vx|\vy)$. Specifically, we want the tree nodes to correspond to a hierarchical CVT of the posterior (analogous to a hierarchical $K$-means process in the discrete setting), where the children of each node constitute a CVT for the cluster represented by the node. Such a tree can organize the different restorations in a manner that facilitates their navigation by a user and can also accompany each node with its relative cluster probability.  

\begin{figure}[t]
\centering
\includegraphics[width=1.0\textwidth]{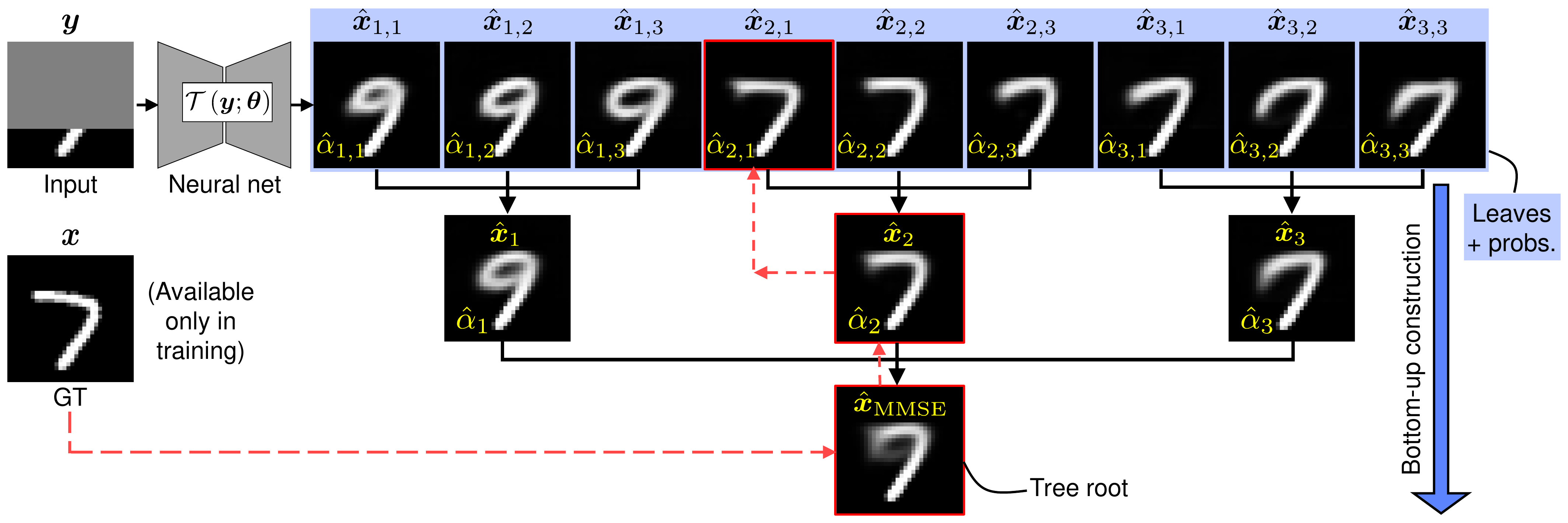}
\caption{\textbf{Method overview}. Our model $\mathcal{T}(\vy;\vtheta)$ receives a degraded image $\vy$ and predicts $\{\hat{\vx}_{k_1,\dots,k_d}\}_{k_1,\dots,k_d=1}^K$, the bottom $K^d$ leaves, and their probabilities $\{\alpha_{k_1,\dots,k_d}\}_{k_1,\dots,k_d=1}^K$ (faint blue box; illustrated here for $K=3$ and $d=2$). Next, the tree is iteratively constructed from the bottom up using weighted averaging, until we reach the root node which is the minimum MSE predictor $\hat{\vx}_{\text{MMSE}}$. During training, starting from the root, the ground truth $\vx$ is propagated through the tree until it reaches the leaves (dashed red lines). At tree level $d$, $x$ is compared to its immediate $K$ children nodes, and the MSE loss to the nearest child is added to the loss trajectory.}
\label{fig:method}
\end{figure}

\subsection{Posterior Trees}\label{subsec:bottom-up-comp}

As discussed above, training a model to predict a tree of degree $K$ and depth $d=1$ with the oracle loss, results in a CVT of the posterior $p_{\rvx|\rvy}(\vx|\vy)$ for every $\vy$. Let $\{\mathcal{A}_k(\vy)\}_{k=1}^K$ denote the resulting Voronoi cells at $d=1$ for some input $\rvy=\vy$, such that $\cup_{k=1}^K\overline{\mathcal{A}_k(\vy)}=\mathcal{X}$, with $\mathcal{X}\in\R^{d_x}$ being the output space. Training with $\ell(\cdot,\cdot) = \ell_2(\hat{\vx},\vx)$, the optimal $K$ predicted leaves and probabilities are given by \citep{rupprecht2017learning,letzelter2023resilient}
\begin{align}
    \vx_k^{\star}(\vy)&=\E[\rvx|\rvy=\vy,\rvx\in\mathcal{A}_k(\vy)], \nonumber \\
    \alpha_k^{\star}(\vy)&=\mathbb{P}(\rvx\in\mathcal{A}_k(\vy)|\rvy=\vy), \quad k=1,\dots,K.
    \label{eq:cond-mean-d1}
\end{align}
We now want to extend the tessellation to the children of each node at the next level. Let $\{\mathcal{A}_{k,q}(\vy)\}_{q=1}^K$ denote the Voronoi cells at $d=2$, partitioning/tessellating the Voronoi cell $\mathcal{A}_k(\vy)$ from level 1 into $K$ sub cells such that $\cup_{q=1}^K\overline{\mathcal{A}_{k,q}(\vy)}=\mathcal{A}_k(\vy)$. We want the $K^2$ leaves and their associated probabilities outputted by our model to be given by
\begin{align}
    \vx_{k,q}^{\star}(\vy)&=\E[\rvx|\rvy=\vy,\rvx\in\mathcal{A}_{k,q}(\vy)], \nonumber \\
    \alpha_{k,q}^{\star}(\vy)&=\mathbb{P}(\rvx\in\mathcal{A}_{k,q}(\vy)|\rvy=\vy), \quad k,q=1,\dots,K.
    \label{eq:cond-mean-d2}
\end{align}
For simplicity of notation, in the following, we omit the dependence of the Voronoi cells on $\vy$, and write $\mathcal{A}_{k}/\mathcal{A}_{k,q}$ instead. \Cref{eq:cond-mean-d1,eq:cond-mean-d2} expose an interesting connection between tree levels. First, recall that since $\{\mathcal{A}_{k,q}\}_{k,q}$ is a tessellation, it satisfies $\cup_{q=1}^K\overline{\mathcal{A}_{k,q}}=\mathcal{A}_k$, and $\mathcal{A}_{k_1,q_1}\cap\mathcal{A}_{k_2,q_2}=\emptyset$ for $(k_1,q_1)\neq(k_2,q_2)$. Invoking the law of total expectation we can write
\begin{equation}
    \E[\rvx|\vy,\rvx\in\mathcal{A}_k] = \sum_{q=1}^K\E[\rvx|\rvy=\vy,\rvx\in\mathcal{A}_k,\rvx\in\mathcal{A}_{k,q}]\mathbb{P}(\rvx\in\mathcal{A}_{k,q}|\rvy=\vy,\rvx\in\mathcal{A}_k).
    \label{eq:cond-mean-d2-to-d1}
\end{equation}
\Cref{eq:cond-mean-d2-to-d1} can be simplified by noting that $\mathcal{A}_k\cap\mathcal{A}_{k,q}=\mathcal{A}_{k,q}$, hence the expectation in the summand is given by $\E[\rvx|\rvy=\vy,\rvx\in\mathcal{A}_{k,q}]$. Moreover, we further simplify the probability in \cref{eq:cond-mean-d2-to-d1} by writing
\begin{align}
    \mathbb{P}(\rvx\in\mathcal{A}_{k,q}|\rvy=\vy,\rvx\in\mathcal{A}_k) & = \frac{\mathbb{P}(\rvx\in\mathcal{A}_{k}|\rvy=\vy,\rvx\in\mathcal{A}_{k,q})\mathbb{P}(\rvx\in\mathcal{A}_{k,q}|\vy)}{\mathbb{P}(\rvx\in\mathcal{A}_{k}|\vy)}  \nonumber\\
    & =\frac{\mathbb{P}(\rvx\in\mathcal{A}_{k,q}|\vy)}{\sum_{q=1}^K{\mathbb{P}(\rvx\in\mathcal{A}_{k,q}|\vy)}},
    \label{eq:prob-d2-to-d1-total-prob}
\end{align}
where the first equality follows from Bayes' rule and the second equality is by the law of total probability and the fact that $\mathcal{A}_{k,q}\subseteq\mathcal{A}_k$, which implies that $\mathbb{P}(\rvx\in\mathcal{A}_{k}|\rvy=\vy,\rvx\in\mathcal{A}_{k,q})=1$.

Substituting \cref{eq:prob-d2-to-d1-total-prob} and the simplified expectation in \cref{eq:cond-mean-d2-to-d1} with the notations of \cref{eq:cond-mean-d1,eq:cond-mean-d2}, we obtain that
\begin{align}
    \vx_k^{\star}(\vy) & = \sum_{q=1}^K {\vx_{k,q}^{\star}(\vy)\frac{\valpha_{k,q}^{\star}(\vy)}{\sum_{q=1}^K{\valpha_{k,q}^{\star}(\vy)}}}, \label{eq:d2-to-d1-result-x} \\
    \valpha_k^{\star}(\vy) & = \sum_{q=1}^K {\valpha_{k,q}^{\star}(\vy)}, \quad k=1,\dots,K, \label{eq:d2-to-d1-result-p}
\end{align}
where \cref{eq:d2-to-d1-result-p} is by the law of total probability. Note that \cref{eq:d2-to-d1-result-x,eq:d2-to-d1-result-p} expose the redundancy in predicting both $\{\vx_k^{\star}(\vy),\valpha_k^{\star}(\vy)\}$ and $\{\vx_{k,q}^{\star}(\vy),\valpha_{k,q}^{\star}(\vy)\}$, as given the latter, we can compute the former in closed form. Furthermore, although our derivation assumed a tree of depth $d=2$, the result trivially generalizes to a tree of depth $d$. Given the leaves and their probabilities at level $d$, $\{\vx_{k_1,\dots,k_d}^{\star}(\vy),\valpha_{k_1,\dots,k_d}^{\star}(\vy)\}_{k_1,\dots,k_d=1}^K$, we can compose their parents at the upper $d-1$ levels, by employing \cref{eq:d2-to-d1-result-x,eq:d2-to-d1-result-p} iteratively from the bottom up.

Here, this is precisely the property we exploit to output an input-adaptive tree $\mathcal{T}(\vy)$ of degree $K$ and depth $d$, using a \emph{single} model. Specifically, we train a model $\smash{\mathcal{T}(\vy;\vtheta)\triangleq(\mX_d(\vy;\vtheta),\valpha_d(\vy;\vtheta))}$ that outputs $K^d$ prediction leaves $\smash{\mX_d(\vy;\vtheta)=\{x_k(\vy;\vtheta)\}_{k=1}^{K^d}}$ and their accompanying probabilities $\smash{\valpha_d(\vy;\vtheta)=\{\alpha_k(\vy;\vtheta)\}_{k=1}^{K^d}}$ (see \cref{fig:method}). Next, starting from the bottom leaves at depth $d$, the rest of the tree is composed level by level, recursively employing \cref{eq:d2-to-d1-result-x,eq:d2-to-d1-result-p} from the bottom up. 

During training, the tree hierarchy is enforced through our loss function. Given some input $\vy_i$, the model $\mathcal{T}(\vy;\vtheta)$ outputs are used as explained earlier to construct the full tree, where the root of the tree approximates the posterior mean $\E[\rvx|\rvy=\vy_i]$. Next, the label $\vx_i$ is first compared to the tree root with an $\ell_2$ loss to ensure our decomposition recovers the MMSE estimator. At each successive level, the label $\vx_i$ is compared to the children of the parent with minimal loss, and the closest child is added to the loss trajectory (\cref{fig:method}). This process is repeated until we arrive at the bottom leaves.

Note that the described training scheme serves two purposes. First, it induces the desired tree structure by employing an amortized hierarchical version of the oracle loss in \cref{eq:mcl-obj}. Second, the loss employed on the root node enables predicting branch probability, mitigating the need for supervising the probabilities with explicit targets.

\subsection{Architecture}\label{subsec:arch-opt}

The approach presented in \cref{subsec:bottom-up-comp} is general, and can be used to augment any architecture outputting $K^d$ images and probabilities. Here, we choose to adapt the U-Net architecture \citep{ronneberger2015u} as a generic established choice for image-to-image regression tasks. The number of output images is $K^d$ to accommodate a tree of degree $K$ and depth $d$. In case all features are shared in the architecture, this amounts to changing the number of output filters at the last layer. However, in challenging tasks such as image inpainting, we found that sharing all parameters leads to reduced prediction diversity. In such cases, to trade off feature sharing and output diversity, we use group convolutions \citep{krizhevsky2012imagenet} in the decoder, such that each prediction has a disjoint set of channels learned separately from other outputs. As for the skip connections from the encoder, the concatenated features are interleaved equally per level such that each prediction has an equal share. In addition to the output images, our architecture also predicts $K^d$ probabilities (see \cref{fig:method,fig:app-architecture}). This is achieved by (global) average pooling all feature maps from the decoder, and feeding their concatenation to an additional lightweight \emph{multi-layer perceptron} (MLP), with four linear layers and a $\softmax$ at the output.

\subsection{Preventing Tree Collapse}\label{subsec:prevent-collapse}

Vanilla training with the oracle loss is known to suffer from ``hypotheses collapse'', where some of the predictions are implausible \citep{guzman2012multiple,lee2015m,lee2016stochastic,rupprecht2017learning,firman2018diversenet, brodie2018alpha,makansi2019overcoming,letzelter2023resilient}. This phenomenon occurs when some predictions are initialized worse than others in stochastic training, and therefore receive little to no gradient updates due to not being chosen in the oracle loss. This results in degenerate outputs that are not encouraged to produce meaningful results. Previous works proposed various regularizations to remedy this undesired effect. For example, \citet{rupprecht2017learning} relaxed the $\argmin$ operator with a small constant $\varepsilon$, such that predictions with non-minimal loss get updated with $\varepsilon$-scaled gradients. Moreover, \citet{makansi2019overcoming} proposed an evolving version of the oracle loss, where the top-$k$ predictions are updated in each step, with $k$ being annealed down to $k=1$ during training, such that it starts with equal weights to all predictions and gradually shifts to updating only the best one towards convergence. Here, we employ a strategy that combines the simplicity of \citep{rupprecht2017learning} with the adaptivity of \citep{makansi2019overcoming}. Specifically, we scale the gradients of non-performing predictions with a constant $\varepsilon$ that is annealed during training according to 
\begin{equation}
    \varepsilon_t=\varepsilon_0\exp\{-[t-t_0]_+/2\},
    \label{eq:epsilon-reg}
\end{equation}
where $t$ is the epoch number, and we fixed $\varepsilon_0=1$ and $t_0=5$ for all experiments. At the beginning of training, this has the benefit of bringing all predictions to a reasonable starting point, alleviating sensitivity to initialization and top prediction domination. As training progresses, this regularization is decayed, and only the relevant prediction is chosen for each sample, hence converging to the desired MCL behavior (see \cref{subsec:app-epsilont}).

\subsection{Weighted Sampling}\label{subsec:weighted-sampler}

Albeit the regularization mentioned in \cref{subsec:prevent-collapse}, for tasks with highly imbalanced posteriors, training with Adam \citep{kingma2015adam} leads to trees with a large disparity in leaf quality. This is because less likely leaves get chosen with lower frequency during training, resulting in transient gradients that highly affect the adaptive normalization in Adam. This hypothesis was also validated in matched settings by using SGD with an appropriate learning rate which resulted in significantly slower, yet improved convergence (see \cref{subsec:app-optimizer}). Therefore, to keep the speed advantage provided by Adam, in the task of image inpainting, we opted for a weighted sampler during training. The purpose of this sampler is to roughly balance out the overall number of occurrences at the bottom leaves during training. This is achieved by keeping track of an association matrix $\smash{\rmA\in\sR^{K^d\times N}}$, where $\ermA_{i,j}$ counts the number of times sample $\vx_j$ was associated with leaf $\vl_i$ over some time window. This allows us to estimate the conditional probability $p_{\rvl_i|\rvx_j}(\vl_i|\vx_j)$, and subsequently derive an optimized sample probability to achieve a roughly uniform marginal leaf probability across the entire training set. The loss of each sample is then adjusted to account for this intervention, avoiding tampering with the original posterior probabilities. See \cref{subsec:app-sampler} for more details.

\section{Experiments}\label{sec:exp}
\begin{figure}[t]
  \centering
  \begin{subfigure}{0.25\linewidth}
    \includegraphics[width=\textwidth, trim={1.4cm 1.4cm 1.8cm 1.2cm},clip]{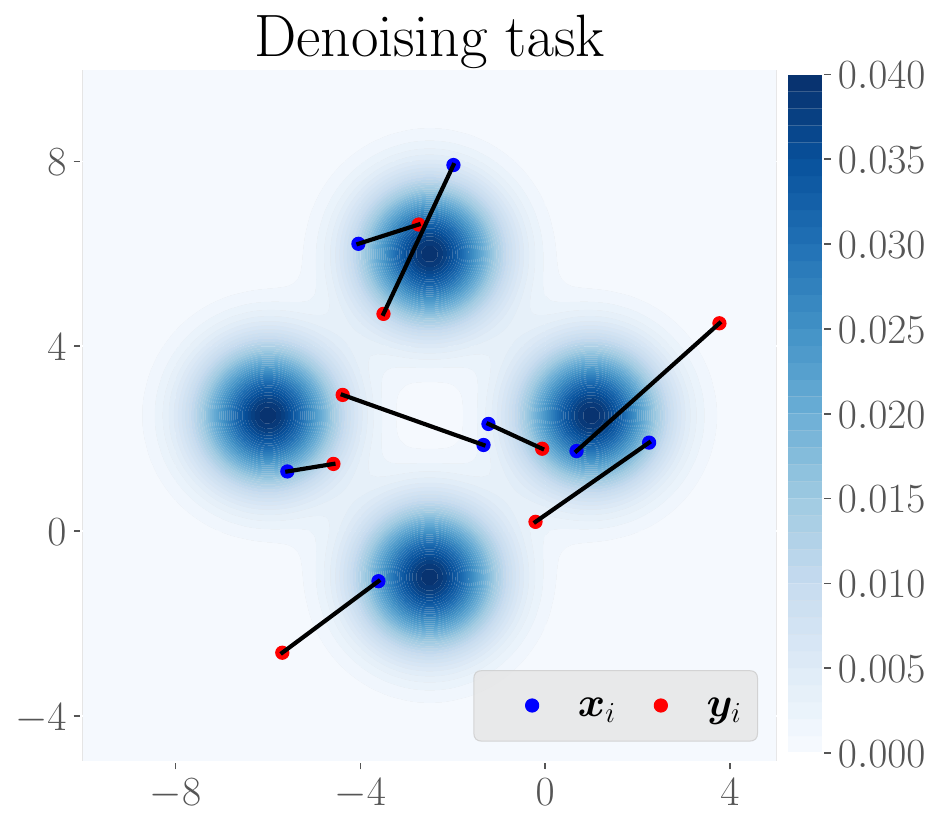}
    \caption{Denoising task}
    \label{fig:gmm-task}
  \end{subfigure}%
  \begin{subfigure}{0.25\linewidth}
    \includegraphics[width=\textwidth, trim={1.4cm 1.4cm 1.8cm 1.2cm},clip]{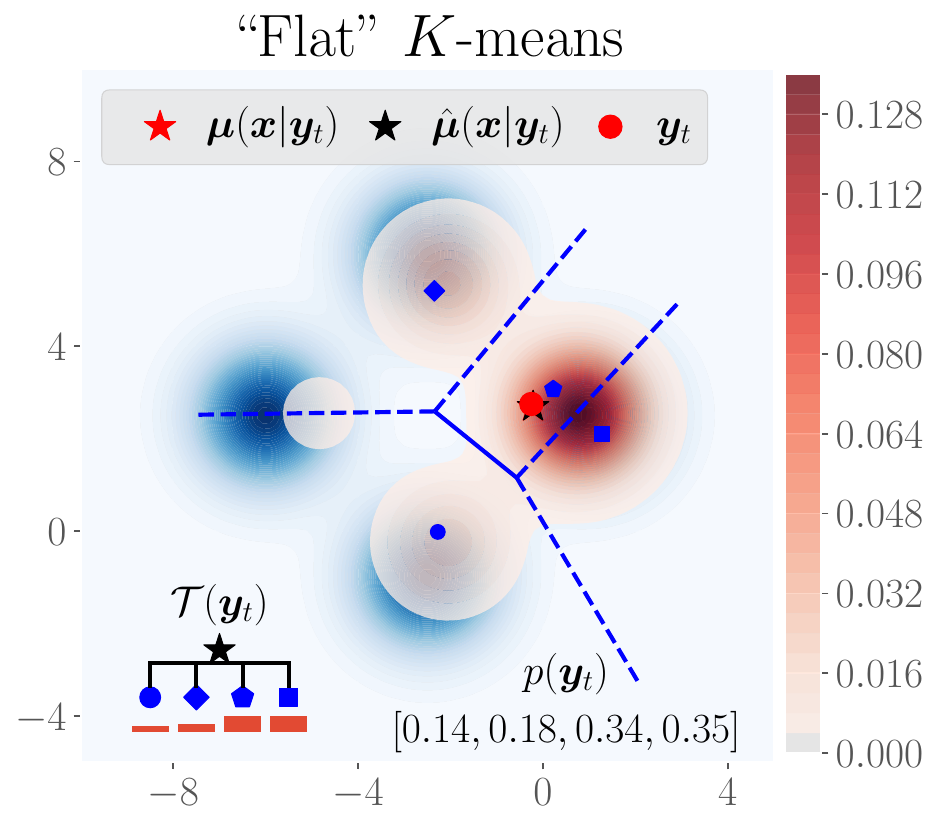}
    \caption{``Flat'' $K$-means}
    \label{fig:gmm-flat-kmeans}
  \end{subfigure}%
  \begin{subfigure}{0.25\linewidth}
    \includegraphics[width=\textwidth, trim={1.4cm 1.4cm 1.8cm 1.2cm},clip]{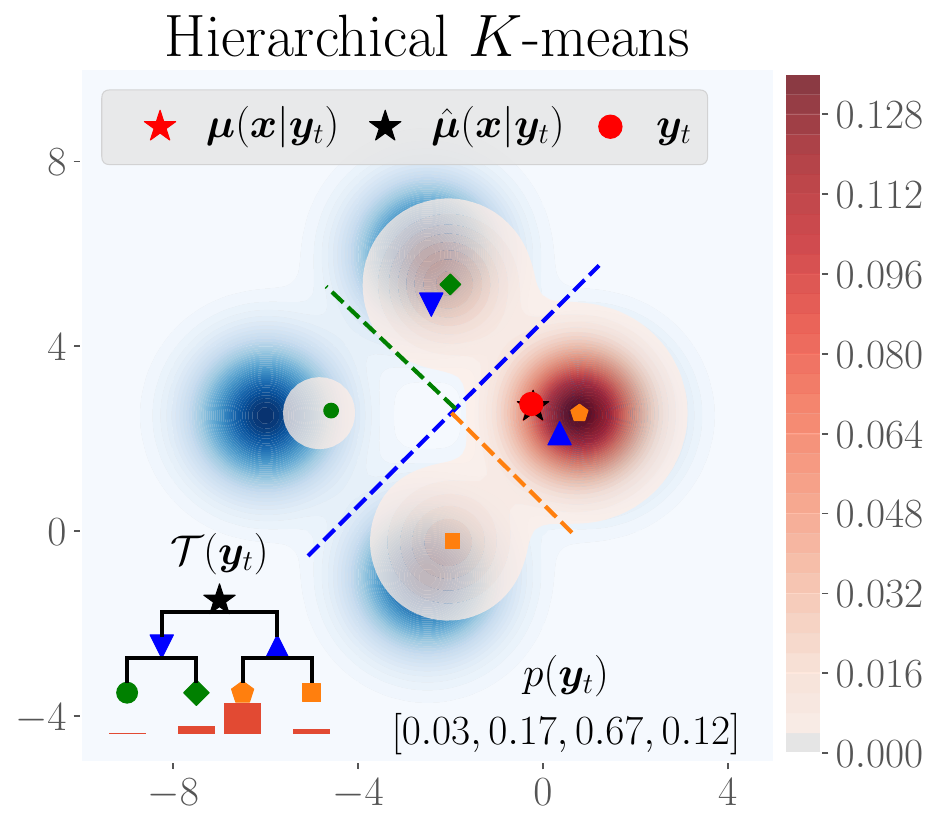}
    \caption{\hspace{-1mm}Hierarchical~$K$-means}
    \label{fig:gmm-hier-kmeans}
  \end{subfigure}%
  \begin{subfigure}{0.25\linewidth}
    \includegraphics[width=\textwidth, trim={1.4cm 1.4cm 1.8cm 1.2cm},clip]{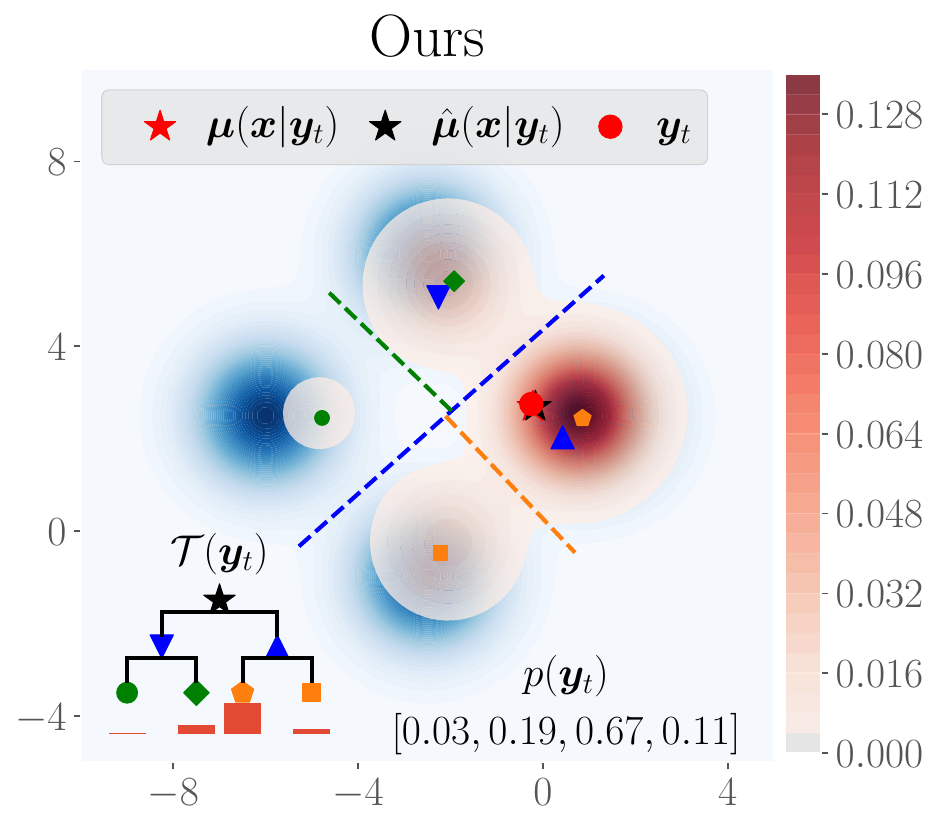}
    \caption{Ours}
    \label{fig:gmm-ours}
  \end{subfigure}%
  \caption{\textbf{2D Gaussian mixture denoising}. \protect\subref{fig:gmm-task} Underlying signal prior \textcolor{blue}{$p_{\rvx}(\vx)$} (blue heatmap), and training samples $(\vx_i,\vy_i)\sim p_{\rvx,\rvy}(\vx,\vy)$. \protect\subref{fig:gmm-flat-kmeans} $K$-means with $K=4$ applied to 10K samples $\vx_i\sim p_{\rvx|\rvy}(\vx|\vy_t)$, for a given test point $\vy_t$ (red circle). The resulting cluster centers (blue markers) partition the underlying posterior \textcolor{red}{$p_{\rvx|\rvy}(\vx|\vy_t)$} (red heatmap), resulting in cluster probabilities $p(\vy_t)$. \protect\subref{fig:gmm-hier-kmeans} Hierarchical $K$-means applied twice with $K=2$ on 10K samples $\vx_i\sim p_{\rvx|\rvy}(\vx|\vy_t)$. At depth $d=1$, the posterior is partitioned by the dashed blue line (blue triangles mark cluster centers). The resulting half spaces are subsequently halved by the dashed orange and green lines respectively. \protect\subref{fig:gmm-ours} Posterior trees (ours) with degree $K=2$ and depth $d=2$. Note that in all cases the estimated posterior mean $\hat{\vmu}(\vx|\vy_t)$ (black star) coincides with the analytical mean $\vmu(\vx|\vy_t)$ (red star), while in (c)-(d) the lowest density mode is better represented. $\mathcal{T}(\vy_t)$/$p(\vy_t)$ are drawn at the bottom of (b)-(d).}
  \label{fig:gmm-2d}
\end{figure}

Here we demonstrate our method in several settings. In all experiments except for the toy example, we used variants of the U-Net architecture \cite{ronneberger2015u}, with a custom number of output images, adjusted to accommodate $K^d$ predictions according to the desired tree layout. Moreover, as explained in \cref{subsec:arch-opt}, each model was also complemented with a small MLP for predicting leaf likelihood. Full details regarding the architectures, optimizer, and per-task settings are in \cref{subsec:app-exp-details}.

\paragraph{\textbf{Toy Example.}}~As a warm-up, we demonstrate posterior trees on a 2D denoising task. Here, $\rvx$ is sampled from a mixture of four Gaussians (arranged in a rhombus-like layout), and $\rvy$ is a noisy version of $\rvx$. 
The prior distribution $p_{\rvx}(\vx)$ and exemplar samples from $p_{\rvx,\rvy}(\vx,\vy)$ are presented in \cref{fig:gmm-task}. 
For this simple case, the posterior distribution $p_{\rvx|\rvy}(\vx|\vy)$ can be calculated analytically (see \cref{subsec:app-toy-example} for the derivation) and is also a mixture of Gaussians. Therefore, this task can serve as a sanity check enabling us to benchmark our results. 
To demonstrate our method on this toy example, we train a model $\mathcal{T}(\vy;\vtheta)$ outputting a tree of depth $d=2$ and degree $K=2$. The results are compared against an approximate ``ground truth'' in \cref{fig:gmm-hier-kmeans}.
Note that even though we have a closed-form expression for the posterior density, we still have to approximate the ``ground truth'' posterior tree by applying hierarchical K-means clustering to samples for every $\vy$.
As seen in \cref{fig:gmm-ours}, our method resulted in highly accurate posterior trees both in terms of cluster centers and cluster likelihoods. Moreover, this result was achieved while being presented only with a \emph{single} posterior sample per $\vy$ in training. In contrast, the approximate ground truth was computed with $10K$ samples.
Moreover, it is important to note that besides trivial cases, our result could not have been achieved with a ``flat'' $K$-means clustering with $K=4$ (\cref{fig:gmm-flat-kmeans}). This is because $K$-means tends to focus on high-density modes, whereas our hierarchical trees better convey rare cases of the posterior. See \cref{subsec:app-toy-example} for more details and examples.

\begin{figure}[t]
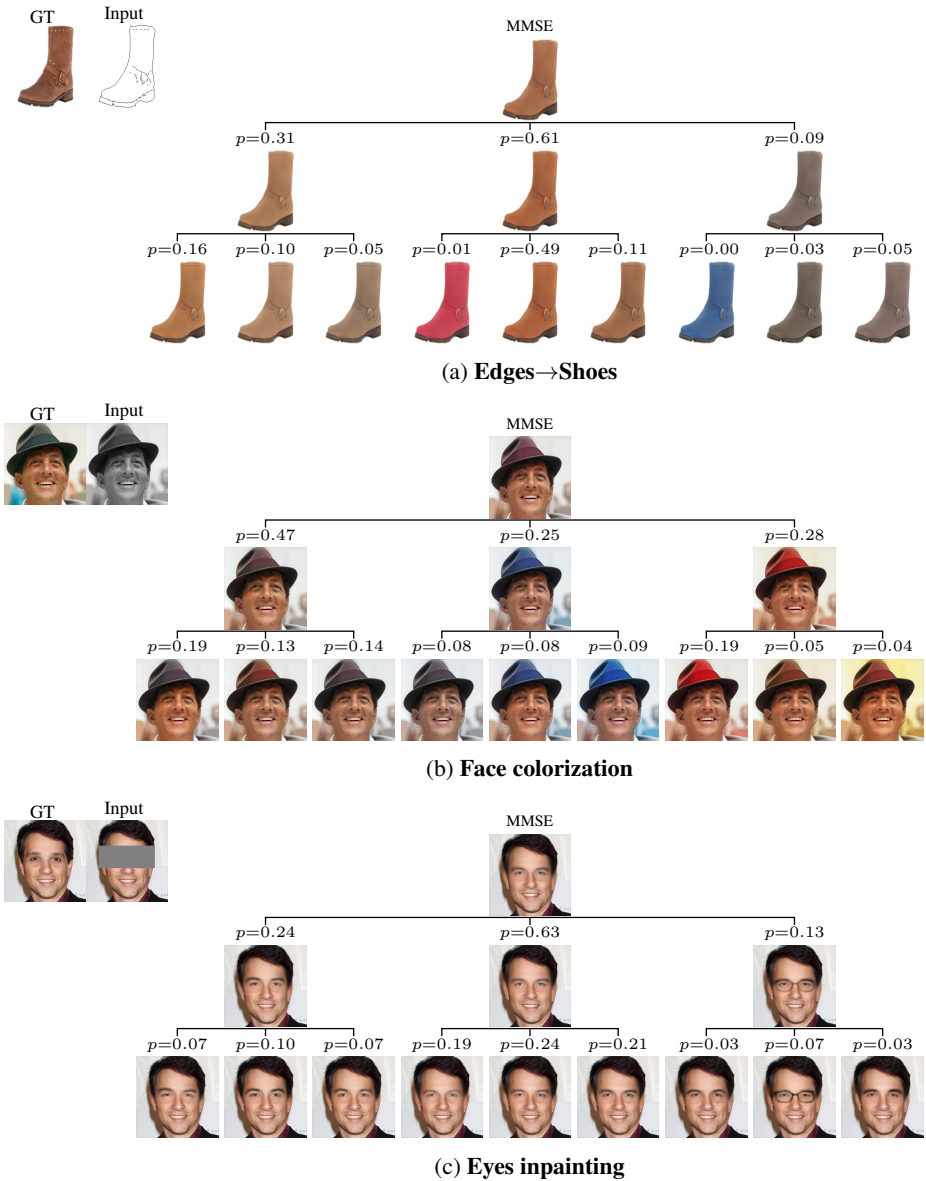

\centering
\begin{subfigure}{\linewidth}
    \input{texfigures/Edges2Shoes_main_subfig1}
    \caption{\textbf{Edges$\rightarrow$Shoes}}
    \label{fig:shoes}
    \vspace{0.2cm}
\end{subfigure}
\begin{subfigure}{\linewidth}
    \input{texfigures/CelebA_colorization_main_subfig2}
    \caption{\textbf{Face colorization}}
    \label{fig:colorization}
    \vspace{0.2cm}
\end{subfigure}
\begin{subfigure}{\linewidth}
    \input{texfigures/CelebA_eye_inpaint_main_subfig3}
    \caption{\textbf{Eyes inpainting}}
    \label{fig:eye_inpainting}
    \vspace{0.2cm}
\end{subfigure}
\caption{\textbf{Diverse applications of posterior trees}. The predicted trees represent inherent task uncertainty: \eg \protect\subref{fig:shoes} Refining the mean estimate by color, grouping similar colors, while still depicting unlikely ones (\eg the blue boot);  \protect\subref{fig:colorization} Presenting various plausible colorizations varying by hat color, skin tone, and background; and \protect\subref{fig:eye_inpainting} Exploring the diverse options of eyebrows/eyeglasses.}
\end{figure}

\paragraph{Handwritten Digits, Edges$\rightarrow$Shoes, and Human Faces.}~\Cref{fig:method} demonstrates posterior trees on inpainting the top $70\%$ of handwritten digits from the MNIST dataset. As can be seen, at depth $d=1$, the learned tree exposes the two likely modes averaged in the mean estimate $\hat{\vx}$, being either a ``7'' or a ``9''. In addition, at deeper tree levels, the different modes are further refined to reveal intricate intra-digit variations. More examples are available in \cref{subsec:app-more-results}. We also applied posterior trees to the edges-to-shoes dataset taken from pix2pix \citep{yu2014fine,isola2017image}. Here, the task is to convert an image of black and white edges to an output RGB image of a shoe. As shown in \cref{fig:shoes}, our tree is capable of representing diverse shoe colors, in an adaptive manner to the input contours. Next, we tested posterior trees on face images from the CelebA-HQ dataset, using the split from CelebA \citep{liu2015faceattributes}. \Cref{fig:colorization} demonstrates our method applied to image colorization, where the input is a grayscale image and the desired output is its RGB version. The resulting trees hierarchically refine the output predictions by background, skin tone, and hat color. We also tested our method on the task of image inpainting. \Cref{fig:teaser,fig:eye_inpainting} show the resulting trees for mouth/eye inpainting respectively. For example, the predicted tree in \cref{fig:eye_inpainting} explores the different options of eye-opening/closing, eyebrow-raising/lowering, eyeglasses, etc. This demonstrates that even shallow trees can still depict diverse reconstruction characteristics, showcasing the benefit of outputting multiple predictions.

\begin{figure}[t]
\begin{subfigure}{1cm}
    \centering
    $ \scriptstyle \text{GT}$\\
    \includegraphics[scale=0.23]{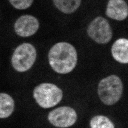}
\end{subfigure}
\begin{subfigure}{1cm}
    \centering
    $ \scriptstyle \text{Input}$\\
\includegraphics[scale=0.23]{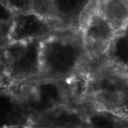}
\end{subfigure}

\vspace{-1.2cm}
\centering
\begin{forest}
  before typesetting nodes={
    for tree={
        font=\footnotesize,
      content/.wrap 2 pgfmath args={    {\footnotesize #2}\\\includegraphics[scale=0.23]{#1}}{content()}{title()},
    },
    where={isodd(n_children())}{calign=child, calign child/.wrap pgfmath arg={#1}{int((n_children()+1)/2)}}{},
  },
  forked edges,
  /tikz/every node/.append style={font=\footnotesize},
  for tree={
    parent anchor=children,
    child anchor=parent,
    align=center,
    l sep'= 1mm,
    s sep'= 2.5mm,
    fork sep'=0mm,
    inner sep = -2.5pt,
    edge={line width=0.5pt},
    font=\footnotesize,
  },
  [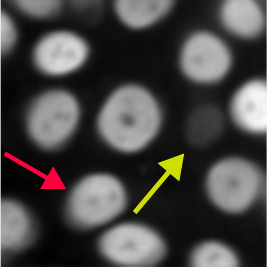, title={ {$ \scriptstyle \text{MMSE}$}}
    [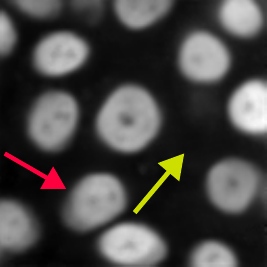, title={ $\scriptstyle p=0.57$}
      [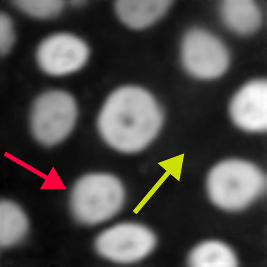, title={ $\scriptstyle p=0.20$}]
      [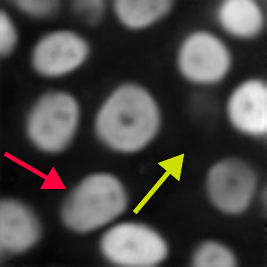, title={ $\scriptstyle p=0.36$}]
      [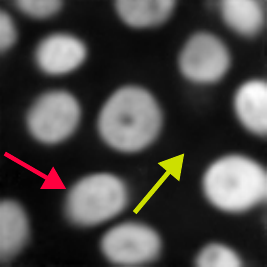, title={ $\scriptstyle p=0.01$}]
    ]
    [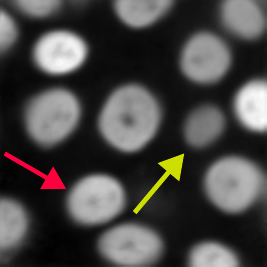, title={ $\scriptstyle p=0.37$}
      [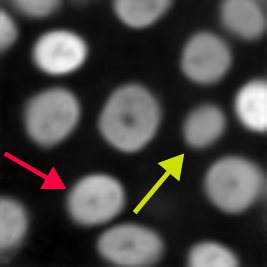, title={ $\scriptstyle p=0.27$}]
      [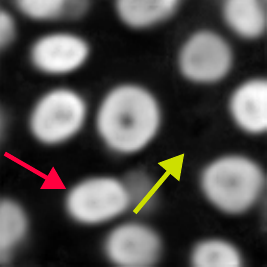, title={ $ \scriptstyle p=0.05$}]
      [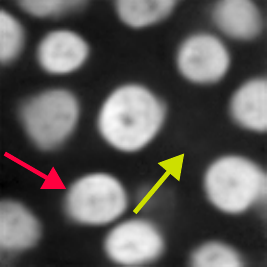, title={ $\scriptstyle p=0.05$}]
    ]
    [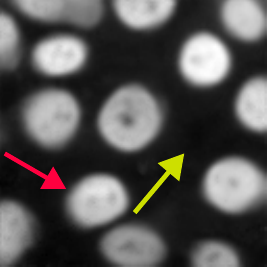, title={ $\scriptstyle p=0.06$}
      [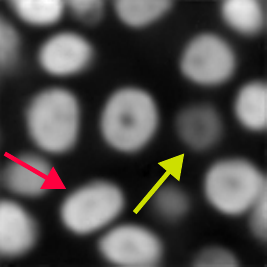, title={{ $\scriptstyle p=0.01$}}]
      [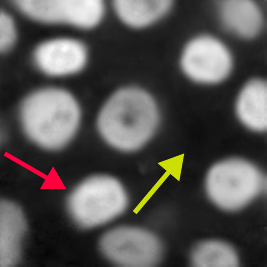, title={$\scriptstyle p=0.02$}]
      [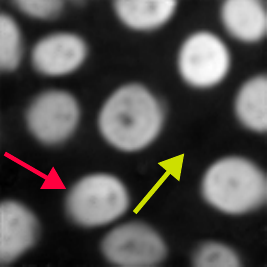, title={ $\scriptstyle p=0.03$}]
    ]
  ]
\end{forest}
\caption{\textbf{Bioimage translation.} Here we explored posterior trees for the task of translating the image of a tissue from one fluorescent dye to another. The resulting trees expose important information regarding uncertain cells (yellow/red arrows), \eg ones that do not consistently appear in all branches, and additionally explore different plausible cellular morphology consistent with the input.}
\label{fig:bioimage}
\end{figure}

\paragraph{\textbf{Bioimage Translation.}}~Transforming images from one domain to match the statistics of images from another is commonly referred to as the task of image-to-image translation \citep{isola2017image}. In the realm of bioimaging, such transformations were utilized to predict fluorescence from bright-field images \citep{ounkomol2018label}, ``virtually stain'' unstained tissue \citep{rivenson2019virtual}, and transfer the images of one fluorescent dye to appear as if they were imaged with another \citep{von2021democratising}.
The common scenario in image-to-image translation is the task being highly ill-posed with a wide range of plausible transformations satisfying the desired statistics. For general image editing/style transfer, this is a desired property adding to the artistic excitement; However, in bioimaging, this is highly problematic as the result often informs a downstream task with high stakes, and therefore output uncertainty should be communicated. Here, we applied our method to a dataset of migrating cells, imaged in a spinning-disk microscope, simultaneously with two different fluorescent dyes (one staining the nuclei and one staining actin filaments) \citep{von2021democratising}. The task was to predict the image of one fluorescent dye (nuclear stain) from another (actin stain). \Cref{fig:bioimage} demonstrates the predicted tree for a $128\times 128$ test patch. The tree conveys important information to the user exposing uncertain cells, and exploring optional cell shapes. These can for example affect downstream tasks such as cell counting and morphological cell analysis.

\subsection{Quantitative Comparisons}\label{subsec:comaprisons}

{
\renewcommand{\arraystretch}{1.1}
\aboverulesep=0ex
\belowrulesep=0ex
\begin{table}[tb]
  \caption{Comparison to the proposed baseline on 100 test images from the FFHQ dataset. Hierarchical $K$-means was applied to 100 posterior samples per test image, and compute is reported in \emph{neural function evaluations} (NFE). The NLL at the root node ($d=0$) is trivial, and therefore omitted.}
  \label{tab:ffhq-comp}
  \centering
  \small
  \begin{tabular}{c|lcccccc} 
    \toprule
     \multirow{2}{1cm}{\centering Task} & \multirow{2}{1.5cm}{\centering Method} & \multicolumn{3}{c}{Optimal PSNR ($\uparrow$)} & \multicolumn{2}{c}{NLL ($\downarrow$)} &  \multirow{2}{*}{NFE ($\downarrow$)}\\
    \cmidrule(r{.45em} ll{.50em}){3-5}
    \cmidrule(r{.45em} ll{.50em}){6-7}
    & & $d=0$ & $d=1$ & $d=2$ & $d=1$ & $d=2$ \\
    \midrule
    \multirow{3}{1cm}{Color.} & DDNM \citep{wang2023ddnm} & \bf{24.6$\pm$3.8} & {25.5$\pm$3.6} & {26.1$\pm$3.6} & \bf{0.9$\pm$0.4} & \bf{2.0$\pm$0.6} & {10000}\\
    & DDRM \citep{kawar2022denoising} & {22.7$\pm$2.7} & {24.0$\pm$3.0} & {24.8$\pm$3.0} & {1.1$\pm$0.4} & {2.0$\pm$0.5} & {2000}\\
    & Ours & \bf{24.6$\pm$4.1} & \bf{25.7$\pm$3.9} & \bf{26.4$\pm$4.0} & \bf{0.9$\pm$0.4} & \bf{2.0$\pm$0.7} & \bf{1}\\
    \midrule
    \multirow{4}{1cm}{Mouth inpaint} & DDNM \citep{wang2023ddnm} & {19.3$\pm$2.6} & {19.9$\pm$2.2} & {20.2$\pm$2.0} & {1.1$\pm$0.4} & {2.1$\pm$0.6} & {10000}\\
    & RePaint \citep{lugmayr2022repaint} & {19.8$\pm$2.7} & \bf{20.5$\pm$2.3} & \bf{20.7$\pm$2.3} & {1.0$\pm$0.3} & \bf{1.9$\pm$0.4} & {457000}\\
    & MAT \citep{li2022mat} & {19.2$\pm$2.3} & 19.4$\pm$2.3 & 19.6$\pm$2.3 & {1.2$\pm$0.3} & {2.5$\pm$0.6} & {100}\\
    & Ours & \bf{20.1$\pm$2.4} & \bf{20.5$\pm$2.3} & {20.4$\pm$2.2} & \bf{0.9$\pm$0.4} & {2.0$\pm$0.8} & \bf{1} \\
    \midrule
    \multirow{4}{1cm}{\centering Eyes inpaint} & DDNM \citep{wang2023ddnm} & {19.3$\pm$2.7} & {19.6$\pm$2.5} & {19.5$\pm$3.7} & {1.1$\pm$0.7} & {2.1$\pm$0.6} & {10000}\\
    & RePaint \citep{lugmayr2022repaint} & \bf{20.1$\pm$2.8} & \bf{20.4$\pm$2.8} & \bf{20.6$\pm$2.8} & \bf{1.0$\pm$0.4} & \bf{1.9$\pm$0.5} & {457000}\\
    & MAT \citep{li2022mat} & {19.0$\pm$3.0} & 19.2$\pm$3.0 & 19.3$\pm$3.0 & {1.2$\pm$0.3} & {2.4$\pm$0.6} & {100}\\
    & Ours & {19.6$\pm$2.7} & {19.9$\pm$2.5} & {19.8$\pm$2.4} & {1.3$\pm$0.6} & {2.5$\pm$0.7} & \bf{1} \\
    \bottomrule
  \end{tabular}
\end{table}
}

\paragraph{Baseline.} Building on the notion of clustering from \cref{subsec:multi-out}, we propose a simple baseline to benchmark our results. As shown in \cref{fig:gmm-ours}, our predicted trees are constructed out of prototypes that yield a hierarchical clustering of the posterior. In discrete settings, this is equivalent to applying Hierarchical $K$-means on samples from $p_{\rvx|\rvy}(\vx|\vy)$. Therefore, given a method to sample from the posterior (\eg using \citep{kawar2022denoising,lugmayr2022repaint,li2022mat,wang2023ddnm}), a natural baseline in our case is a two-step procedure: (i) Generate $N_s$ samples from the posterior, and (ii) Apply hierarchical $K$-means $d$ times to generate a tree of degree $K$ and depth $d$. Note that due to computational reasons, the comparison is performed over a random subset of 100 test images from FFHQ, where for each image we generate $N_s=100$ samples, and apply hierarchical $K$-means with $K=3$ and $d=2$. To ensure a fair comparison, at each clustering step, $K$-means was run 5 times, keeping only the clusters with the best objective.

\paragraph{Metrics.} To compare our predicted trees to the proposed baseline, we opted for two different metrics. The first metric is the PSNR (equivalent to MSE) between the ground truth test image and the tree nodes along the optimal path starting from the root and ending at the leaves. Intuitively, an accurate posterior clustering implies our tree nodes should maximize the PSNR, representing the ground truth test image with increasingly higher accuracy (lower MSE) as a function of (optimal) node depth. The second metric we adopt is the sample negative log-likelihood (NLL) (using the natural logarithm) of the ground truth test image under the predicted posterior partitioning. Here, an accurate tree is expected to maximize the sample log-likelihood. This metric serves the purpose of verifying the predicted probabilities, as in practice we do not have access to the ground truth posterior distribution, and estimating cluster probability based on posterior samples becomes worse as a function of depth. \Cref{tab:ffhq-comp} compares posterior trees to the proposed baseline, implemented with various state-of-the-art samplers. The results suggest that our method yields comparable results in both the PSNR along the optimal path and in sample log-likelihood while requiring a single NFE ($\approx$ 7 ms on a 1080Ti GPU). This is $10^2-10^6\times$ faster than the competition (see \cref{subsec:app-gan-sampler,subsec:app-vis-baselines} for more details and visual comparisons).

\section{Discussion and Conclusion}\label{sec:disc-conc}

We demonstrated the wide applicability of posterior trees across diverse tasks and datasets. However, our method is not free of limitations. First, the optimal hyper-parameters $K$ and $d$ are task-dependent, with no rule of thumb for determining them a-priori. Second, in this work, we focused on balanced trees with a fixed degree $K$ which might be sub-optimal for certain posteriors. Devising a strategy for an input-adaptive tree layout is an exciting direction for future research. Third, our method is limited in the number of output leaves, as we amortize the entire tree inference to a single forward pass. For predicting significantly deeper trees, an iterative inference procedure conditioning the model on the node index (in an analogous fashion to the timestep in diffusion models) is required. Finally, our method is tailored towards visualizing uncertainty and not sampling realistic-looking images. Although with increased depth our prototypes become increasingly sharper, they are nonetheless still cluster centers that average multiple plausible solutions and hence are not expected to lie on the image data manifold for shallow trees. A possible solution is to apply posterior trees in the latent space of an autoencoder such as VQ-VAE \citep{van2017neural, razavi2019generating}, however, this is beyond the scope of this paper.

To conclude, in this work, we proposed a technique to output a hierarchical quantization of the posterior in a single forward pass. We discussed key design choices underlying our approach, including bottom-up tree construction, a principled training scheme, and proper regularization techniques to prevent tree collapse. We further demonstrated the benefit of hierarchical clustering over flat trees and discussed the intuition behind it on a toy example. In our experiments, we applied our method to highly ill-posed inverse problems and showed that diverse (prototypical) reconstructions are possible with a simple training scheme exposing uncertainty. Additionally, we also proposed an appropriate baseline based on posterior sampling, and quantitatively compared our approach to several strong samplers. Our method demonstrated at least comparable results while being orders of magnitude faster. Finally, we applied our method to the challenging task of bioimage translation, demonstrating its practical relevance in a real-world application.

\clearpage
\bibliographystyle{unsrtnat}
\bibliography{references}

\begin{thebibliography}{63}
\providecommand{\natexlab}[1]{#1}
\providecommand{\url}[1]{\texttt{#1}}
\expandafter\ifx\csname urlstyle\endcsname\relax
  \providecommand{\doi}[1]{doi: #1}\else
  \providecommand{\doi}{doi: \begingroup \urlstyle{rm}\Url}\fi

\bibitem[Ounkomol et~al.(2018)Ounkomol, Seshamani, Maleckar, Collman, and Johnson]{ounkomol2018label}
Chawin Ounkomol, Sharmishtaa Seshamani, Mary~M Maleckar, Forrest Collman, and Gregory~R Johnson.
\newblock Label-free prediction of three-dimensional fluorescence images from transmitted-light microscopy.
\newblock \emph{Nature methods}, 15\penalty0 (11):\penalty0 917--920, 2018.

\bibitem[Christiansen et~al.(2018)Christiansen, Yang, Ando, Javaherian, Skibinski, Lipnick, Mount, O’neil, Shah, Lee, et~al.]{christiansen2018silico}
Eric~M Christiansen, Samuel~J Yang, D~Michael Ando, Ashkan Javaherian, Gaia Skibinski, Scott Lipnick, Elliot Mount, Alison O’neil, Kevan Shah, Alicia~K Lee, et~al.
\newblock In silico labeling: predicting fluorescent labels in unlabeled images.
\newblock \emph{Cell}, 173\penalty0 (3):\penalty0 792--803, 2018.

\bibitem[Rivenson et~al.(2019)Rivenson, Wang, Wei, de~Haan, Zhang, Wu, G{\"u}nayd{\i}n, Zuckerman, Chong, Sisk, et~al.]{rivenson2019virtual}
Yair Rivenson, Hongda Wang, Zhensong Wei, Kevin de~Haan, Yibo Zhang, Yichen Wu, Harun G{\"u}nayd{\i}n, Jonathan~E Zuckerman, Thomas Chong, Anthony~E Sisk, et~al.
\newblock Virtual histological staining of unlabelled tissue-autofluorescence images via deep learning.
\newblock \emph{Nature biomedical engineering}, 3\penalty0 (6):\penalty0 466--477, 2019.

\bibitem[Falk et~al.(2019)Falk, Mai, Bensch, {\c{C}}i{\c{c}}ek, Abdulkadir, Marrakchi, B{\"o}hm, Deubner, J{\"a}ckel, Seiwald, et~al.]{falk2019u}
Thorsten Falk, Dominic Mai, Robert Bensch, {\"O}zg{\"u}n {\c{C}}i{\c{c}}ek, Ahmed Abdulkadir, Yassine Marrakchi, Anton B{\"o}hm, Jan Deubner, Zoe J{\"a}ckel, Katharina Seiwald, et~al.
\newblock U-net: deep learning for cell counting, detection, and morphometry.
\newblock \emph{Nature methods}, 16\penalty0 (1):\penalty0 67--70, 2019.

\bibitem[Song et~al.(2021)Song, Shen, Xing, and Ermon]{song2021solving}
Yang Song, Liyue Shen, Lei Xing, and Stefano Ermon.
\newblock Solving inverse problems in medical imaging with score-based generative models.
\newblock In \emph{International Conference on Learning Representations}, 2021.

\bibitem[Ohayon et~al.(2021)Ohayon, Adrai, Vaksman, Elad, and Milanfar]{ohayon2021high}
Guy Ohayon, Theo Adrai, Gregory Vaksman, Michael Elad, and Peyman Milanfar.
\newblock High perceptual quality image denoising with a posterior sampling cgan.
\newblock In \emph{Proceedings of the IEEE/CVF International Conference on Computer Vision}, pages 1805--1813, 2021.

\bibitem[Kawar et~al.(2022)Kawar, Elad, Ermon, and Song]{kawar2022denoising}
Bahjat Kawar, Michael Elad, Stefano Ermon, and Jiaming Song.
\newblock Denoising diffusion restoration models.
\newblock In \emph{Advances in Neural Information Processing Systems}, 2022.

\bibitem[Wang et~al.(2023)Wang, Yu, and Zhang]{wang2023ddnm}
Yinhuai Wang, Jiwen Yu, and Jian Zhang.
\newblock Zero-shot image restoration using denoising diffusion null-space model.
\newblock \emph{The Eleventh International Conference on Learning Representations}, 2023.

\bibitem[Chung et~al.(2022)Chung, Kim, Mccann, Klasky, and Ye]{chung2022diffusion}
Hyungjin Chung, Jeongsol Kim, Michael~Thompson Mccann, Marc~Louis Klasky, and Jong~Chul Ye.
\newblock Diffusion posterior sampling for general noisy inverse problems.
\newblock In \emph{The Eleventh International Conference on Learning Representations}, 2022.

\bibitem[Li et~al.(2022)Li, Lin, Zhou, Qi, Wang, and Jia]{li2022mat}
Wenbo Li, Zhe Lin, Kun Zhou, Lu~Qi, Yi~Wang, and Jiaya Jia.
\newblock Mat: Mask-aware transformer for large hole image inpainting.
\newblock In \emph{Proceedings of the IEEE/CVF conference on computer vision and pattern recognition}, pages 10758--10768, 2022.

\bibitem[Lugmayr et~al.(2022)Lugmayr, Danelljan, Romero, Yu, Timofte, and Van~Gool]{lugmayr2022repaint}
Andreas Lugmayr, Martin Danelljan, Andres Romero, Fisher Yu, Radu Timofte, and Luc Van~Gool.
\newblock Repaint: Inpainting using denoising diffusion probabilistic models.
\newblock In \emph{Proceedings of the IEEE/CVF Conference on Computer Vision and Pattern Recognition}, pages 11461--11471, 2022.

\bibitem[Saharia et~al.(2022{\natexlab{a}})Saharia, Ho, Chan, Salimans, Fleet, and Norouzi]{saharia2022image}
Chitwan Saharia, Jonathan Ho, William Chan, Tim Salimans, David~J Fleet, and Mohammad Norouzi.
\newblock Image super-resolution via iterative refinement.
\newblock \emph{IEEE Transactions on Pattern Analysis and Machine Intelligence}, 2022{\natexlab{a}}.

\bibitem[Saharia et~al.(2022{\natexlab{b}})Saharia, Chan, Chang, Lee, Ho, Salimans, Fleet, and Norouzi]{saharia2022palette}
Chitwan Saharia, William Chan, Huiwen Chang, Chris Lee, Jonathan Ho, Tim Salimans, David Fleet, and Mohammad Norouzi.
\newblock Palette: Image-to-image diffusion models.
\newblock In \emph{ACM SIGGRAPH 2022 Conference Proceedings}, pages 1--10, 2022{\natexlab{b}}.

\bibitem[Bendel et~al.(2022)Bendel, Ahmad, and Schniter]{bendel2022regularized}
Matthew Bendel, Rizwan Ahmad, and Philip Schniter.
\newblock A regularized conditional gan for posterior sampling in inverse problems.
\newblock \emph{arXiv preprint arXiv:2210.13389}, 2022.

\bibitem[Cohen et~al.(2023)Cohen, Manor, Bahat, and Michaeli]{cohen2023posterior}
Noa Cohen, Hila Manor, Yuval Bahat, and Tomer Michaeli.
\newblock From posterior sampling to meaningful diversity in image restoration.
\newblock \emph{arXiv preprint arXiv:2310.16047}, 2023.

\bibitem[Sehwag et~al.(2022)Sehwag, Hazirbas, Gordo, Ozgenel, and Canton]{sehwag2022generating}
Vikash Sehwag, Caner Hazirbas, Albert Gordo, Firat Ozgenel, and Cristian Canton.
\newblock Generating high fidelity data from low-density regions using diffusion models.
\newblock In \emph{Proceedings of the IEEE/CVF Conference on Computer Vision and Pattern Recognition}, pages 11492--11501, 2022.

\bibitem[Nehme et~al.(2023)Nehme, Yair, and Michaeli]{nehme2023uncertainty}
Elias Nehme, Omer Yair, and Tomer Michaeli.
\newblock Uncertainty quantification via neural posterior principal components.
\newblock In \emph{Thirty-seventh Conference on Neural Information Processing Systems}, 2023.

\bibitem[Belhasin et~al.(2023)Belhasin, Romano, Freedman, Rivlin, and Elad]{belhasin2023principal}
Omer Belhasin, Yaniv Romano, Daniel Freedman, Ehud Rivlin, and Michael Elad.
\newblock Principal uncertainty quantification with spatial correlation for image restoration problems.
\newblock \emph{arXiv preprint arXiv:2305.10124}, 2023.

\bibitem[Manor and Michaeli(2024)]{manor2024posterior}
Hila Manor and Tomer Michaeli.
\newblock On the posterior distribution in denoising: Application to uncertainty quantification.
\newblock In \emph{The Twelfth International Conference on Learning Representations}, 2024.
\newblock URL \url{https://openreview.net/forum?id=adSGeugiuj}.

\bibitem[Yair et~al.(2023)Yair, Nehme, and Michaeli]{yair2023uncertainty}
Omer Yair, Elias Nehme, and Tomer Michaeli.
\newblock Uncertainty visualization via low-dimensional posterior projections.
\newblock \emph{arXiv preprint arXiv:2312.07804}, 2023.

\bibitem[Kendall and Gal(2017)]{kendall2017uncertainties}
Alex Kendall and Yarin Gal.
\newblock What uncertainties do we need in bayesian deep learning for computer vision?
\newblock \emph{Advances in neural information processing systems}, 30, 2017.

\bibitem[Wang et~al.(2019)Wang, Li, Aertsen, Deprest, Ourselin, and Vercauteren]{wang2019aleatoric}
Guotai Wang, Wenqi Li, Michael Aertsen, Jan Deprest, S{\'e}bastien Ourselin, and Tom Vercauteren.
\newblock Aleatoric uncertainty estimation with test-time augmentation for medical image segmentation with convolutional neural networks.
\newblock \emph{Neurocomputing}, 338:\penalty0 34--45, 2019.

\bibitem[Angelopoulos et~al.(2022)Angelopoulos, Kohli, Bates, Jordan, Malik, Alshaabi, Upadhyayula, and Romano]{angelopoulos2022image}
Anastasios~N Angelopoulos, Amit~P Kohli, Stephen Bates, Michael~I Jordan, Jitendra Malik, Thayer Alshaabi, Srigokul Upadhyayula, and Yaniv Romano.
\newblock Image-to-image regression with distribution-free uncertainty quantification and applications in imaging.
\newblock In \emph{International Conference on Machine Learning}, 2022.

\bibitem[Lugmayr et~al.(2020)Lugmayr, Danelljan, Van~Gool, and Timofte]{lugmayr2020srflow}
Andreas Lugmayr, Martin Danelljan, Luc Van~Gool, and Radu Timofte.
\newblock Srflow: Learning the super-resolution space with normalizing flow.
\newblock In \emph{ECCV}, 2020.

\bibitem[Kawar et~al.(2021)Kawar, Vaksman, and Elad]{kawar2021snips}
Bahjat Kawar, Gregory Vaksman, and Michael Elad.
\newblock Snips: Solving noisy inverse problems stochastically.
\newblock \emph{Advances in Neural Information Processing Systems}, 34:\penalty0 21757--21769, 2021.

\bibitem[Chung et~al.(2023)Chung, Lee, and Ye]{chung2023fast}
Hyungjin Chung, Suhyeon Lee, and Jong~Chul Ye.
\newblock Fast diffusion sampler for inverse problems by geometric decomposition.
\newblock \emph{arXiv preprint arXiv:2303.05754}, 2023.

\bibitem[Feng et~al.(2023)Feng, Smith, Rubinstein, Chang, Bouman, and Freeman]{feng2023score}
Berthy~T Feng, Jamie Smith, Michael Rubinstein, Huiwen Chang, Katherine~L Bouman, and William~T Freeman.
\newblock Score-based diffusion models as principled priors for inverse imaging.
\newblock \emph{arXiv preprint arXiv:2304.11751}, 2023.

\bibitem[Hyv{\"a}rinen and Dayan(2005)]{hyvarinen2005estimation}
Aapo Hyv{\"a}rinen and Peter Dayan.
\newblock Estimation of non-normalized statistical models by score matching.
\newblock \emph{Journal of Machine Learning Research}, 6\penalty0 (4), 2005.

\bibitem[Sohl-Dickstein et~al.(2015)Sohl-Dickstein, Weiss, Maheswaranathan, and Ganguli]{sohl2015deep}
Jascha Sohl-Dickstein, Eric Weiss, Niru Maheswaranathan, and Surya Ganguli.
\newblock Deep unsupervised learning using nonequilibrium thermodynamics.
\newblock In \emph{International conference on machine learning}, pages 2256--2265. PMLR, 2015.

\bibitem[Song and Ermon(2019)]{song2019generative}
Yang Song and Stefano Ermon.
\newblock Generative modeling by estimating gradients of the data distribution.
\newblock \emph{Advances in neural information processing systems}, 32, 2019.

\bibitem[Ho et~al.(2020)Ho, Jain, and Abbeel]{ho2020denoising}
Jonathan Ho, Ajay Jain, and Pieter Abbeel.
\newblock Denoising diffusion probabilistic models.
\newblock \emph{Advances in neural information processing systems}, 33:\penalty0 6840--6851, 2020.

\bibitem[Salimans and Ho(2022)]{salimans2022progressive}
Tim Salimans and Jonathan Ho.
\newblock Progressive distillation for fast sampling of diffusion models.
\newblock \emph{arXiv preprint arXiv:2202.00512}, 2022.

\bibitem[Luhman and Luhman(2021)]{luhman2021knowledge}
Eric Luhman and Troy Luhman.
\newblock Knowledge distillation in iterative generative models for improved sampling speed.
\newblock \emph{arXiv preprint arXiv:2101.02388}, 2021.

\bibitem[Meng et~al.(2022)Meng, Gao, Kingma, Ermon, Ho, and Salimans]{meng2022on}
Chenlin Meng, Ruiqi Gao, Diederik~P Kingma, Stefano Ermon, Jonathan Ho, and Tim Salimans.
\newblock On distillation of guided diffusion models.
\newblock In \emph{NeurIPS 2022 Workshop on Score-Based Methods}, 2022.

\bibitem[Song et~al.(2023)Song, Dhariwal, Chen, and Sutskever]{song2023consistency}
Yang Song, Prafulla Dhariwal, Mark Chen, and Ilya Sutskever.
\newblock Consistency models.
\newblock \emph{arXiv preprint arXiv:2303.01469}, 2023.

\bibitem[Dorta et~al.(2018)Dorta, Vicente, Agapito, Campbell, and Simpson]{dorta2018structured}
Garoe Dorta, Sara Vicente, Lourdes Agapito, Neill~DF Campbell, and Ivor Simpson.
\newblock Structured uncertainty prediction networks.
\newblock In \emph{Proceedings of the IEEE conference on computer vision and pattern recognition}, pages 5477--5485, 2018.

\bibitem[Monteiro et~al.(2020)Monteiro, Le~Folgoc, Coelho~de Castro, Pawlowski, Marques, Kamnitsas, van~der Wilk, and Glocker]{monteiro2020stochastic}
Miguel Monteiro, Lo{\"\i}c Le~Folgoc, Daniel Coelho~de Castro, Nick Pawlowski, Bernardo Marques, Konstantinos Kamnitsas, Mark van~der Wilk, and Ben Glocker.
\newblock Stochastic segmentation networks: Modelling spatially correlated aleatoric uncertainty.
\newblock \emph{Advances in neural information processing systems}, 33:\penalty0 12756--12767, 2020.

\bibitem[Meng et~al.(2021)Meng, Song, Li, and Ermon]{meng2021estimating}
Chenlin Meng, Yang Song, Wenzhe Li, and Stefano Ermon.
\newblock Estimating high order gradients of the data distribution by denoising.
\newblock \emph{Advances in Neural Information Processing Systems}, 34:\penalty0 25359--25369, 2021.

\bibitem[Nussbaum et~al.(2022)Nussbaum, Gawlikowski, and Niebling]{nussbaum2022structuring}
Frank Nussbaum, Jakob Gawlikowski, and Julia Niebling.
\newblock Structuring uncertainty for fine-grained sampling in stochastic segmentation networks.
\newblock \emph{Advances in Neural Information Processing Systems}, 35:\penalty0 27678--27691, 2022.

\bibitem[Guzman-Rivera et~al.(2012)Guzman-Rivera, Batra, and Kohli]{guzman2012multiple}
Abner Guzman-Rivera, Dhruv Batra, and Pushmeet Kohli.
\newblock Multiple choice learning: Learning to produce multiple structured outputs.
\newblock \emph{Advances in neural information processing systems}, 25, 2012.

\bibitem[Guzman-Rivera et~al.(2014)Guzman-Rivera, Kohli, Batra, and Rutenbar]{guzman2014efficiently}
Abner Guzman-Rivera, Pushmeet Kohli, Dhruv Batra, and Rob Rutenbar.
\newblock Efficiently enforcing diversity in multi-output structured prediction.
\newblock In \emph{Artificial Intelligence and Statistics}, pages 284--292. PMLR, 2014.

\bibitem[Lee et~al.(2015)Lee, Purushwalkam, Cogswell, Crandall, and Batra]{lee2015m}
Stefan Lee, Senthil Purushwalkam, Michael Cogswell, David Crandall, and Dhruv Batra.
\newblock Why m heads are better than one: Training a diverse ensemble of deep networks.
\newblock \emph{arXiv preprint arXiv:1511.06314}, 2015.

\bibitem[Lee et~al.(2016)Lee, Purushwalkam Shiva~Prakash, Cogswell, Ranjan, Crandall, and Batra]{lee2016stochastic}
Stefan Lee, Senthil Purushwalkam Shiva~Prakash, Michael Cogswell, Viresh Ranjan, David Crandall, and Dhruv Batra.
\newblock Stochastic multiple choice learning for training diverse deep ensembles.
\newblock \emph{Advances in Neural Information Processing Systems}, 29, 2016.

\bibitem[Lee et~al.(2017)Lee, Hwang, Park, and Shin]{lee2017confident}
Kimin Lee, Changho Hwang, KyoungSoo Park, and Jinwoo Shin.
\newblock Confident multiple choice learning.
\newblock In \emph{International Conference on Machine Learning}, pages 2014--2023. PMLR, 2017.

\bibitem[Rupprecht et~al.(2017)Rupprecht, Laina, DiPietro, Baust, Tombari, Navab, and Hager]{rupprecht2017learning}
Christian Rupprecht, Iro Laina, Robert DiPietro, Maximilian Baust, Federico Tombari, Nassir Navab, and Gregory~D Hager.
\newblock Learning in an uncertain world: Representing ambiguity through multiple hypotheses.
\newblock In \emph{Proceedings of the IEEE international conference on computer vision}, pages 3591--3600, 2017.

\bibitem[Firman et~al.(2018)Firman, Campbell, Agapito, and Brostow]{firman2018diversenet}
Michael Firman, Neill~DF Campbell, Lourdes Agapito, and Gabriel~J Brostow.
\newblock Diversenet: When one right answer is not enough.
\newblock In \emph{Proceedings of the IEEE Conference on Computer Vision and Pattern Recognition}, pages 5598--5607, 2018.

\bibitem[Ilg et~al.(2018)Ilg, Cicek, Galesso, Klein, Makansi, Hutter, and Brox]{ilg2018uncertainty}
Eddy Ilg, Ozgun Cicek, Silvio Galesso, Aaron Klein, Osama Makansi, Frank Hutter, and Thomas Brox.
\newblock Uncertainty estimates and multi-hypotheses networks for optical flow.
\newblock In \emph{Proceedings of the European Conference on Computer Vision (ECCV)}, pages 652--667, 2018.

\bibitem[Tian et~al.(2019)Tian, Xu, Zhou, and Guan]{tian2019versatile}
Kai Tian, Yi~Xu, Shuigeng Zhou, and Jihong Guan.
\newblock Versatile multiple choice learning and its application to vision computing.
\newblock In \emph{Proceedings of the IEEE/CVF Conference on Computer Vision and Pattern Recognition}, pages 6349--6357, 2019.

\bibitem[Stein and Oshman(2022)]{stein2022double}
Nitai Stein and Yaakov Oshman.
\newblock Double-opportunity estimation via altruism.
\newblock \emph{IEEE Transactions on Aerospace and Electronic Systems}, 2022.

\bibitem[Letzelter et~al.(2023)Letzelter, Fontaine, P{\'e}rez, Richard, Essid, and Chen]{letzelter2023resilient}
Victor Letzelter, Mathieu Fontaine, Patrick P{\'e}rez, Gael Richard, Slim Essid, and Micka{\"e}l Chen.
\newblock Resilient multiple choice learning: A learned scoring scheme with application to audio scene analysis.
\newblock In \emph{Thirty-seventh Conference on Neural Information Processing Systems}, 2023.

\bibitem[Ronneberger et~al.(2015)Ronneberger, Fischer, and Brox]{ronneberger2015u}
Olaf Ronneberger, Philipp Fischer, and Thomas Brox.
\newblock U-net: Convolutional networks for biomedical image segmentation.
\newblock In \emph{Medical Image Computing and Computer-Assisted Intervention--MICCAI 2015: 18th International Conference, Munich, Germany, October 5-9, 2015, Proceedings, Part III 18}, pages 234--241. Springer, 2015.

\bibitem[Krizhevsky et~al.(2012)Krizhevsky, Sutskever, and Hinton]{krizhevsky2012imagenet}
Alex Krizhevsky, Ilya Sutskever, and Geoffrey~E Hinton.
\newblock Imagenet classification with deep convolutional neural networks.
\newblock \emph{Advances in neural information processing systems}, 25, 2012.

\bibitem[Brodie et~al.(2018)Brodie, Tensmeyer, Ackerman, and Martinez]{brodie2018alpha}
Mike Brodie, Chris Tensmeyer, Wes Ackerman, and Tony Martinez.
\newblock Alpha model domination in multiple choice learning.
\newblock In \emph{2018 17th IEEE International Conference on Machine Learning and Applications (ICMLA)}, pages 879--884. IEEE, 2018.

\bibitem[Makansi et~al.(2019)Makansi, Ilg, Cicek, and Brox]{makansi2019overcoming}
Osama Makansi, Eddy Ilg, Ozgun Cicek, and Thomas Brox.
\newblock Overcoming limitations of mixture density networks: A sampling and fitting framework for multimodal future prediction.
\newblock In \emph{Proceedings of the IEEE/CVF Conference on Computer Vision and Pattern Recognition}, pages 7144--7153, 2019.

\bibitem[Kingma and Ba(2015)]{kingma2015adam}
Diederik~P Kingma and Jimmy Ba.
\newblock Adam: A method for stochastic optimization.
\newblock \emph{International Conference on Learning Representations}, 3, 2015.

\bibitem[Yu and Grauman(2014)]{yu2014fine}
Aron Yu and Kristen Grauman.
\newblock Fine-grained visual comparisons with local learning.
\newblock In \emph{Proceedings of the IEEE conference on computer vision and pattern recognition}, pages 192--199, 2014.

\bibitem[Isola et~al.(2017)Isola, Zhu, Zhou, and Efros]{isola2017image}
Phillip Isola, Jun-Yan Zhu, Tinghui Zhou, and Alexei~A Efros.
\newblock Image-to-image translation with conditional adversarial networks.
\newblock In \emph{Proceedings of the IEEE conference on computer vision and pattern recognition}, pages 1125--1134, 2017.

\bibitem[Liu et~al.(2015)Liu, Luo, Wang, and Tang]{liu2015faceattributes}
Ziwei Liu, Ping Luo, Xiaogang Wang, and Xiaoou Tang.
\newblock Deep learning face attributes in the wild.
\newblock In \emph{Proceedings of International Conference on Computer Vision (ICCV)}, December 2015.

\bibitem[von Chamier et~al.(2021)von Chamier, Laine, Jukkala, Spahn, Krentzel, Nehme, Lerche, Hern{\'a}ndez-P{\'e}rez, Mattila, Karinou, et~al.]{von2021democratising}
Lucas von Chamier, Romain~F Laine, Johanna Jukkala, Christoph Spahn, Daniel Krentzel, Elias Nehme, Martina Lerche, Sara Hern{\'a}ndez-P{\'e}rez, Pieta~K Mattila, Eleni Karinou, et~al.
\newblock Democratising deep learning for microscopy with zerocostdl4mic.
\newblock \emph{Nature communications}, 12\penalty0 (1):\penalty0 2276, 2021.

\bibitem[Van Den~Oord et~al.(2017)Van Den~Oord, Vinyals, et~al.]{van2017neural}
Aaron Van Den~Oord, Oriol Vinyals, et~al.
\newblock Neural discrete representation learning.
\newblock \emph{Advances in neural information processing systems}, 30, 2017.

\bibitem[Razavi et~al.(2019)Razavi, Van~den Oord, and Vinyals]{razavi2019generating}
Ali Razavi, Aaron Van~den Oord, and Oriol Vinyals.
\newblock Generating diverse high-fidelity images with vq-vae-2.
\newblock \emph{Advances in neural information processing systems}, 32, 2019.

\bibitem[Diamond and Boyd(2016)]{diamond2016cvxpy}
Steven Diamond and Stephen Boyd.
\newblock {CVXPY}: {A} {P}ython-embedded modeling language for convex optimization.
\newblock \emph{Journal of Machine Learning Research}, 17\penalty0 (83):\penalty0 1--5, 2016.

\bibitem[Choi et~al.(2020)Choi, Uh, Yoo, and Ha]{choi2020stargan}
Yunjey Choi, Youngjung Uh, Jaejun Yoo, and Jung-Woo Ha.
\newblock Stargan v2: Diverse image synthesis for multiple domains.
\newblock In \emph{Proceedings of the IEEE/CVF conference on computer vision and pattern recognition}, pages 8188--8197, 2020.

\end{thebibliography}

\clearpage
\appendix
\beginsupplement
\section*{Appendices}\label{app}
\section{Experimental Details}\label{subsec:app-exp-details}
\subsection{Architectures}\label{subsec:app-arch}
As mentioned in \cref{subsec:arch-opt}, in our experiments we adopted the U-Net \citep{ronneberger2015u} architecture. Our architecture consisted of 4 downsampling/upsampling blocks. Downsampling by $2\times$ was performed using average pooling, and upsampling by $2\times$ was implemented with a nearest-neighbor interpolation. In both cases, the feature maps at the updated spatial resolution were processed by 2 convolution blocks consisting of 2D convolution, group normalization, and LeakyReLU activation with a negative slope of 0.2. The number of features at each spatial resolution was adapted to the predicted tree layout, \ie for predicting a tree of degree $K$ and depth $d$, the initial number of channels before downsampling was set to $c_{\text{init}}K^d$, where $K^d$ is the number of output leaves and $c_{\text{init}}\in\{4,8\}$. At each successive encoder block, the number of channels was doubled, reaching $16c_{\text{init}}K^d$ at the bottleneck. For example, assuming a tree with degree $K=3$ and depth $d=2$, and setting $c_{\text{init}}=4$, the number of channels per level in the encoder is given by $[36, 72, 144, 288, 576]$.

\begin{figure}[ht]
\begin{subfigure}{\linewidth}
\begin{subfigure}{1cm}
    \centering
    $ \scriptstyle \text{GT}$\\
    \includegraphics[scale=0.12]{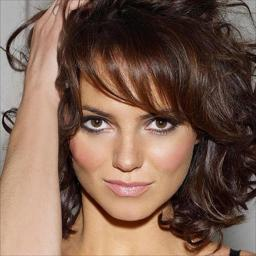}
\end{subfigure}
\begin{subfigure}{1cm}
    \centering
    $ \scriptstyle \text{Input}$\\
\includegraphics[scale=0.12]{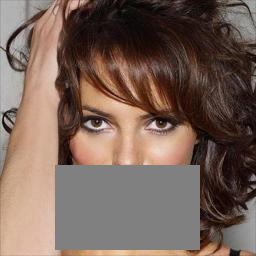}
\end{subfigure}

\vspace{-1.2cm}
\begin{nscenter}
\begin{forest}
  before typesetting nodes={
    for tree={
        font=\footnotesize,
      content/.wrap 2 pgfmath args={    {\footnotesize #2}\\\includegraphics[scale=0.12]{#1}}{content()}{title()},
    },
    where={isodd(n_children())}{calign=child, calign child/.wrap pgfmath arg={#1}{int((n_children()+1)/2)}}{},
  },
  forked edges,
  /tikz/every node/.append style={font=\footnotesize},
  for tree={
    parent anchor=children,
    child anchor=parent,
    align=center,
    l sep'= 1mm,
    s sep'= 2.5mm,
    fork sep'=0mm,
    inner sep = -2.5pt,
    edge={line width=0.5pt},
    font=\footnotesize,
  },
  [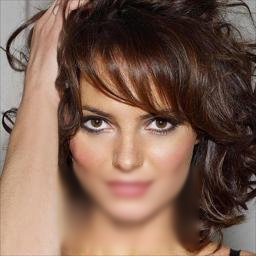, title={ {$ \scriptstyle \text{MMSE}$}}
    [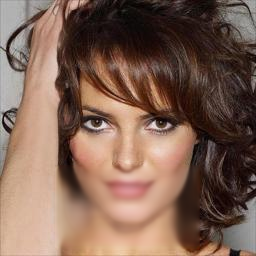, title={ $\scriptstyle p=0.23$}
      [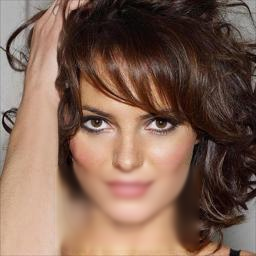, title={ $\scriptstyle p=0.14$}]
      [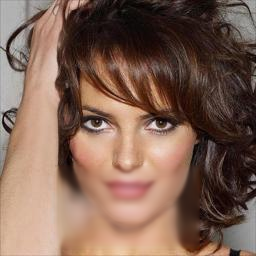, title={ $\scriptstyle p=0.07$}]
      [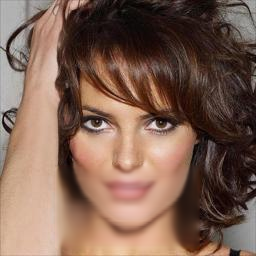, title={ $\scriptstyle p=0.03$}]
    ]
    [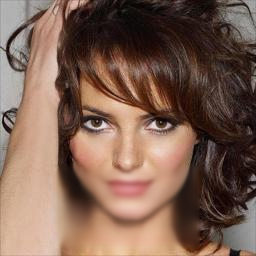, title={ $\scriptstyle p=0.50$}
      [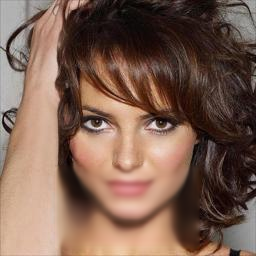, title={ $\scriptstyle p=0.11$}]
      [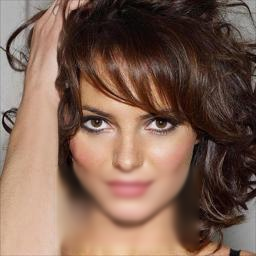, title={ $ \scriptstyle p=0.08$}]
      [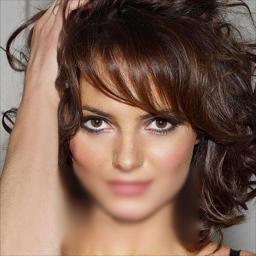, title={ $\scriptstyle p=0.31$}]
    ]
    [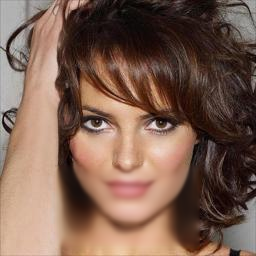, title={ $\scriptstyle p=0.26$}
      [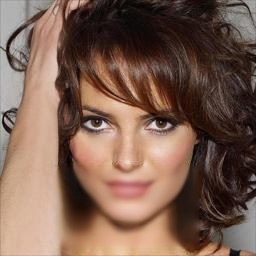, title={{ $\scriptstyle p=0.14$}}]
      [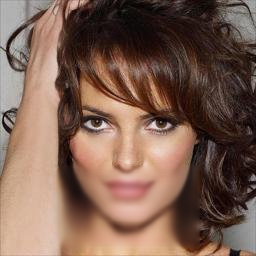, title={$\scriptstyle p=0.06$}]
      [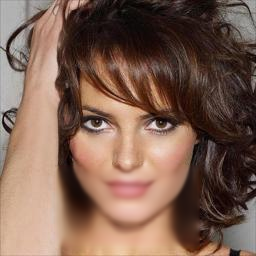, title={ $\scriptstyle p=0.07$}]
    ]
  ]
\end{forest}
\end{nscenter}
\centering
\caption{Fully shared}
\label{fig:app-full-weight-sharing}\vspace{0.2cm}
\end{subfigure}

\input{texfigures/Encoder_shared_subfig_appendix}
\caption{\textbf{Leaf weight sharing strategy.} \protect\subref{fig:app-full-weight-sharing} Fully shared architecture, with all leaves predicted jointly. \protect\subref{fig:app-encoder-weight-sharing} Leaves only share encoder (see \cref{fig:app-architecture}).}
\label{fig:app-leaf-weight-sharing}
\end{figure}

Similarly, in the decoder, the number of channels was halved at each upsampling step in a symmetric fashion. However, as mentioned in \cref{subsec:arch-opt}, in some tasks (\eg image inpainting), we noticed that fully sharing parameters between all leaves led to reduced diversity between predictions (see \cref{fig:app-leaf-weight-sharing}). On the other hand, learning a separate U-Net for each prediction leaf is computationally intensive, and highly inefficient as it is expected that the initial feature extraction stage for predicting the different leaves would be similar. Therefore, as a middle ground between these two extremes, we opted for an architecture that shares the encoder between the different leaves, while having disjoint decoders, each with a dedicated set of weights $\{\vvarphi_k\}_{k=1}^{K^d}$, learned separately from others. As for the skip connections from the encoder, the concatenated feature maps are interleaved equally per level such that each prediction has an equal share. 

\begin{figure}[h!]
\centering
\includegraphics[width=1.0\textwidth]{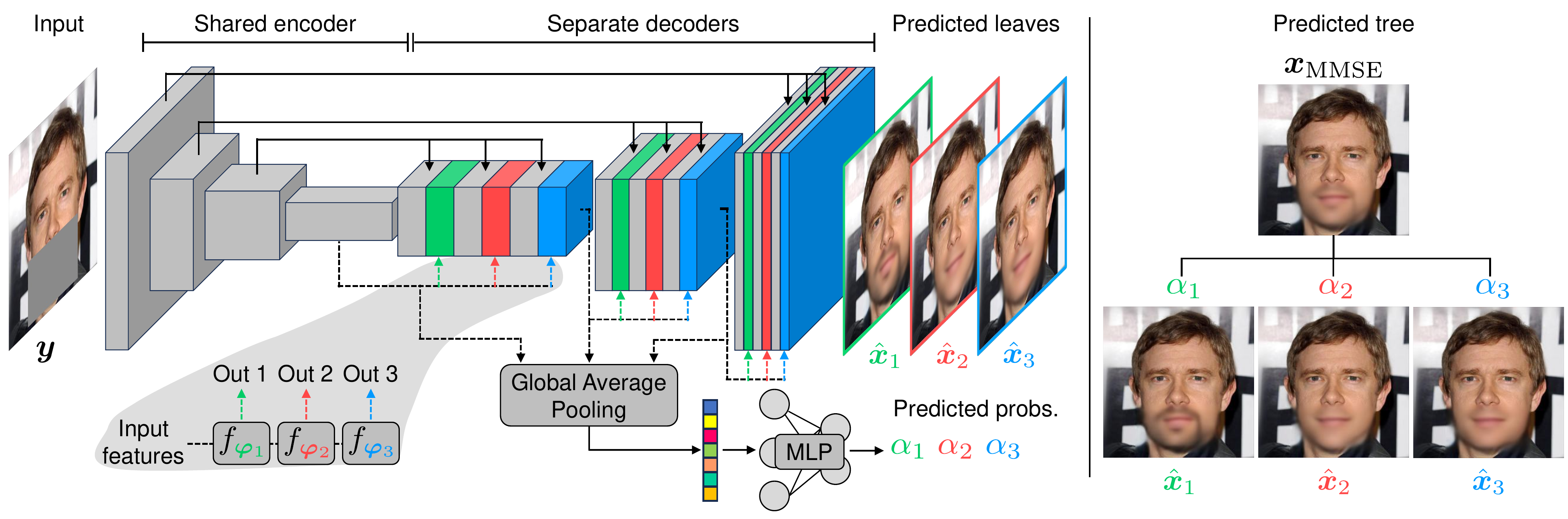}
\caption{\textbf{Model architecture}. Our model receives a degraded image $\vy$ and predicts the bottom $K^d$ leaves and their probabilities $\{\hat{\vx}_{k_1,\dots,k_d},\alpha_{k_1,\dots,k_d}\}_{k_1,\dots,k_d=1}^K$ (illustrated here for $K=3$ and $d=1$). The encoder is shared between all leaves, however, in the decoding stage, each leaf has a separate decoder (\eg with parameters \textcolor{green}{$\vvarphi_1$},\textcolor{red}{$\vvarphi_2$},\textcolor{blue}{$\vvarphi_3$}) to enable diverse outputs. To predict leaf probability, all features from the decoder are passed through a global average pooling layer, and the resulting concatenated vector is passed to a lightweight MLP with a softmax layer at the output. Afterward, as explained earlier, the output tree is constructed iteratively from the bottom up (right).}
\label{fig:app-architecture}
\end{figure}

In addition to the output images, our architecture also predicts $K^d$ scalars, which are the probabilities of the different leaves. This is achieved by global average pooling of all feature maps from the decoder, and feeding their concatenation to an additional lightweight MLP. This MLP has four linear layers with dimensions $[d_f,256,64,K^d]$, where $d_f$ is the dimension of the concatenated pooled features from the decoder. Each linear layer is followed by a 1D batch normalization and SiLU non-linearity, and the output is passed through a $\softmax$ layer to produce a valid probability vector. \Cref{fig:app-architecture} summarizes our architecture. 

\subsection{Per-task Details}\label{subsec:app-task-details}

In all tasks, we only learned the required residual from the input to produce the predictions. For MNIST experiments, the images were padded to $32\times32$ to enable proper downsampling in the encoder. For Edges$\rightarrow$Shoes experiments, the images were kept the same as in the pix2pix \citep{isola2017image} paper, with a resolution of $256\times256$. For CelebA-HQ experiments, the images were resized to $256\times256$. Finally, for the bioimage translation dataset, we trained on $128\times128$ patches cropped from the full $1024\times1024$ images, as cell information tends to be local.

\subsection{Optimization}\label{subsec:app-optimizer}

We used the Adam optimizer \citep{kingma2015adam} with $\beta_1=0.9,\beta_2=0.999$ for all experiments. For the U-Net predicting the output leaves, we used an initial learning rate of 0.001. For the MLP predicting leaf probability, however, we found it beneficial to use a smaller initial learning rate of 0.0002. This is because at the initial phase of training, the predicted leaves are still not converged, and hence the MLP can easily ``classify'' which leaf is the most likely one. This collapses the predicted probabilities to a sparse vector, leading to leaves with zero probability. Hence, to avoid this instability, we used a lower learning rate for the probabilities such that the leaves are first allowed to converge, leading to the desired learning dynamics. For both components, the learning rate was dropped by a factor of 10 if the validation loss stagnated for more than 10 epochs, and the minimum learning rate was set to \(5\cdot 10^{-6}\). 
We used a batch size of 32 for 70 epochs for all tasks. This resulted in training times of $\approx$40 minutes, 10 hours, 5 hours, and 7 hours for MNIST, Edges$\rightarrow$Shoes, CelebA-HQ, and the bioimage datasets respectively.

\clearpage
\section{Role of $\varepsilon_t$ From \cref{eq:epsilon-reg}}\label{subsec:app-epsilont}

\begin{figure}[h]
\begin{subfigure}{0.5\linewidth}
    \includegraphics[width=\linewidth]{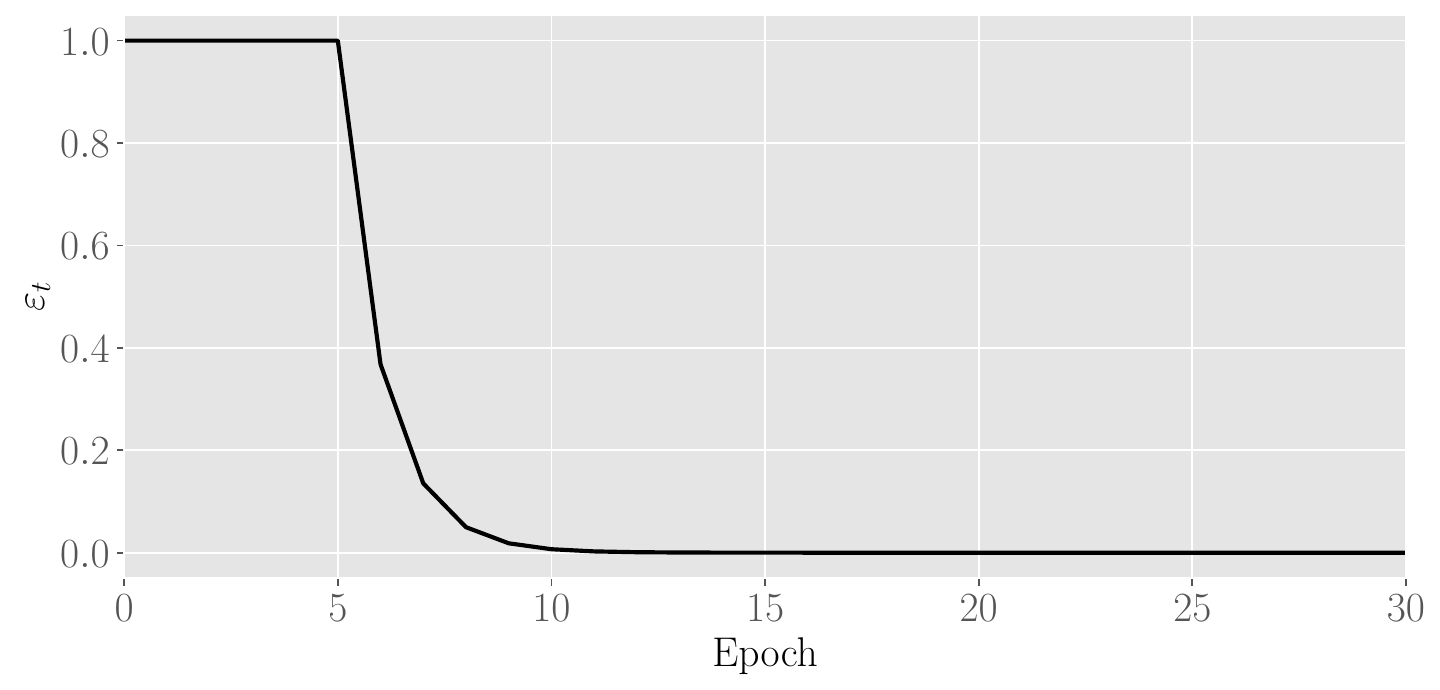}
    \caption{$\varepsilon_t$}
    \label{fig:app-epsilont-vs-epoch}
\end{subfigure}%
\begin{subfigure}{0.5\linewidth}
    \includegraphics[width=\linewidth]{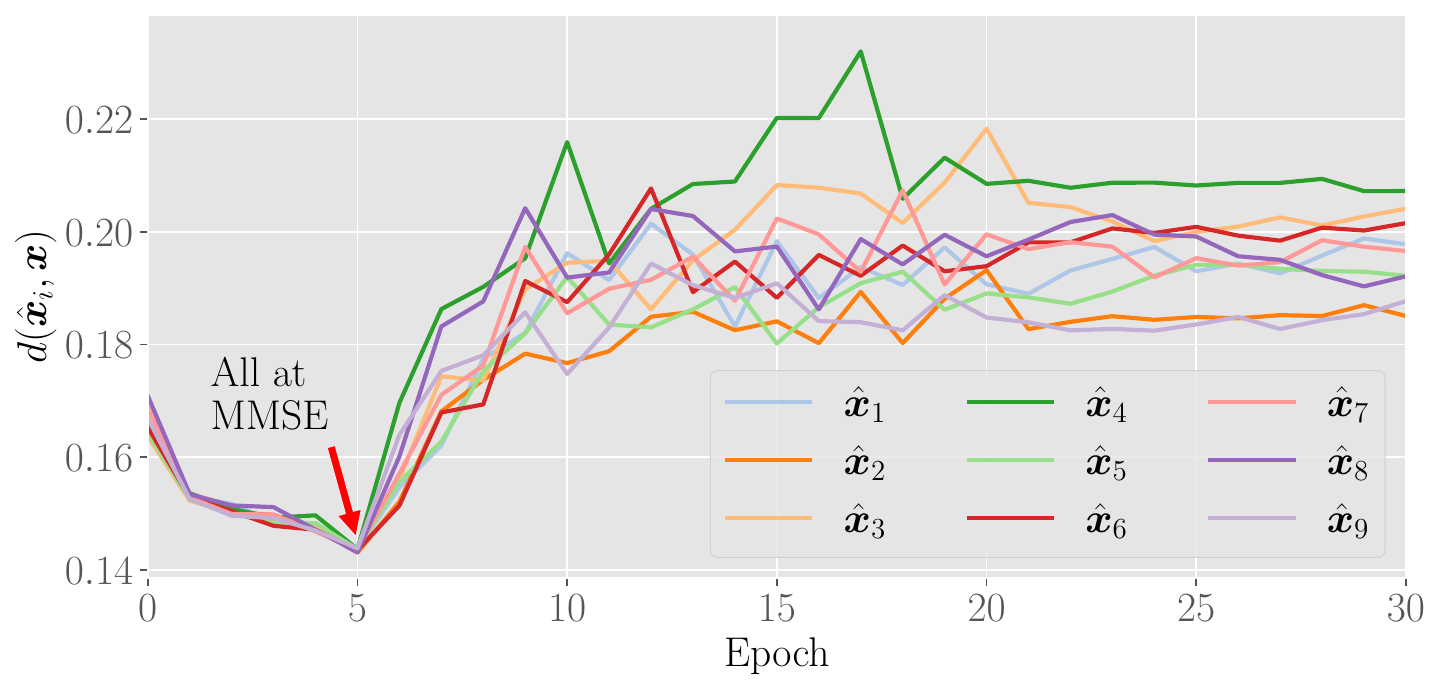}
    \caption{Distance of leaves $\{\hat{\vx}_i\}$ from the GT $\vx$}
    \label{fig:app-val-dists-vs-epoch}
\end{subfigure}

\centering
\begin{subfigure}{\linewidth}
    \input{texfigures/epsilon_t_before_subfig}
    \caption{Predicted tree at epoch $t=5$}
    \label{fig:app-tree-at-epoch5}
    \vspace{0.2cm}
\end{subfigure}

\begin{subfigure}{\linewidth}
    \input{texfigures/epsilon_t_after_subfig}
    \caption{Predicted tree at convergence (epoch $t=22$)}
    \label{fig:app-tree-at-converge}
\end{subfigure}

\caption{\textbf{Role of $\varepsilon_t$}. \protect\subref{fig:app-epsilont-vs-epoch} $\varepsilon_t$. \protect\subref{fig:app-val-dists-vs-epoch} Distances of leaf predictions $\{\hat{\vx}_i\}$ from the GT $\vx$ throughout epochs. For the first $t_0=5$ epochs all predictions are brought to the vicinity of the MMSE estimator. Afterward, $\varepsilon_t$ is decayed and each of the leaves is free to specialize in a subset of the posterior, leading to higher distances on average for the entire test set. \protect\subref{fig:app-tree-at-epoch5} Predicted tree at epoch $t=t_0=5$ where all leaves are near the MMSE estimator. \protect\subref{fig:app-tree-at-converge} Predicted tree at convergence (epoch $t=22$) where each leaf specializes in a different posterior mode (\eg a ``3'' or a ``5'').}
\label{fig:app-role-of-epsilont}
\end{figure}

As explained in \cref{subsec:prevent-collapse}, the convergence of MCL is strongly affected by initialization. This is because leaves that are better initialized are more likely to be chosen in the oracle loss, and hence will dominate the remaining leaves in training. This results in output leaves that practically do not train, and therefore give meaningless results at test time. To remedy this, we scale the gradients of non-performing predictions with a constant $\varepsilon$ that is annealed during training according to \cref{eq:epsilon-reg}. For the first $t_0=5$, our training degenerates to standard MSE minimization, bringing all leaves to a reasonable starting point near the posterior mean (\cref{fig:app-tree-at-epoch5}). This results in all prediction leaves having roughly the same initialization, with a similar number of associated training samples. Afterward, $\varepsilon$ is gradually decayed, and the leaves start to specialize in their respective posterior mode, converging to the desired MCL behavior (\cref{fig:app-tree-at-converge}).

\clearpage

\section{Weighted Sampler}\label{subsec:app-sampler}

\begin{figure}[t]
\begin{subfigure}{0.5\linewidth}
    \includegraphics[width=\linewidth]{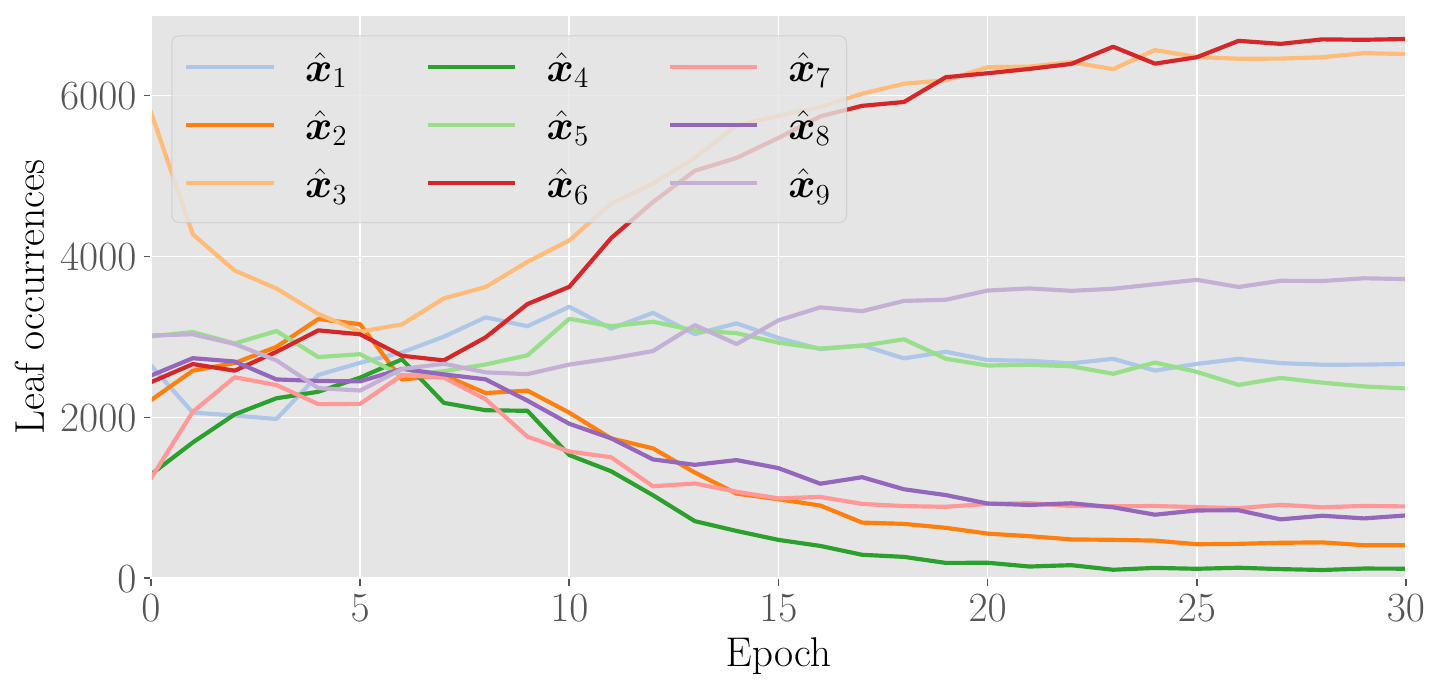}
    \caption{Standard sampling}
    \label{fig:app-occurrences-before-sampler}
\end{subfigure}%
\begin{subfigure}{0.5\linewidth}
    \includegraphics[width=\linewidth]{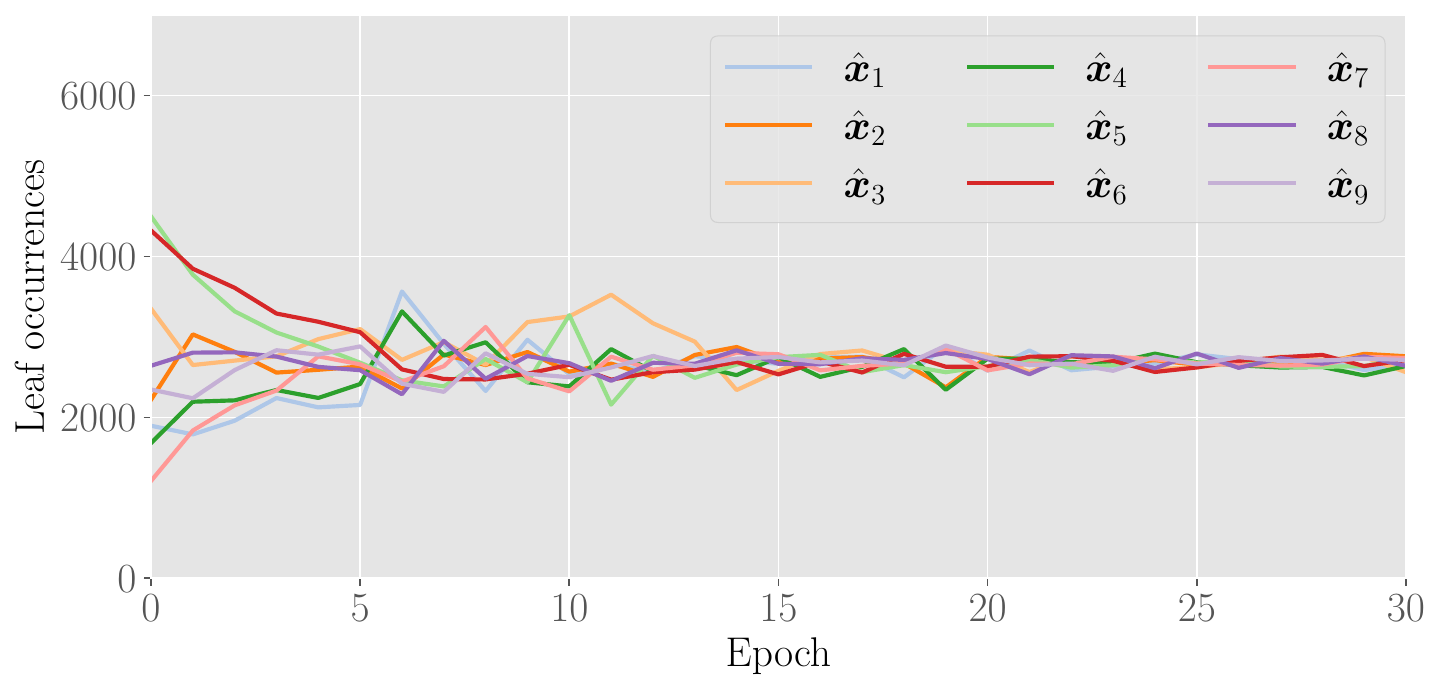}
    \caption{Weighted sampling}
    \label{fig:app-occurrences-after-sampler}
\end{subfigure}

\centering

\input{texfigures/No_importance_sampling_subfigure_appendix}

\input{texfigures/SGD_subfigure_appendix}

\input{texfigures/Adam_with_sampler_appendix_subfig}

\caption{\textbf{Weighted sampling effect.} Optimization with Adam requires weighted sampling to train properly (see text in \cref{subsec:app-sampler}).}
\label{fig:app-weighted-sampler}
\end{figure}

For highly imbalanced posteriors where the dominant mode is significantly more likely than the weakest mode (\eg $10\times$), we found that an additional regularization complementary to the one provided by $\varepsilon_t$ is needed. This is because even if we eliminate the dependence of training on leaf initialization, still, leaves associated with (much) less likely posterior modes will be chosen with a lower frequency during training, resulting in transient gradients that highly affect the adaptive normalization in Adam. To test this hypothesis, we trained two models on the task of image inpainting: (i) one with Adam using an initial learning rate of 0.001, and (ii) one with SGD using a momentum of 0.9 and an initial learning rate of 0.1. \Cref{fig:app-weighted-sampler} demonstrates the results of this experiment. As evident in the resulting trees, Adam leads to highly implausible predictions in leaves with low likelihoods (\eg $p=0.0$ in \cref{fig:app-adam-no-is}). On the other hand, SGD does not lead to nonsensical predictions; however, it requires significantly longer training time and is more challenging to converge (\cref{fig:app-sgd}).

Therefore, in our method, we proposed a novel weighted sampling scheme as a simple fix (\cref{fig:app-adam-with-is}). The proposed weighted sampler enables us to still enjoy the convergence speed of Adam while tackling the aforementioned optimization deficiency of MCL training for posteriors that are far from uniform. The goal of our sampler is to ensure that on average the number of occurrences at each output leaf during training is roughly the same (\cref{fig:app-occurrences-after-sampler}). This is done by undersampling training samples associated with leaves that represent probable posterior modes while oversampling training samples associated with leaves that represent rare posterior modes.

Formally, let $\vl_i$ denote the $i$th prediction leaf ($i=1,\dots,K^d$), and $\vs_j=(\vx_j,\vy_j)$ denote the $j$th (paired) training sample. We assume we are given a training set $\mathcal{D}=\{\vs_j\}_{j=1}^N$ of $N$ \emph{i.i.d}.\ samples, such that $p_{\rvs}(\vs_j)=\frac{1}{N}$. The goal of the sampler is to manipulate $p_{\rvs}(\vs_j)$ via oversampling/undersampling, such that the new sample probability in training $q_{\rvs}(\vs_j)$, leads to a uniform marginal leaf distribution, \ie $q_{\rvl}(\vl_i)=\frac{1}{K^d}$. To estimate the marginal leaf probability during training, we kept track of an association matrix $\mA\in\sR^{K^d\times N}$, where $\emA_{i,j}$ counts the number of times sample $\vs_j$ was associated with leaf $\vl_i$ over some time window. After each training batch $\vb_{t'}$ of size $B=\lvert \vb_{t'} \rvert$, the association matrix is updated with momentum $\mu=2^{-\frac{B}{N}}$, such that
\begin{equation}
    \mA_{t'+1} \leftarrow \mu \mA_{t'} + (1 - \mu) \mA_{t'+1}.
    \label{eq:app-assoc-mat}
\end{equation}
This smoothing was necessary to make sample probability change gracefully and avoid changing the probability of uncertain samples that switch leaf association in between batches. Next, given the updated association matrix, we can normalize the result to obtain an estimate of the current joint distribution $p_{\rvl,\rvs}(\vl_i,\vs_j)$, and further estimate the current conditional leaf probability via marginalization
\begin{equation}
    p_{\rvl|\rvs}(\vl_i|\vs_j) = \frac{p_{\rvl,\rvs}(\vl_i,\vs_j)}{\sum_{i=1}^{K^d} p_{\rvl,\rvs}(\vl_i,\vs_j)}.
    \label{eq:app-curr-pleaf}
\end{equation}
Let $\mP\in\sR^{K^d\times N}$ denote the matrix where $\emP_{i,j}=p_{\rvl|\rvs}(\vl_i|\vs_j)$. Recall that our goal was to dictate a new sample probability $q_{\rvs}(\vs_j)$ such that the induced marginal leaf probability $q_{\rvl}(\vl_i)$ is uniform. In matrix form, we are looking for a probability vector $\vq\in\sR^{N\times1}$, that satisfies $\frac{1}{K^d}\bm{1}_{K^d\times1}=\mP\vq$. Naturally, since $K^d\ll N$, there are many different options $\vq$ that satisfy this constraint. Therefore, we opted to solve the Tikhonov-regularized problem
\begin{equation}
\begin{aligned}
\vq^{\star} = \argmin_{\vq} \quad & \frac{1}{2}\|\mP\vq-\frac{1}{K^d}\bm{1}\|_2^2 + \frac{\lambda}{2}\|\vq\|_2^2 \\
\textrm{s.t.} \quad & \vq_j\geq0, \quad j=1,\dots,N \\
& \vq^{\top}\bm{1}_{N\times1}=1
\end{aligned}
\label{eq:app-qs-opt-prob}
\end{equation}

Obtaining the exact solution to \cref{eq:app-qs-opt-prob}, required roughly $\approx$2 seconds in typical settings using cvxpy \citep{diamond2016cvxpy}. This resulted in significantly longer training times since it is done after every batch during training. Therefore, we opted to use an approximation. Specifically, if we drop the positivity constraint $\vq_j\geq0, \forall j$, and solve \cref{eq:app-qs-opt-prob} accounting only for the constraint $\vq^{\top}\bm{1}_{N\times1}=1$, we obtain the following closed-form solution
\begin{equation}
\begin{aligned}
    \tilde{\vq} & = (\mI_{N\times N}-\mP^{\top}(\mP\mP^{\top}+\lambda\mI_{K^d \times K^d})^{-1}\mP)\bm{1}_{N\times1} \\
    \vq^{\diamond} & = \frac{1}{\tilde{\vq}^{\top}\bm{1}_{N\times1}} \tilde{\vq}. 
\end{aligned}
\label{eq:app-qs-approx}
\end{equation}
In practice, in most cases, the positivity constraint is not active, and therefore $\vq^{\diamond}=\vq^{\star}$. In the rare cases where \cref{eq:app-qs-approx} resulted in negative values, we clipped entries below zero in $\vq^{\diamond}$ and renormalized the result.

During training, we activated our weighted sampler one epoch after $\varepsilon_t$ started decaying to 0. At each training step, the next batch of samples was chosen such that the difference between the current estimated sample probability $p_{\rvs}(\vs_j)=\sum_{i=1}^{K^d}p_{\rvl,\rvs}(\vl_i,\vs_j)$ and the desired sample probability $\vq^{\diamond}$ is minimized.

Finally, note that switching from $\vp=\frac{1}{N}\bm{1}_{N\times1}$ to $\vq^{\diamond}$ during training effectively changes the dataset by repetition/omission, and if not accounted for will distort the learned posterior probabilities. Therefore, to undo this undesired effect while still maintaining an increased number of optimization steps for weak posterior modes, we scaled the loss of sample $\vs_j$ by the ratio $\gamma_j=\frac{1}{N\vq_j^{\diamond}}$. Hence, we effectively only changed the number of optimization steps taken per output leaf, enabling all leaves to continuously receive gradients and train equally as well with Adam. 

\clearpage

\section{Unconditional Hierarchical $K$-means}

In \cref{subsec:multi-out} we mentioned that in a discrete setting, our approach is equivalent to a \emph{hierarchical} $K$-means clustering. Here, we describe the hierarchical $K$-means algorithm and review its properties in the standard (unconditional) setting.

Given a set of $N$ data points $\{\vx_i\}$ where $\vx_i\in\R^{d_x}$, the goal of $K$-means is to partition the data points into $K$ clusters/sets $\{\gC_1,\dots,\gC_K\}$, such that each data point belongs to the cluster with the nearest mean/centroid serving as a prototype of the cluster. This results in a partitioning/tessellation of the data space into centroidal Voronoi cells, where for each cluster the within-cluster variances (squared Euclidean distances) are minimized. Formally, the objective function for finding the clusters is given by
\begin{equation}
    \argmin_{\gC_1,\dots,\gC_K} {\sum_{k=1}^K\sum_{\vx_i\in\gC_k}\|\vx_i-\vmu_k\|_2^2},
\label{eq:kmeans-obj}
\end{equation}
where $\vmu_k$ is the mean/centroid of cluster $\gC_k$, given by $\vmu_k = \frac{1}{|\gC_k|}\sum_{\vx_i\in\gC_k}{\vx_i}$. 
While in general Problem (\ref{eq:kmeans-obj}) is NP-hard, given some initial means/centroids, the $K$-means algorithm (also known as Lloyd's method) finds a local minimum by alternating between (i) assigning each data point to the cluster with the nearest mean, and (ii) updating the cluster means based on the new assignment. 

Extending $K$-means hierarchically is rather straightforward. Starting from a single cluster comprised of all data points $\{\vx_i\}$, at each level of the hierarchy, the data points belonging to the $k$th cluster $\gC_k$ are split further into $K$ sub-clusters $\{\gC_{k,1},\dots,\gC_{k,K}\}$ by applying $K$-means. We refer to this extended algorithm as \emph{hierarchical} $K$-means. The result of applying hierarchical $K$-means $d$ times is a dendrogram/top-down balanced tree $\mathcal{T}$ of degree $K$, depth $d$, and breadth (final number of leaves) $K^d$. Note that this successive process is different from applying $K$-means with $K^d$ clusters once. This is because the hierarchical memberships are enforced, restricting data points in subclusters $\{\gC_{i,1},\dots,\gC_{i,K}\}$ from being assigned to subclusters $\{\gC_{j,1},\dots,\gC_{j,K}\}$ if $i\neq j$.

Hierarchical $K$-means enjoys several advantages compared to the standard (``Flat'') $K$-means algorithm. One of these advantages is increased robustness to initialization as demonstrated in \cref{subsec:app-flat-vs-hier}.

\newpage
\section{Toy Example}\label{subsec:app-toy-example}

\subsection{Model Architecture}

The model used to produce our results in \cref{fig:gmm-2d} consisted of two 5-layer MLPs with 256 hidden units, one for predicting the leaves and one for predicting the likelihoods. Both MLPs consisted of linear layers interleaved with the SiLU activation. In addition, the output of the second MLP was passed through a $\softmax$ to represent a valid probability distribution. 

\subsection{Analytical Posterior}

\begin{figure}[t]
  \centering
  \begin{subfigure}{0.25\linewidth}
    \includegraphics[width=\textwidth, trim={0.2cm 0.2cm 0.2cm 0.2cm}, clip]{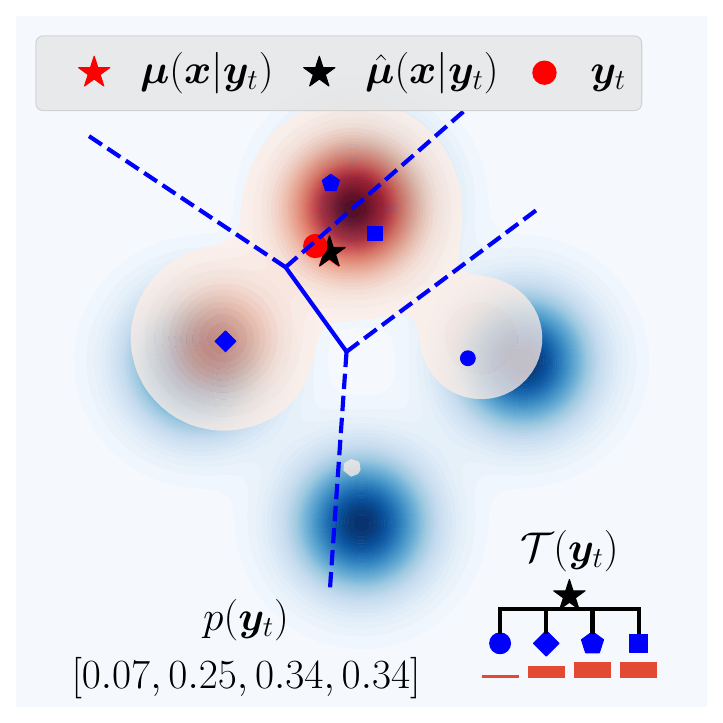}
  \end{subfigure}%
  \begin{subfigure}{0.25\linewidth}
    \includegraphics[width=\textwidth, trim={0.2cm 0.2cm 0.2cm 0.2cm}, clip]{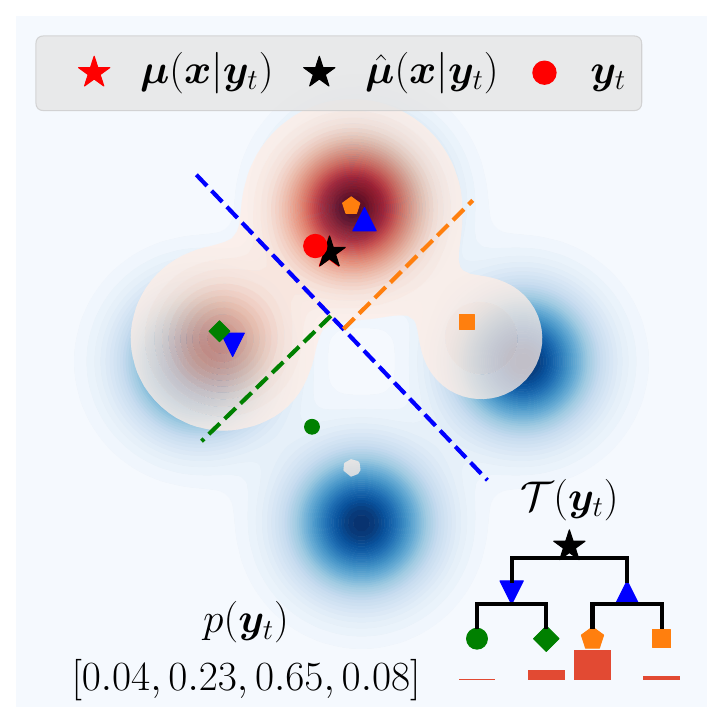}
  \end{subfigure}%
  \begin{subfigure}{0.25\linewidth}
    \includegraphics[width=\textwidth, trim={0.2cm 0.2cm 0.2cm 0.2cm}, clip]{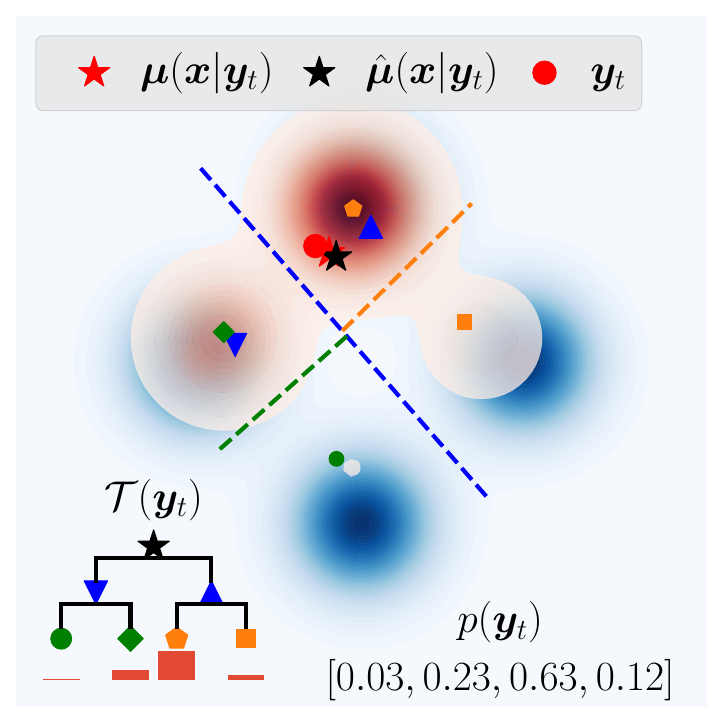}
  \end{subfigure}%
  \begin{subfigure}{0.25\linewidth}
    \includegraphics[width=\textwidth, trim={0.2cm 0.2cm 0.2cm 0.2cm}, clip]{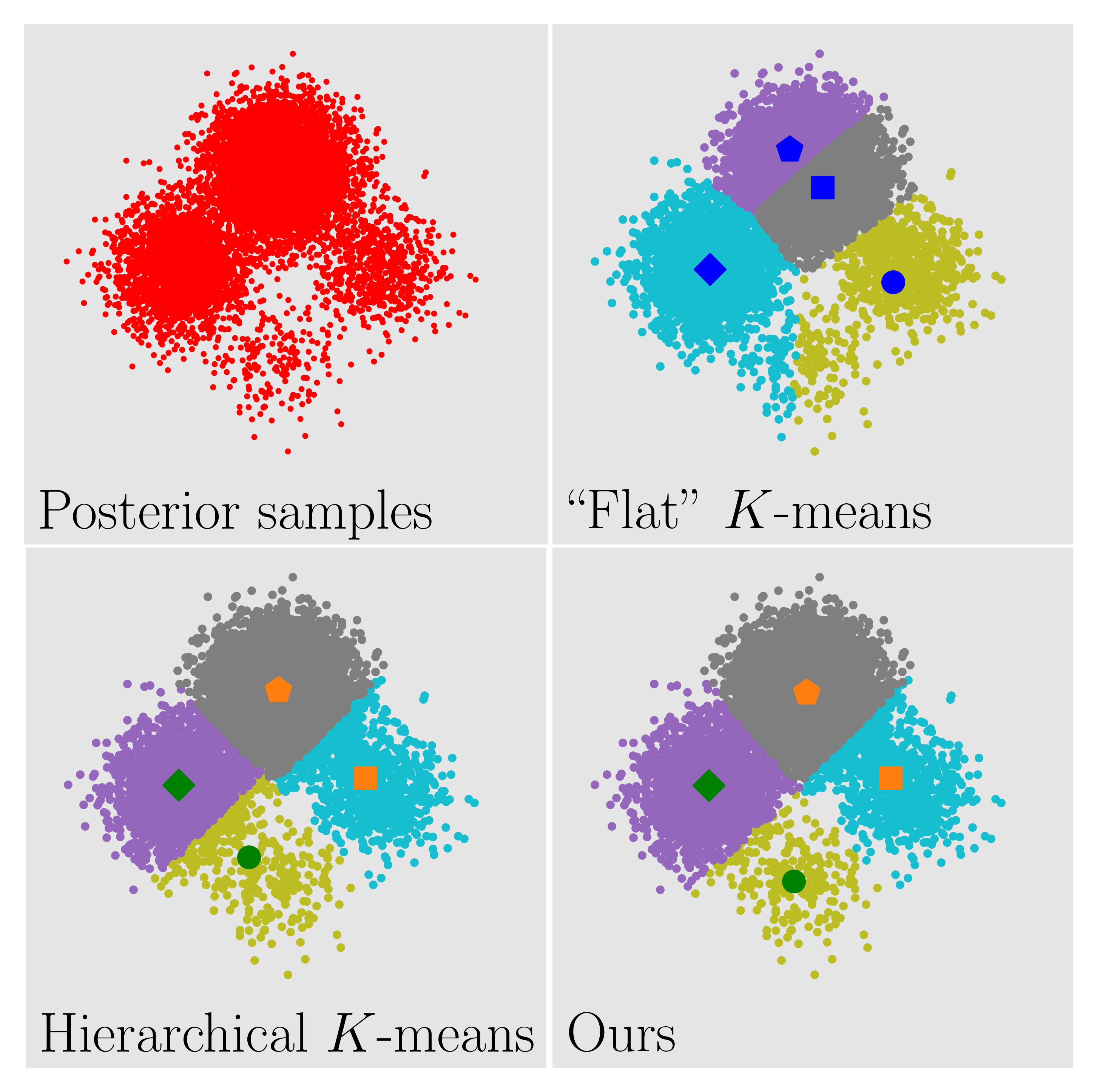}
  \end{subfigure}
  \begin{subfigure}{0.25\linewidth}
    \includegraphics[width=\textwidth, trim={0.2cm 0.2cm 0.2cm 0.2cm}, clip]{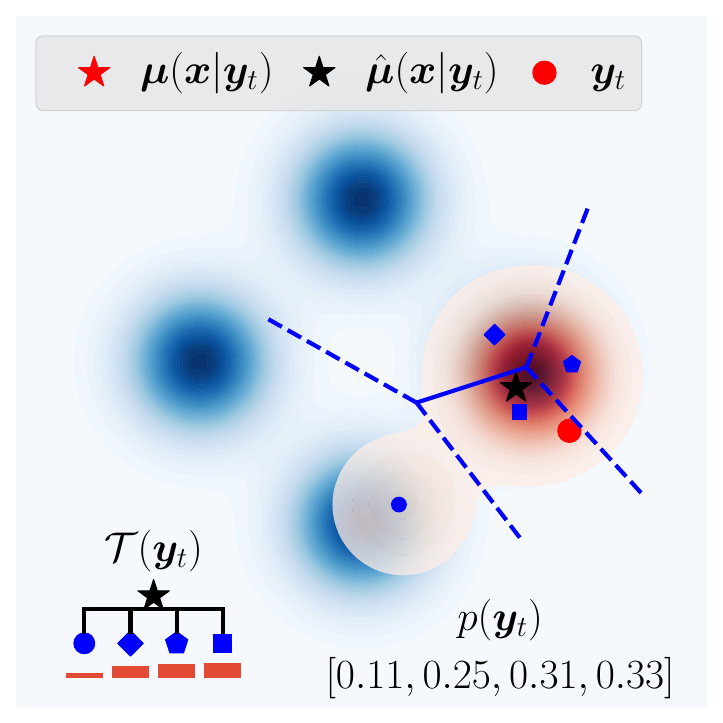}
  \end{subfigure}%
  \begin{subfigure}{0.25\linewidth}
    \includegraphics[width=\textwidth, trim={0.2cm 0.2cm 0.2cm 0.2cm}, clip]{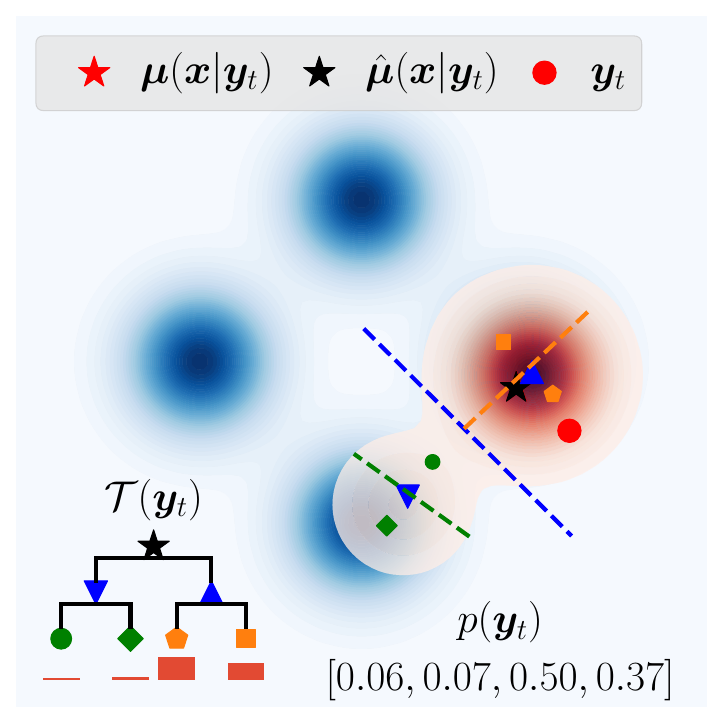}
  \end{subfigure}%
  \begin{subfigure}{0.25\linewidth}
    \includegraphics[width=\textwidth, trim={0.2cm 0.2cm 0.2cm 0.2cm}, clip]{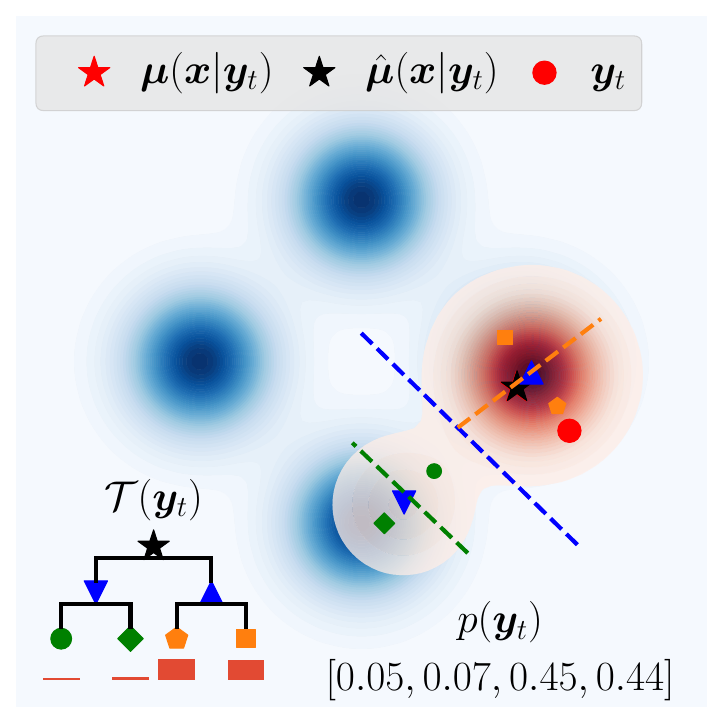}
  \end{subfigure}%
  \begin{subfigure}{0.25\linewidth}
    \includegraphics[width=\textwidth, trim={0.2cm 0.2cm 0.2cm 0.2cm}, clip]{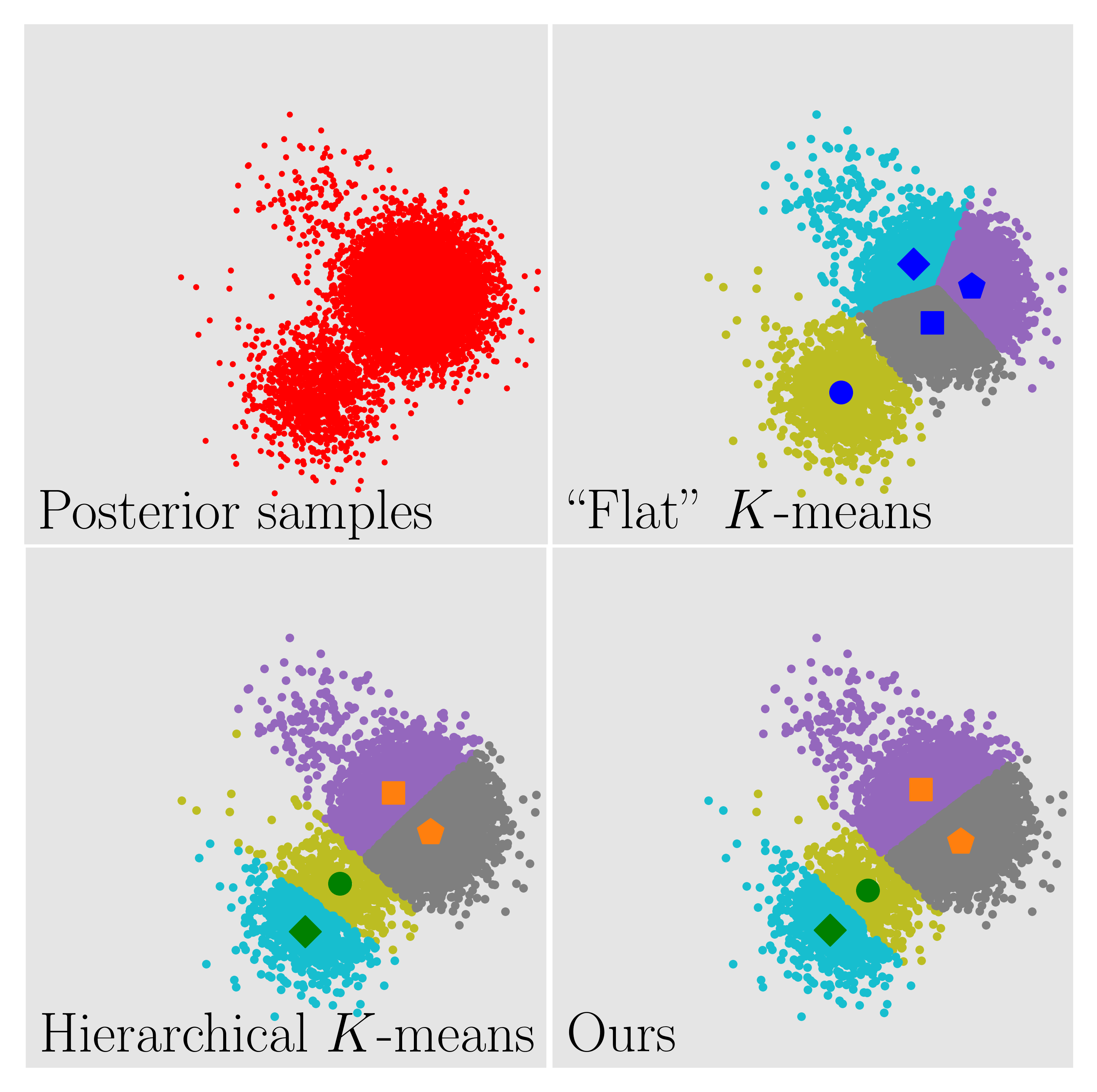}
  \end{subfigure}
  \begin{subfigure}{0.25\linewidth}
    \includegraphics[width=\textwidth, trim={0.2cm 0.2cm 0.2cm 0.2cm}, clip]{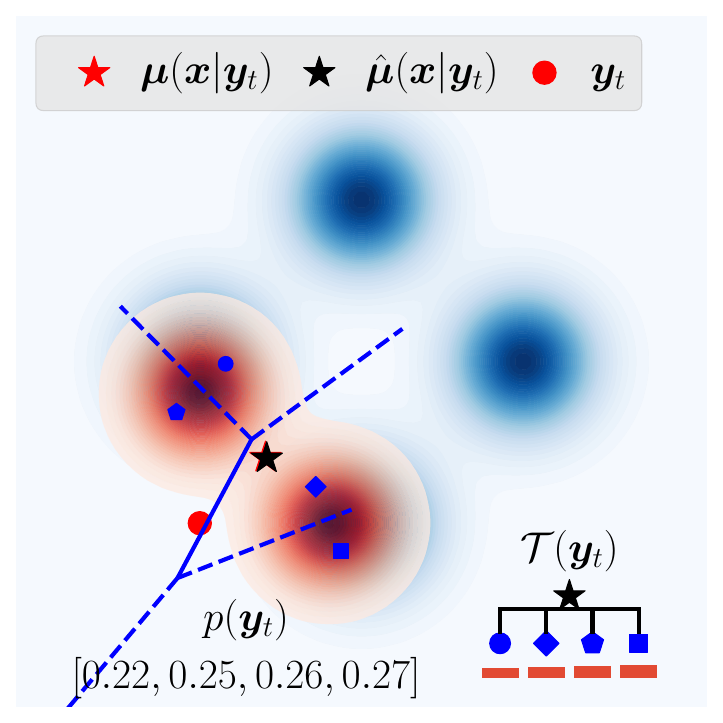}
    \caption{``Flat'' $K$-means}
  \end{subfigure}%
  \begin{subfigure}{0.25\linewidth}
    \includegraphics[width=\textwidth, trim={0.2cm 0.2cm 0.2cm 0.2cm}, clip]{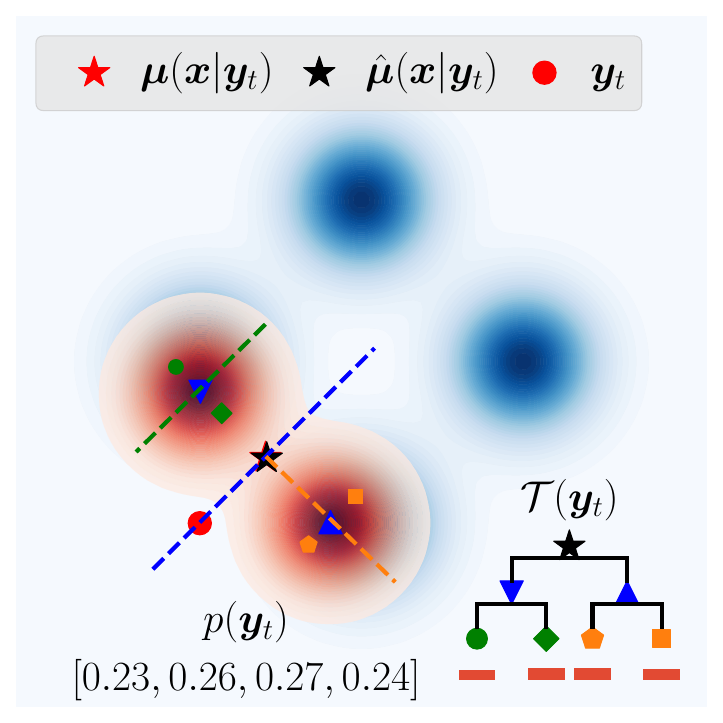}
    \caption{Hier. $K$-means}
  \end{subfigure}%
  \begin{subfigure}{0.25\linewidth}
    \includegraphics[width=\textwidth, trim={0.2cm 0.2cm 0.2cm 0.2cm}, clip]{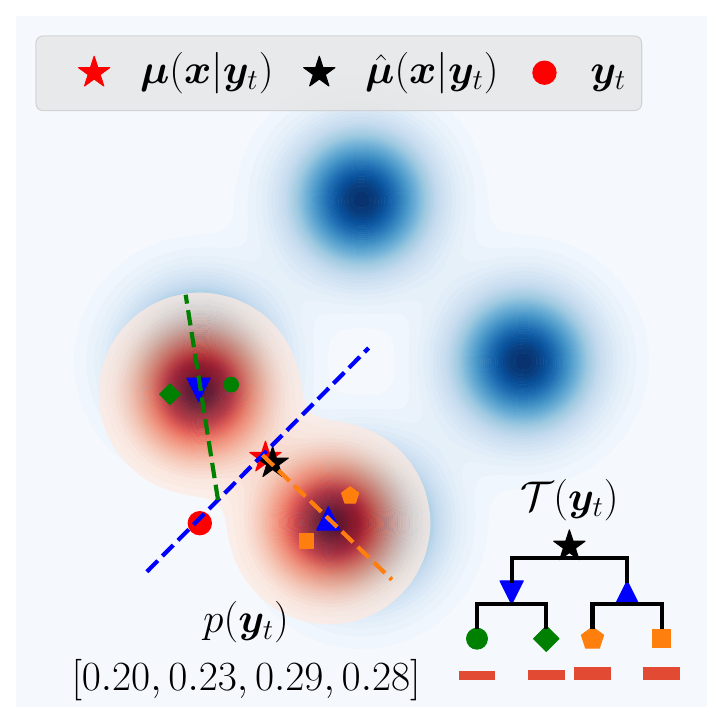}
    \caption{Ours}
  \end{subfigure}%
  \begin{subfigure}{0.25\linewidth}
    \includegraphics[width=\textwidth, trim={0.2cm 0.2cm 0.2cm 0.2cm}, clip]{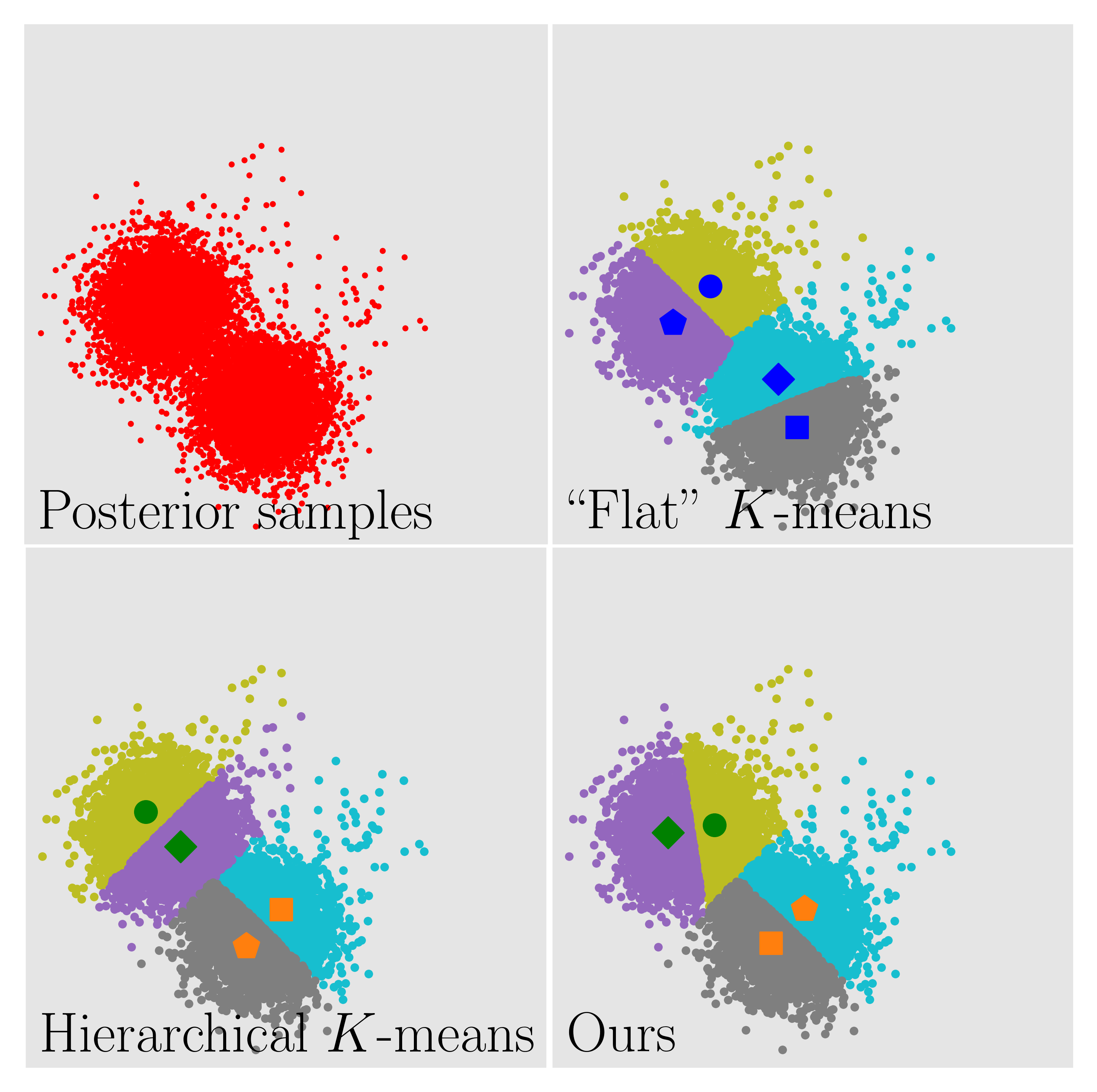}
    \caption{Clusters}
  \end{subfigure}
  \caption{\textbf{Additional test points $\vy_t$}. Each row above shows the results for a different test point $\vy_t$ (red dot) when clustering with (a) ``Flat'' $K$-means, (b) Hierarchical $K$-means, (c) Posterior trees (ours). In (d) we show the resulting partition/clustering induced by each method by coloring 10K samples from the posterior $p_{\rvx|\rvy}(\vx|\vy_t)$ according to their nearest cluster. Our method consistently recovers the result of hierarchical $K$-means in all cases. Moreover, other than trivial cases such as the bottom row, the hierarchy better represents weaker modes compared to ``Flat'' $K$-means.}
  \label{fig:gmm-supp-more-pts}
\end{figure}

In \cref{fig:gmm-2d}, we plotted the posterior trees on the (unknown) ground truth posterior. Here we provide the closed-form expression for the analytically derived posterior for completeness. The denoising task we assumed was $\rvy = \rvx + \rvn$, where $\rvx$ comes from a mixture of $L=4$ Gaussians, $p_{\rvx}(\vx)=\sum_{\ell=1}^L \pi_{\ell}\mathcal{N}(\vx;\vmu_{\ell},\mSigma_{\ell})$, and $\rvn\sim\mathcal{N}(\cdot;\bm{0},\sigma_{\varepsilon}^2\mI)$ is a white Gaussian noise. Specifically, in our toy example we used equally probable spherical Gaussians (\ie $\pi_\ell=\frac{1}{4}, \mSigma_\ell=\mI$), with the following means
\begin{equation}
    \vmu_1 = \begin{pmatrix}-6.0\\+2.5\end{pmatrix}, \;\vmu_2 = \begin{pmatrix}+1.0\\+2.5\end{pmatrix}, \;\vmu_3 = \begin{pmatrix}-2.5\\+6.0\end{pmatrix}, \;\vmu_4 = \begin{pmatrix}-2.5\\-1.5\end{pmatrix}.  
\end{equation}
Let $\rc$ be an auxiliary random variable taking values in $\{1,\dots,L\}$ with probabilities $\{\pi_1,\dots,\pi_L\}$. Then we can write the posterior by invoking the law of total probability conditioned on the event $\rc=\ell$,
\begin{align}
    p(\vx|\vy)&=\sum_{\ell=1}^L p_{\rvx|\rvy,\rc}(\vx|\vy,\ell)p_{\rc|\rvy}(\ell|\vy)\nonumber\\ 
    &=\sum_{\ell=1}^L p_{\rvx|\rvy,\rc}(\vx|\vy,\ell)\frac{p_{\rvy|\rc}(\vy|\ell)p_{\rc}(\ell)}{p_{\rvy}(\vy)}\nonumber\\
    &=\sum_{\ell=1}^L \mathcal{N}(\vx;\tilde{\vmu}_{\ell},\tilde{\mSigma}_{\ell})\frac{q_{\ell} \pi_{\ell}}{\sum_{\ell'=1}^L q_{\ell'} \pi_{\ell'}},
    \label{eq:gmm-posterior}
\end{align}
where the first step is by Bayes rule, and in the result we denoted
\begin{align}
    q_\ell&=\mathcal{N}(\vy;\vmu_\ell,\mSigma_\ell+\sigma_{\varepsilon}^2\mI),\nonumber\\ 
    \tilde{\vmu}_\ell&=\vmu_\ell + \mSigma_\ell(\mSigma_\ell+\sigma_{\varepsilon}^2\mI)^{-1}(\vy-\vmu_\ell)\nonumber\\
    \tilde{\mSigma}_\ell&=\mSigma_\ell-\mSigma_\ell(\mSigma_\ell+\sigma_{\varepsilon}^2\mI)^{-1}\mSigma_\ell, \quad \ell=1,\dots,L.
    \label{eq:gmm-posterior-inner-terms}
\end{align}
In \cref{fig:gmm-supp-more-pts}, in a similar fashion to \cref{fig:gmm-2d} from the main text, we visualize posterior trees for additional test inputs $\vy_t$. As clearly evident in all cases, our method recovers the (approximated) ground truth posterior tree with high accuracy.

\subsection{Stability of ``Flat'' vs. Hierarchical $K$-means}\label{subsec:app-flat-vs-hier}

\begin{figure}[t]
  \centering
  \begin{subfigure}{1.0\linewidth}
  \begin{subfigure}{0.25\linewidth}
    \includegraphics[width=\textwidth, trim={0.2cm 0.2cm 0.2cm 0.2cm}, clip]{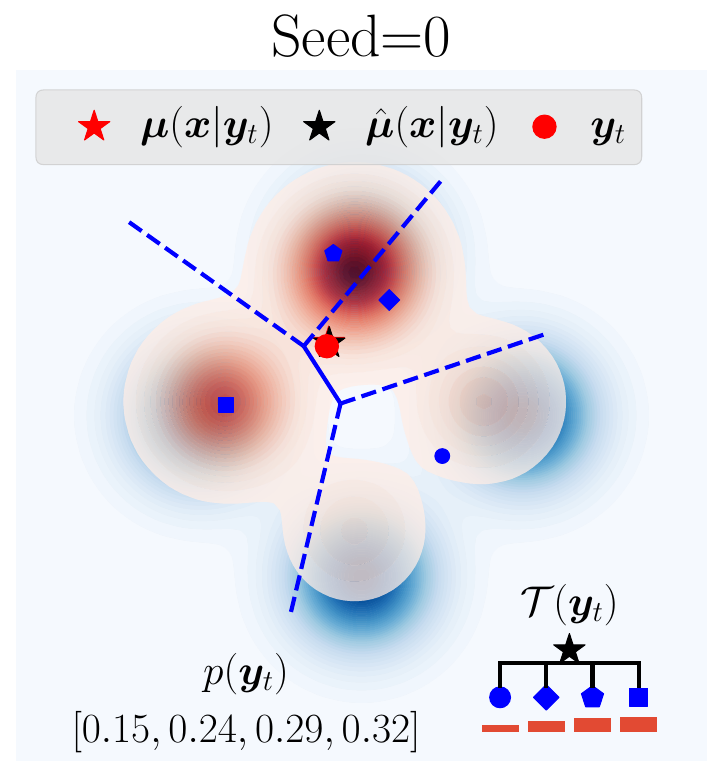}
  \end{subfigure}%
  \begin{subfigure}{0.25\linewidth}
    \includegraphics[width=\textwidth, trim={0.2cm 0.2cm 0.2cm 0.2cm}, clip]{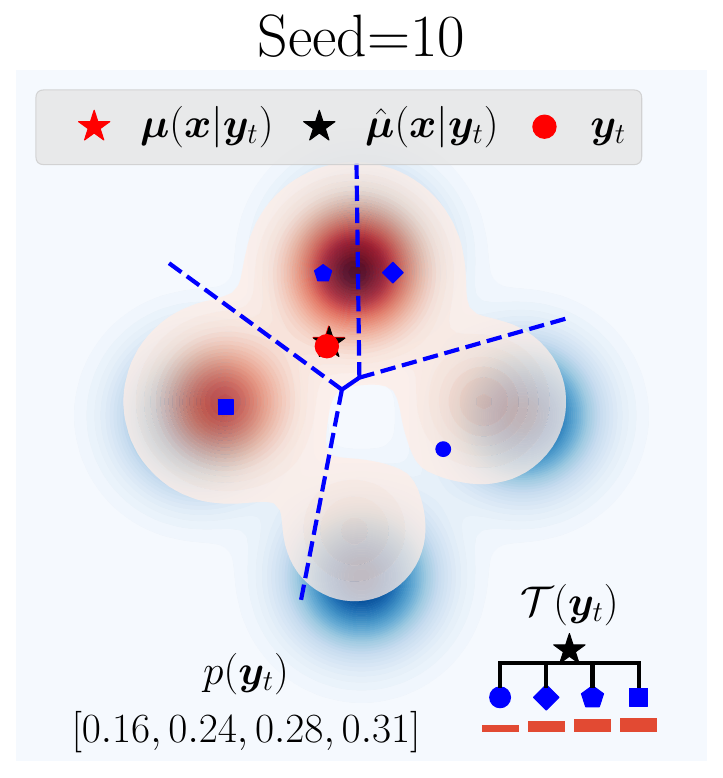}
  \end{subfigure}%
  \begin{subfigure}{0.25\linewidth}
    \includegraphics[width=\textwidth, trim={0.2cm 0.2cm 0.2cm 0.2cm}, clip]{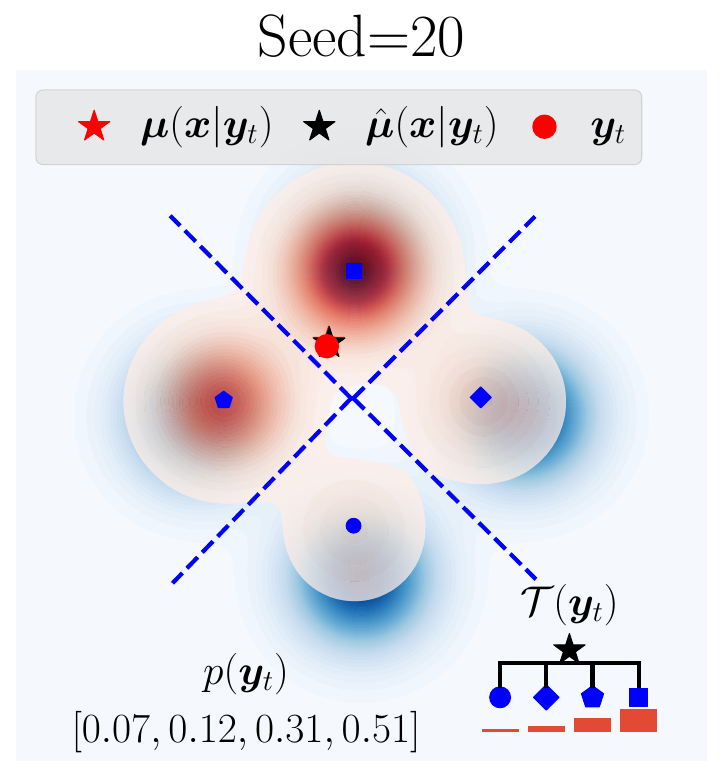}
  \end{subfigure}%
  \begin{subfigure}{0.25\linewidth}
    \includegraphics[width=\textwidth, trim={0.2cm 0.2cm 0.2cm 0.2cm}, clip]{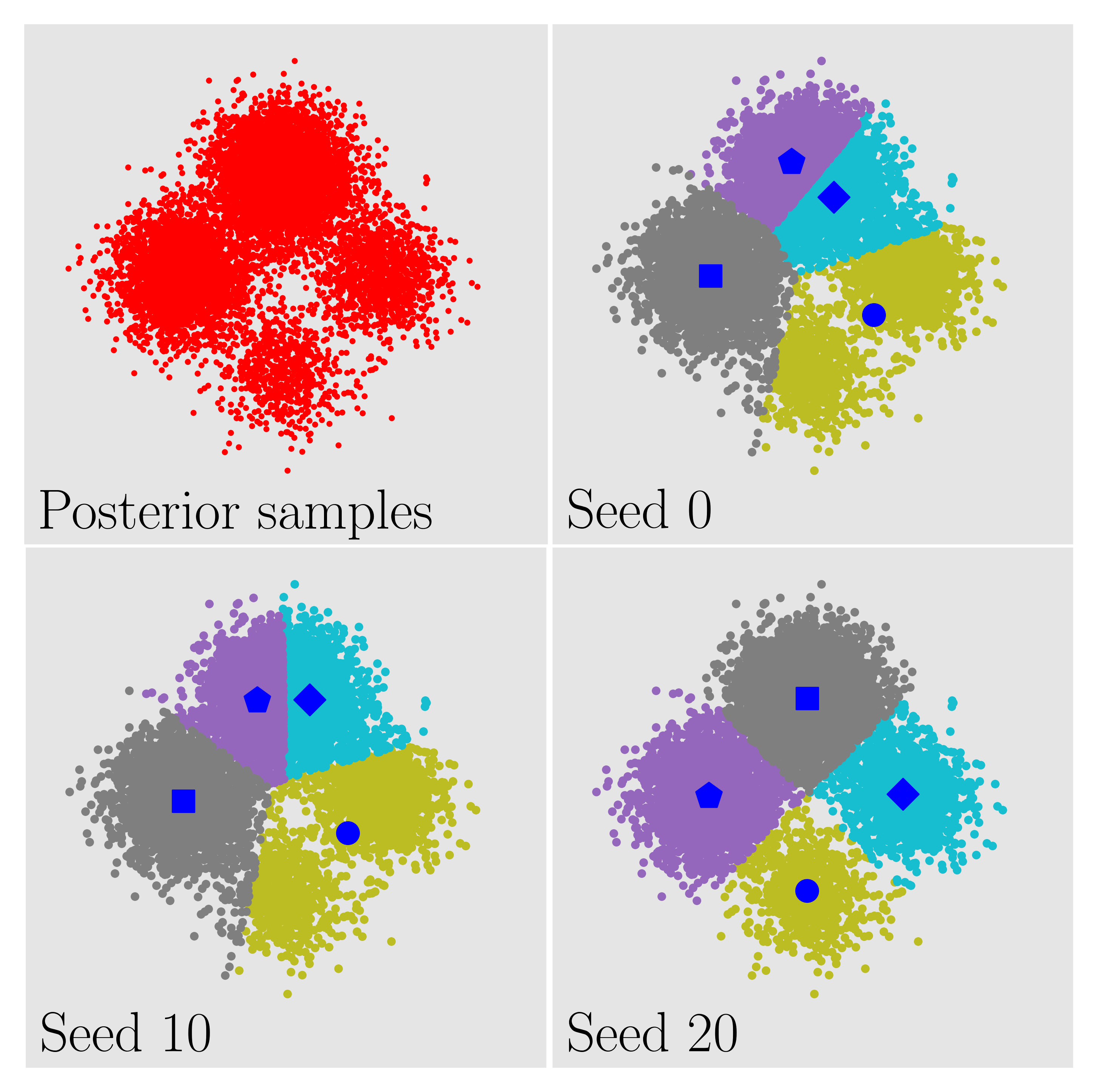}
  \end{subfigure}
  \caption{``Flat'' $K$-means}
  \end{subfigure}
  \begin{subfigure}{1.0\linewidth}
  \begin{subfigure}{0.25\linewidth}
    \includegraphics[width=\textwidth, trim={0.2cm 0.2cm 0.2cm 0.2cm}, clip]{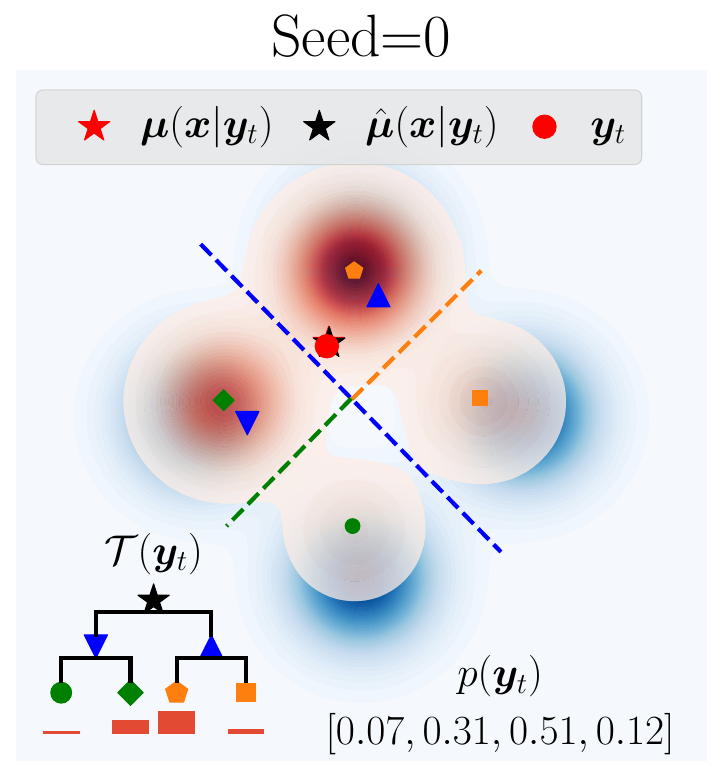}
  \end{subfigure}%
  \begin{subfigure}{0.25\linewidth}
    \includegraphics[width=\textwidth, trim={0.2cm 0.2cm 0.2cm 0.2cm}, clip]{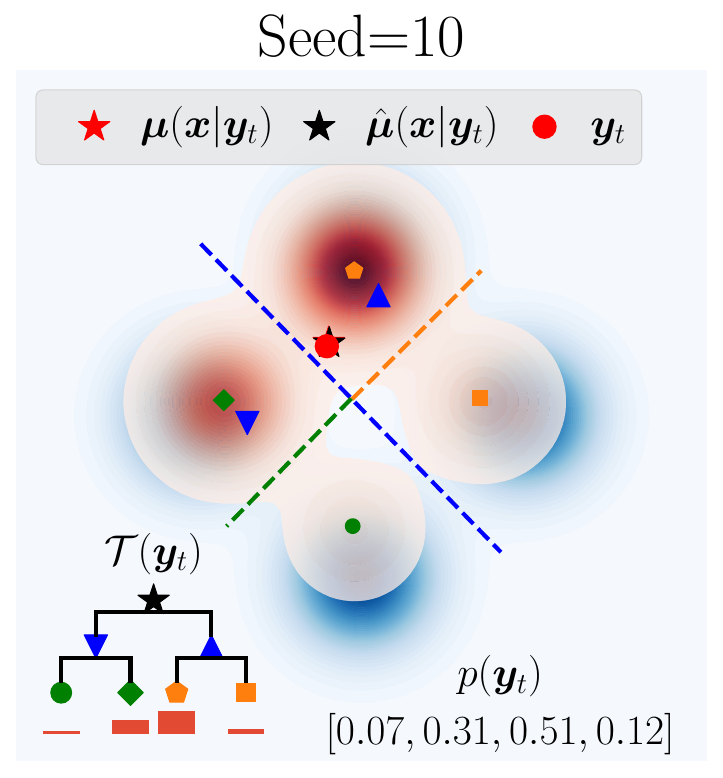}
  \end{subfigure}%
  \begin{subfigure}{0.25\linewidth}
    \includegraphics[width=\textwidth, trim={0.2cm 0.2cm 0.2cm 0.2cm}, clip]{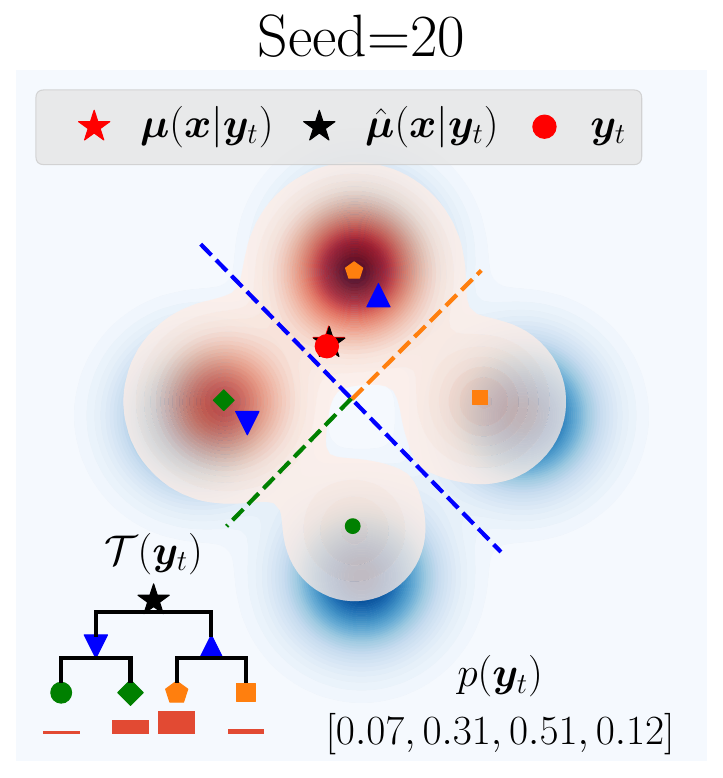}
  \end{subfigure}%
  \begin{subfigure}{0.25\linewidth}
    \includegraphics[width=\textwidth, trim={0.2cm 0.2cm 0.2cm 0.2cm}, clip]{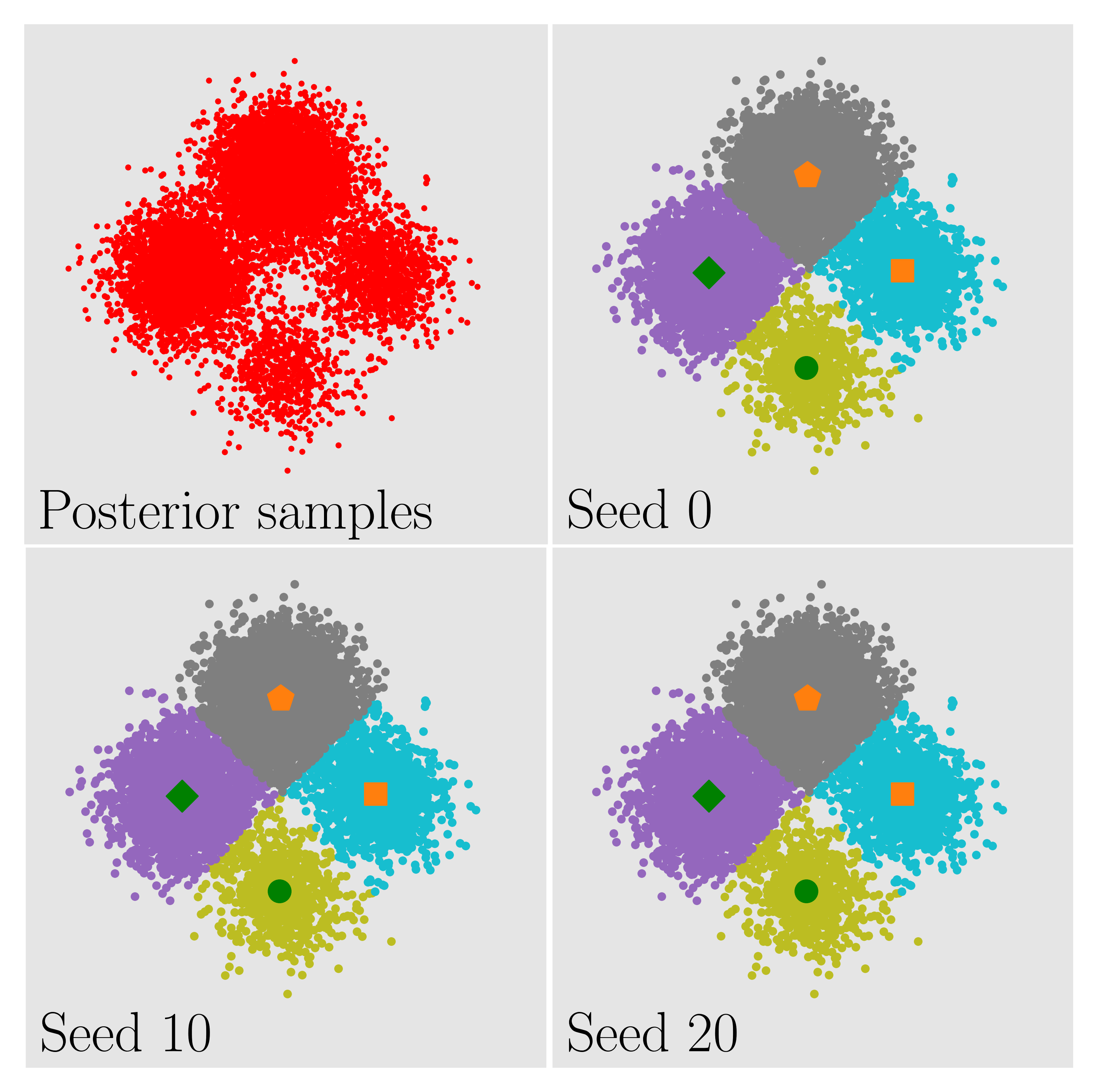}
  \end{subfigure}
  \caption{Hierarchical $K$-means}
  \end{subfigure}
  \caption{\textbf{Resilience to initialization}. Here we inspect the effect of the seed used for initialization, on the resulting clusters recovered by (a) ``Flat'' $K$-means and (b) Hierarchical $K$-means. The hierarchy results in a more resilient algorithm that recovers the same result regardless of the seed used for initialization.}
  \label{fig:gmm-supp-seed-stability}
\end{figure}

As mentioned earlier, the hierarchical $K$-means algorithm has several advantages over the classical ``Flat'' $K$-means algorithm. Here, we examine the effect of the random initialization on the resulting clusters found by ``Flat''/Hierarchical K-means. In both cases, we applied the respective method to 10K posterior samples to avoid errors resulting from an insufficient sample size. \Cref{fig:gmm-supp-seed-stability} demonstrates the resulting clusters for 3 different seeds. As can be seen by the result, despite using the widely adopted $K$-means$++$ initialization for both algorithms, ``Flat'' $K$-means was less resilient to a bad initialization compared to its hierarchical counterpart. This is yet another advantage of working with trees of depth bigger than $d=1$.  

\clearpage
\section{Tree Width vs. Depth}

\begin{figure}[t]

\begin{subfigure}{\linewidth}
    \input{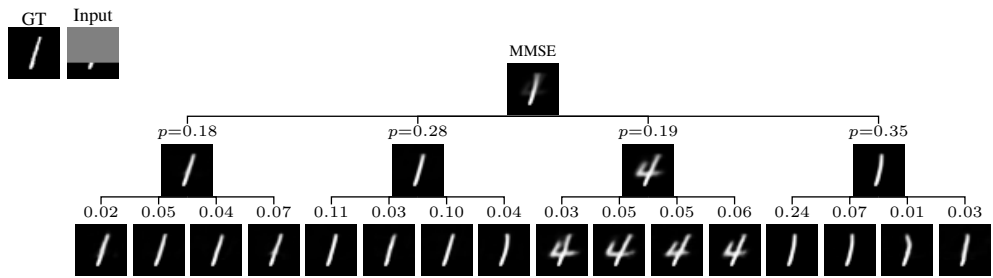}
    \caption{Tree with $K=4,d=2$}
    \label{fig:app-tree-4by2}
\end{subfigure}\vspace{0.2cm}\\
\begin{subfigure}{\linewidth}
    \input{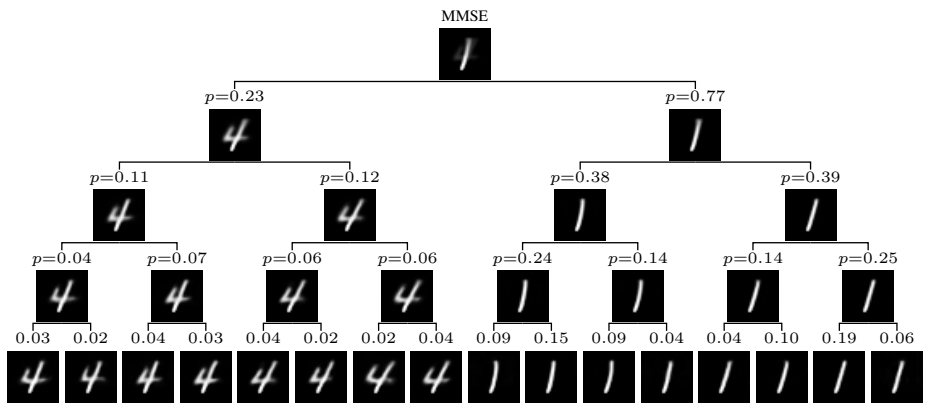}
    \caption{Tree with $K=2,d=4$}
    \label{fig:app-tree-2by4}
\end{subfigure}%
\caption{\textbf{Tree depth vs. width.} \protect\subref{fig:app-tree-4by2} Tree with $K=4,d=2$. \protect\subref{fig:app-tree-2by4} Binary tree with $K=2,d=4$ also resulting in 16 leaves. The binary tree emphasizes more the ``4'' mode relative to the ``1'' mode, although in both cases the probability mass associated with the ``4'' is $\approx 20\%$.}
\label{fig:app-tree-depth-vs-width}
\end{figure}

Throughout \cref{sec:exp} we presented results with trees of degree $K=3$ and depth $d=2$. This was done only for the convenience of the exposition and our method is not restricted to this specific layout. Here we examine the effect of tree width compared to tree depth. \Cref{fig:app-flat-vs-2by2-tree,fig:app-tree-depth-vs-width} presents the resulting trees for four different model configurations on the task of digit inpainting. For a fair comparison between model pairs, we fix the number of output leaves which dictates the number of model parameters, and only change the hierarchical structure. \Cref{fig:app-flat-tree-4by1} compares a layout of $K=4,d=1$ to $K=2,d=2$. As evident in the result, the hierarchy better organizes the different options, presenting the ``4'' cluster more faithfully than the flat tree which presents a cluster midway between a ``4'' and a ``1''. \Cref{fig:app-tree-depth-vs-width} repeats this experiment for deeper trees comparing $K=4,d=2$ to $K=2,d=4$. A couple of remarks are in order regarding the results. First, in both cases, the ``4'' cluster has a similar probability relative to the ``1'' cluster. This indicates that our method is consistent in the predicted posterior modes and their likelihoods, with the only change between different layouts being the chosen tessellation of the output space. Second, the degree $K$ controls the emphasis/over-representation devoted to weaker posterior modes. A smaller degree leads to more emphasis on weaker modes as tree depth grows. Lastly, it is important to note that the optimal layout $K$ and $d$ is task and input-dependant, and setting these adaptively is an interesting direction for future research. 

\begin{figure}[h]

\begin{subfigure}{0.5\linewidth}
    \input{texfigures/tree_depth_width_N4D1}
    \caption{Tree with $K=4, d=1$}
    \label{fig:app-flat-tree-4by1}
\end{subfigure}%
\begin{subfigure}{0.5\linewidth}
    \input{texfigures/tree_depth_width_N2D2}
    \caption{Tree with $K=2, d=2$}
    \label{fig:app-bin-tree-2by2}
\end{subfigure}%
\caption{\textbf{Flat vs. hierarchical trees.} \protect\subref{fig:app-flat-tree-4by1} Flat tree with $K=4,d=1$. \protect\subref{fig:app-bin-tree-2by2} Binary tree with $K=2,d=2$, with overall 4 leaves. The binary tree categorizes the leaves and better represents the ``4'' mode.}
\label{fig:app-flat-vs-2by2-tree}
\end{figure}

\clearpage
\section{GAN-based Posterior Samplers}\label{subsec:app-gan-sampler}

\begin{figure}[h]

\begin{subfigure}{\linewidth}
    \input{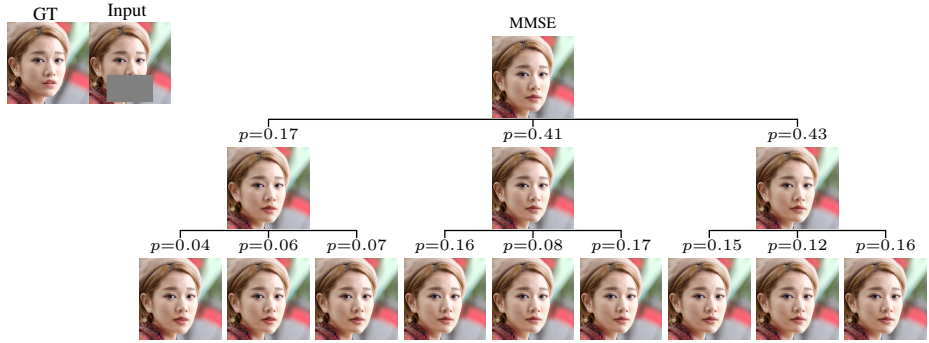}
    \caption{Mouth inpainting}
    \label{fig:app-gan-sampler-mouth}
\end{subfigure}\vspace{0.2cm}\\
\begin{subfigure}{\linewidth}
    \input{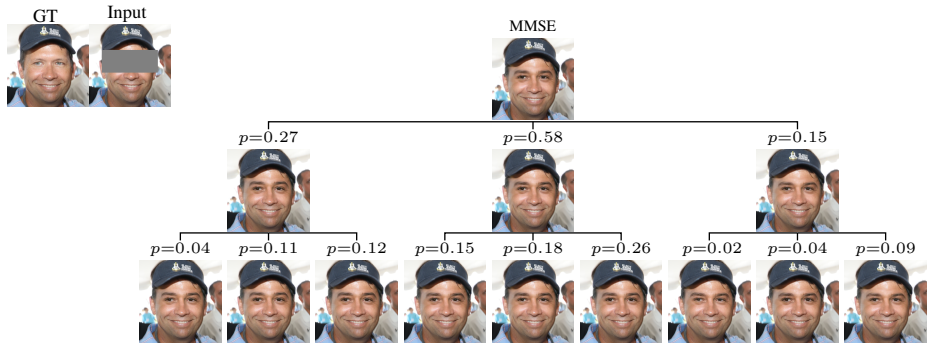}
    \caption{Eyes inpainting}
    \label{fig:app-gan-sampler-eyes}
\end{subfigure}%
\caption{\textbf{GAN-based posterior sampler.} \protect\subref{fig:app-gan-sampler-mouth}-\protect\subref{fig:app-gan-sampler-eyes} Posterior trees constructed with MAT, by applying hierarchical $K$-means to 100 samples in the tasks of mouth/eyes inpainting respectively. As evident in both cases, the resulting tree exhibits little to no variance due to mode collapse.}
\label{fig:app-gan-sampler}
\end{figure}

In \cref{sec:exp} we compared our method mainly to diffusion-based posterior samplers (\eg \citep{wang2023ddnm,kawar2022denoising,lugmayr2022repaint}). While these samplers often lead to state-of-the-art sample quality, they are known to be computationally intensive. In theory, GAN-based samplers such as MAT \citep{li2022mat}, can significantly speed up the proposed two-step baseline of sampling followed by hierarchical $K$-means; however, as evident in \cref{fig:app-gan-sampler}, even top-performing GAN-based samplers often collapse to a single mode of the posterior, producing samples with little variability that do not faithfully reflect the full distribution.

\clearpage
\section{Visual Comparison to Baselines}\label{subsec:app-vis-baselines}

\begin{figure}[h]
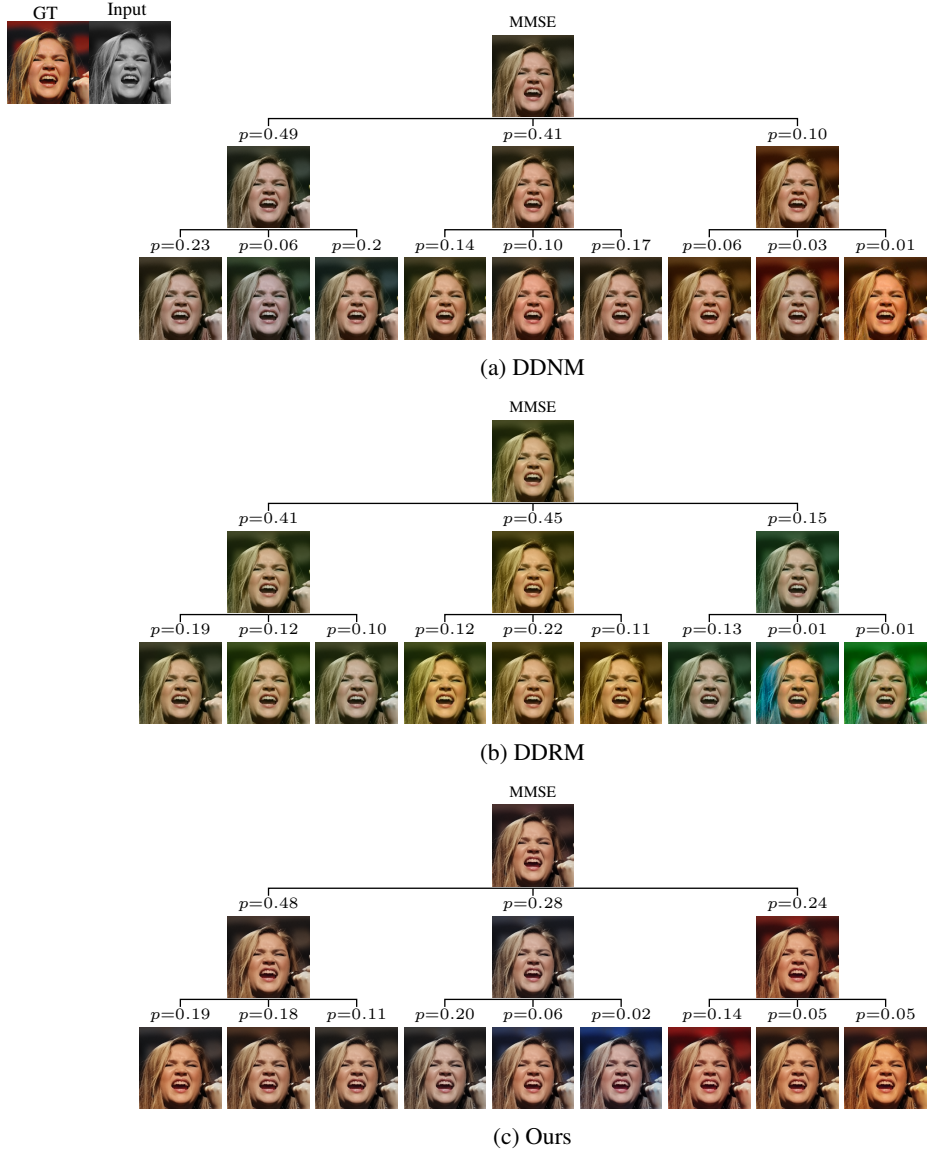


\begin{subfigure}{\linewidth}
    \input{texfigures/Tree_comparison_colorization_appendix_subfig1}
    \caption{DDNM}
    \label{fig:app-vis-bl-colorization-ddnm}
\end{subfigure}\vspace{0.2cm}\\
\begin{subfigure}{\linewidth}
    \input{texfigures/Tree_comparison_colorization_appendix_subfig2}
    \caption{DDRM}
    \label{fig:app-vis-bl-colorization-ddrm}
\end{subfigure}\vspace{0.2cm}\\
\begin{subfigure}{\linewidth}
    \input{texfigures/Tree_comparison_colorization_appendix_subfig3}
    \caption{Ours}
    \label{fig:app-vis-bl-colorization-ours}
\end{subfigure}%
\caption{\textbf{Tree comparison in colorization.}}
\label{fig:app-vis-bl-colorization}
\end{figure}

\begin{figure}[t]

\begin{subfigure}{\linewidth}
    \input{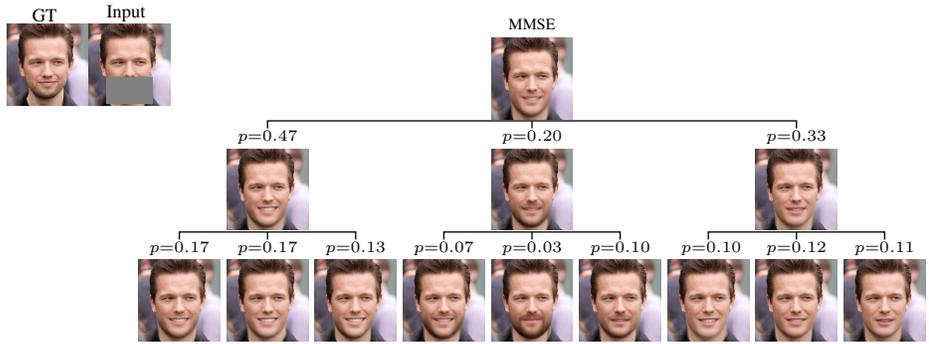}
    \caption{DDNM}
    \label{fig:app-vis-bl-mouth-inpaint-ddnm}
\end{subfigure}\vspace{0.2cm}\\
\begin{subfigure}{\linewidth}
    \input{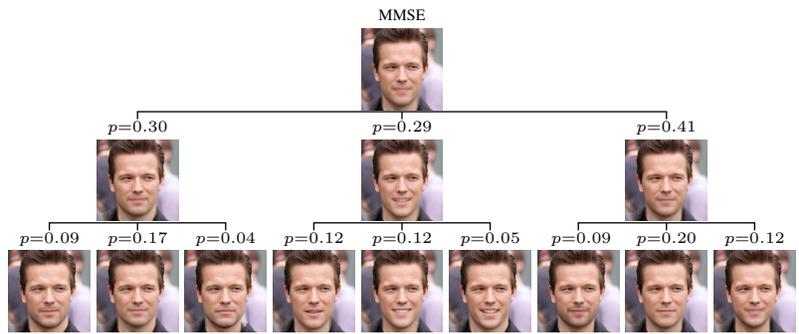}
    \caption{RePaint}
    \label{fig:app-vis-bl-mouth-inpaint-repaint}
\end{subfigure}\vspace{0.2cm}\\
\begin{subfigure}{\linewidth}
    \input{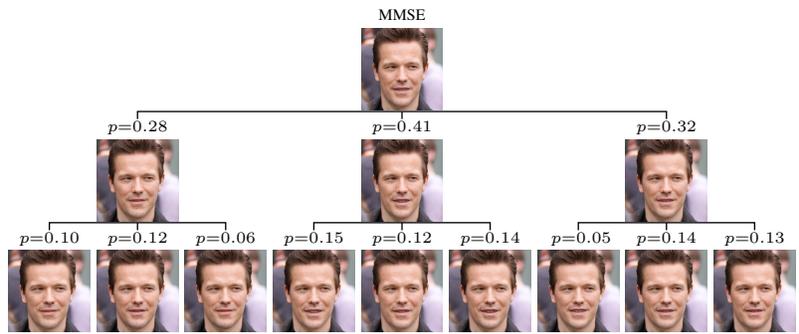}
    \caption{MAT}
    \label{fig:app-vis-bl-mouth-inpaint-mat}
\end{subfigure}\vspace{0.2cm}\\
\begin{subfigure}{\linewidth}
    \input{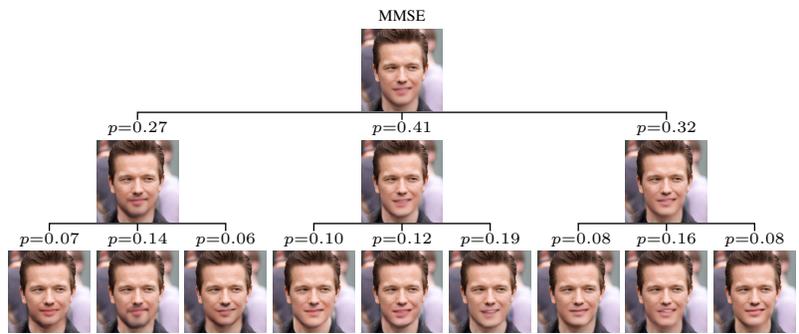}
    \caption{Ours}
    \label{fig:app-vis-bl-mouth-inpaint-ours}
\end{subfigure}%
\caption{\textbf{Tree comparison in mouth inpainting.}}
\label{fig:app-vis-bl-mouth-inpaint}
\end{figure}

In \cref{tab:ffhq-comp} we reported quantitative comparisons to the proposed baseline implemented with various samplers. In \cref{fig:app-vis-bl-mouth-inpaint,fig:app-vis-bl-colorization}, we include two sample visual comparisons of the resulting trees with our method and the baselines for the tasks of face colorization and mouth inpainting.


\clearpage
\section{More Results}\label{subsec:app-more-results}

Here we include more results for each of the tasks presented in \cref{sec:exp} from the main text. Additionally, we also include colorization results on the AFHQ(v2) dataset \citep{choi2020stargan}, with a similar performance to that obtained on CelebA-HQ.

\begin{figure}[h]
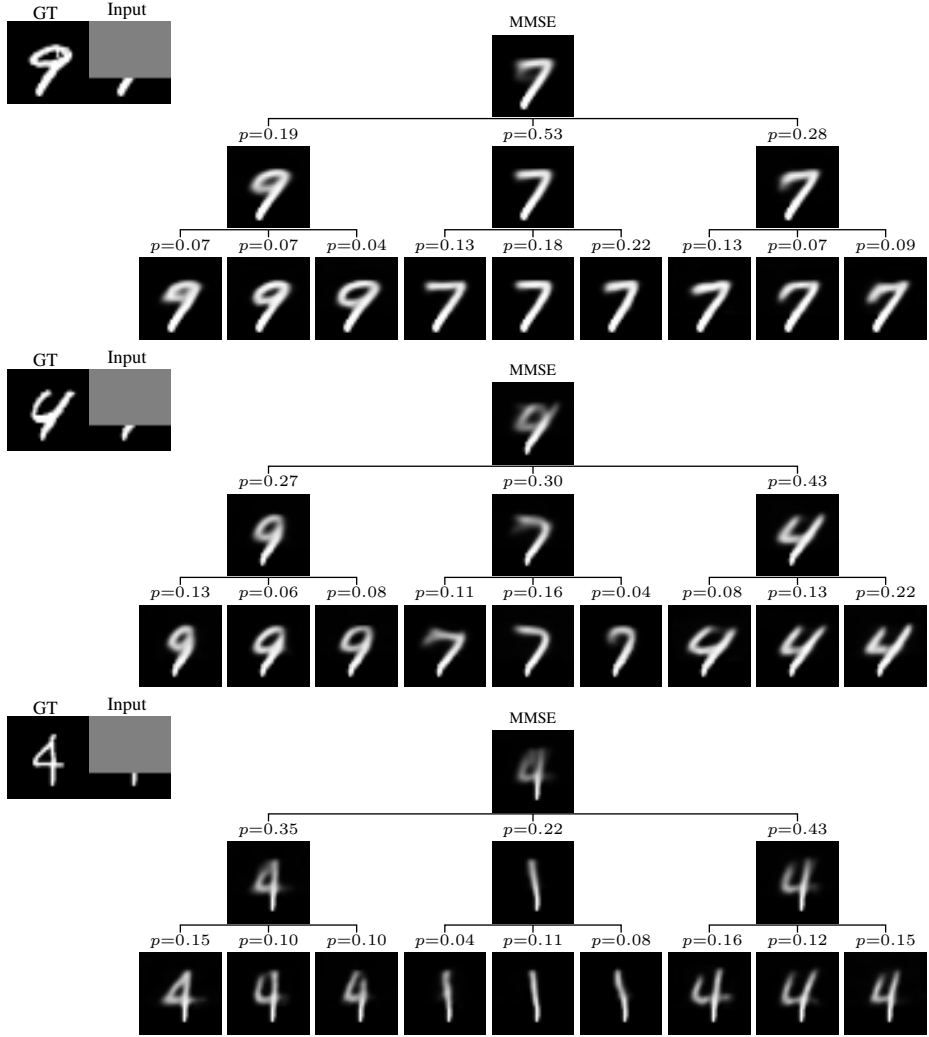


{\input{texfigures/MNIST_appendix_subfig1}}

\vspace{0.1cm}
{\input{texfigures/MNIST_appendix_subfig2}}

\vspace{0.1cm}
{\input{texfigures/MNIST_appendix_subfig3}}

\caption{\textbf{More digit inpainting results.}}
\label{fig:app-more-mnist}
\end{figure}

\begin{figure}[ht]

{\input{texfigures/Shoes_appendix_subfig3}}

\vspace{0.1cm}
{\input{texfigures/Shoes_appendix_subfig4}}

\caption{\textbf{More Edges$\rightarrow$Shoes results.}}
\label{fig:app-more-edges2shoes}
\end{figure}

\begin{figure}[ht]
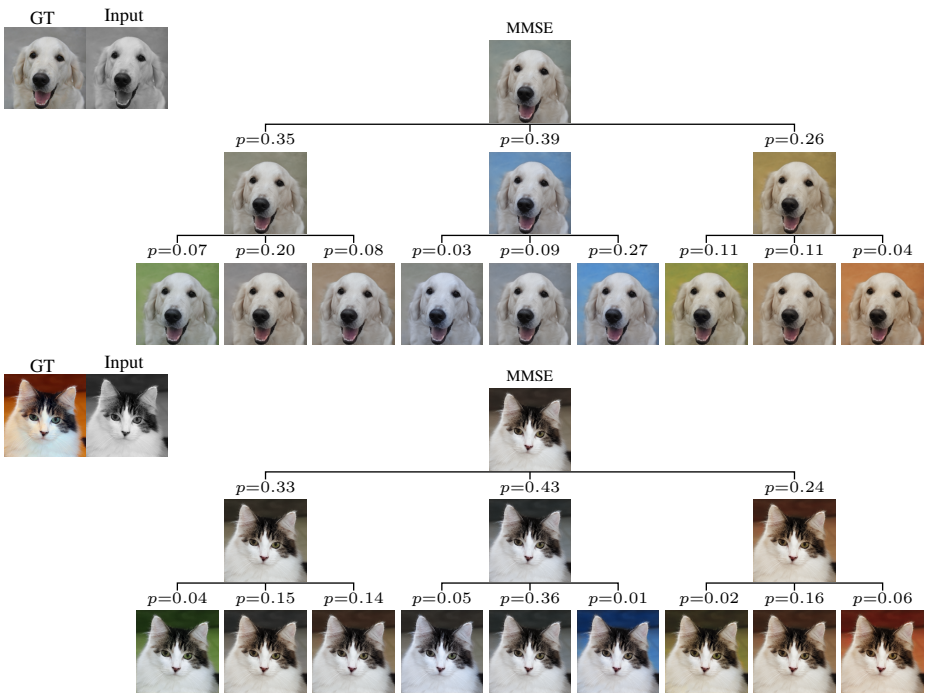


{\input{texfigures/AFHQ_colorization_appendix_subfig1}}

\vspace{0.1cm}
{\input{texfigures/AFHQ_colorization_appendix_subfig3}}

\caption{\textbf{AFHQ(v2) colorization results.}}
\label{fig:app-afhq-colorization}
\end{figure}

\begin{figure}[t]
\centering
\begin{subfigure}{\linewidth}
    {\input{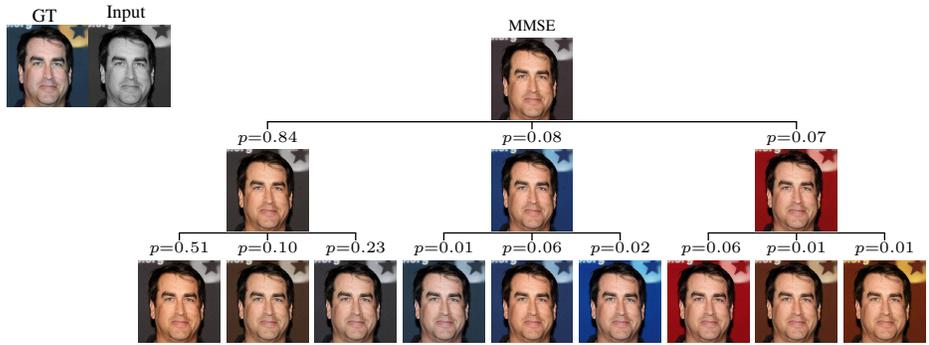}}
    \caption{\textbf{Colorization}}
    \label{fig:app-colorization}
    \vspace{0.2cm}
\end{subfigure}
\begin{subfigure}{\linewidth}
    \input{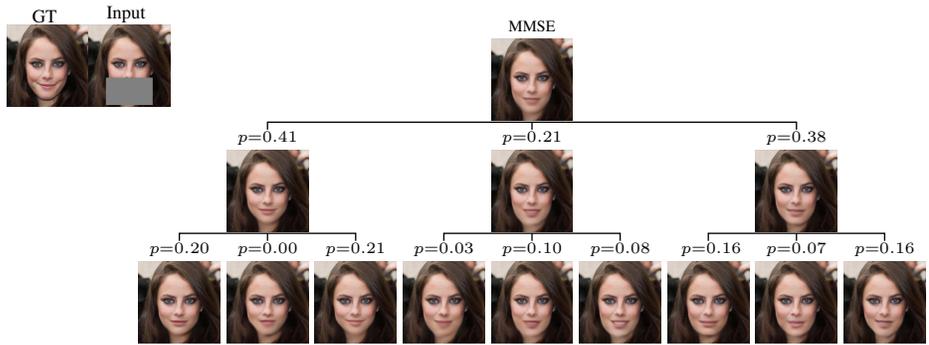}
    \caption{\textbf{Mouth inpainting}}
    \label{fig:app-mouth-inpaint}
    \vspace{0.2cm}
\end{subfigure}
\begin{subfigure}{\linewidth}
    \input{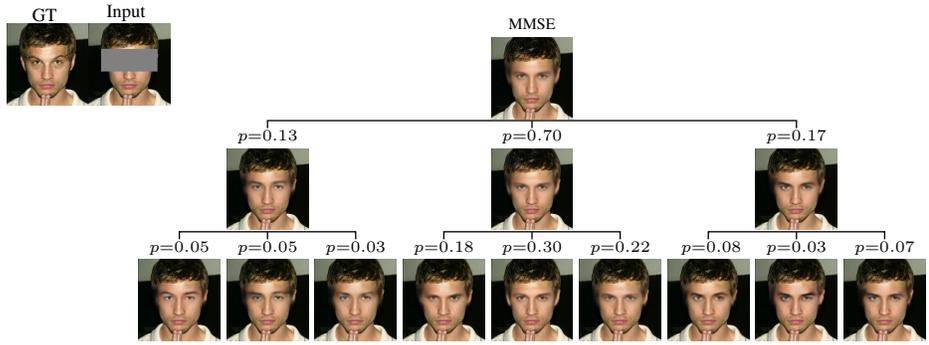}
    \caption{\textbf{Eyes inpainting}}
    \label{fig:app-eyes-inpainting}
    \vspace{0.2cm}
\end{subfigure}
\caption{\textbf{More CelebA-HQ results}.}
\end{figure}

\begin{figure}[ht]

{\input{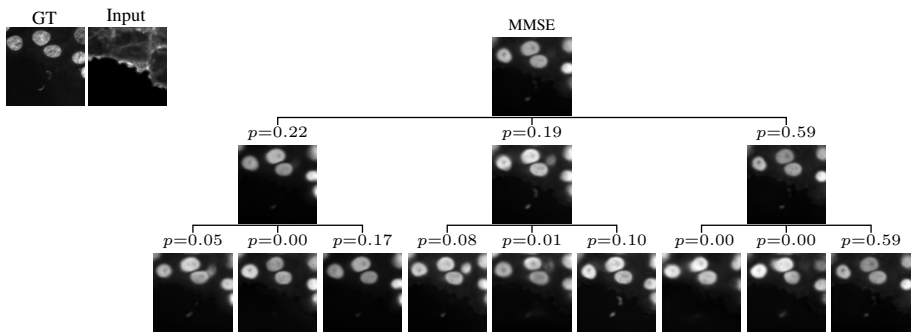}}


\caption{\textbf{More Bioimage translation results.}}
\label{fig:app-more-bio}
\end{figure}

\end{document}